\documentclass[twoside,11pt]{article}
\usepackage[margin=0.9in]{geometry}
\usepackage{authblk}

\usepackage[utf8]{inputenc} 
\usepackage[T1]{fontenc}    
\usepackage{hyperref}       
\usepackage{url}            
\usepackage{booktabs}       
\usepackage{amsfonts}       
\usepackage{nicefrac}       
\usepackage{microtype}      
\usepackage{natbib}
\usepackage{graphicx}
\usepackage{amssymb}
\PassOptionsToPackage{round}{natbib}

\usepackage{afterpage}
\usepackage{algorithm,algorithmic}
\newcommand{\vect}[1]{\boldsymbol{#1}}
\usepackage[usenames, dvipsnames]{color}
\def\therefore{\boldsymbol{\text{ }
\leavevmode
\lower0.4ex\hbox{$\cdot$}
\kern-.5em\raise0.7ex\hbox{$\cdot$}
\kern-0.55em\lower0.4ex\hbox{$\cdot$}
\thinspace\text{ }}}

\usepackage{mathtools}
\newcommand{\defeq}{\vcentcolon=}

\newcommand{\argmin}{\mathop{\mathrm{argmin}}}
\newcommand{\argmax}{\mathop{\mathrm{argmax}}}

\newcommand{\norm}{\mathcal{N}}

\newcommand{\dd}{\mathrm{d}}
\newcommand{\avec}{\mathbf{a}}
\newcommand{\bvec}{\mathbf{b}}

\newcommand{\uvec}{\mathbf{u}}

\newcommand{\xvec}{\mathbf{x}}

\newcommand{\yvec}{\mathbf{y}}

\newcommand{\zvec}{\mathbf{z}}

\newcommand{\appropto}{\mathrel{\vcenter{
  \offinterlineskip\halign{\hfil$##$\cr
    \propto\cr\noalign{\kern2pt}\sim\cr\noalign{\kern-2pt}}}}}

\usepackage{pdflscape}

\newtheorem{property}{Property}

\usepackage{cancel}

\usepackage{caption}
\usepackage{rotating}
\usepackage{lipsum}

\usepackage{subcaption}
\usepackage{cleveref}

\usepackage{xargs}                      
\usepackage[pdftex,dvipsnames]{xcolor}  
\usepackage[colorinlistoftodos,prependcaption,textsize=tiny]{todonotes}
\newcommandx{\unsure}[2][1=]{\todo[linecolor=red,backgroundcolor=red!25,bordercolor=red,#1]{#2}}
\newcommandx{\change}[2][1=]{\todo[linecolor=blue,backgroundcolor=blue!25,bordercolor=blue,#1]{#2}}
\newcommandx{\info}[2][1=]{\todo[linecolor=OliveGreen,backgroundcolor=OliveGreen!25,bordercolor=OliveGreen,#1]{#2}}
\newcommandx{\improvement}[2][1=]{\todo[linecolor=Plum,backgroundcolor=Plum!25,bordercolor=Plum,#1]{#2}}
\newcommandx{\thiswillnotshow}[2][1=]{\todo[disable,#1]{#2}}
\makeatletter
\newcommand{\mypm}{\mathbin{\mathpalette\@mypm\relax}}
\newcommand{\@mypm}[2]{\ooalign{%
  \raisebox{.1\height}{$#1+$}\cr
  \smash{\raisebox{-.6\height}{$#1-$}}\cr}}
\makeatother

\begin{document}

\title{Partitioned Variational Inference: A unified framework encompassing federated and continual learning}
\author[1]{Thang D.~Bui}
\author[2]{Cuong V.~Nguyen}
\author[2]{Siddharth Swaroop}
\author[2]{Richard E.~Turner}
\affil[1]{University of Sydney, Australia; thang.buivn@gmail.com}
\affil[2]{University of Cambridge, UK; \{vcn22,ss2163,ret26\}@cam.ac.uk}
\date{}                     
\setcounter{Maxaffil}{0}

\maketitle

\begin{abstract}
Variational inference (VI) has become the method of choice for fitting many modern probabilistic models. However, practitioners are faced with a fragmented literature that offers a bewildering array of algorithmic options. First, the variational family. Second, the granularity of the updates e.g.~whether the updates are local to each data point and employ message passing or global. Third, the method of optimization (bespoke or blackbox, closed-form or stochastic updates, etc.). This paper presents a new framework, termed Partitioned Variational Inference (PVI), that explicitly acknowledges these algorithmic dimensions of VI, unifies disparate literature, and provides guidance on usage. Crucially, the proposed PVI framework allows us to identify new ways of performing VI that are ideally suited to challenging learning scenarios including federated learning (where distributed computing is leveraged to process non-centralized data) and continual learning (where new data and tasks arrive over time and must be accommodated quickly). We showcase these new capabilities by developing communication-efficient federated training of Bayesian neural networks and continual learning for Gaussian process models with private pseudo-points. The new methods significantly outperform the state-of-the-art, whilst being almost as straightforward to implement as standard VI.

\end{abstract}



\section{Introduction}
\label{sec:introduction}
%
Variational methods recast approximate inference as an optimization problem, thereby enabling advances in optimization  
to be leveraged for inference. VI has enabled approaches including natural gradient methods, mirror-descent, trust region and stochastic (mini-batch) optimization to be tapped in this way. The approach has been successful, with VI methods often lying on the efficient frontier of approximate inference's speed-accuracy trade-off. VI has consequently become one of the most popular varieties of approximate inference. For example, it is now a standard approach for Gaussian process models \citep{titsias:2009a}, latent topic models \citep{blei+al:2003}, and deep generative models \citep{kingma+welling:2014}.

Deployment of VI requires the practitioner to make three fundamental choices. First, the {\it form of the approximate family} which ranges from simple mean-field or factorized distributions, through unfactorized exponential families to complex non-exponential family distributions. Second, the {\it granularity of variational inference} which includes, on the one hand, approaches based on the global variational free-energy, and on the other those that consider a single data point at a time and employ local message passing. Third, the {\it form of the variational updates} which encompasses the optimization method employed for maximizing the global variational free-energy or the form of the message passing updates. 

A large body of work has investigated how the choice of approximating family affects the accuracy of VI \citep{mackay:2003,wang+titterington:2004,turner+sahani:2011} and  how additional approximations can enable VI to support more complex approximate families \citep{jaakkola+jordan:1998,rezende+mohamed:2015,salimans+al:2015, ranganath+al:2016, mescheder+al:2017}. This is a fundamental question, but it is orthogonal to the focus of the current paper. Instead, we focus on the second two choices. The granularity of variational inference is an important algorithmic dimension. Whilst global variational inference has more theoretical guarantees and is arguably simpler to implement, local variational inference offers unique opportunities for online or continual learning (e.g.~allowing `old' data to be sporadically revisited) and distributed computing (e.g.~supporting asynchronous lock-free updates). The form of the updates is equally important with a burgeoning set of alternatives. For global VI these including gradient ascent, natural gradient and mirror descent, approximate second-order methods, stochastic versions thereof, collapsed VI and fixed-point updates to name but a few. For local VI, there has been less exploration of the options, but damping in natural and moment space is often employed.  

The goal of this paper is to develop a unifying framework, termed Partitioned Variational Inference (PVI), that explicitly acknowledges that the granularity and the optimization method are two fundamental algorithmic dimensions of VI. The new framework 1.~generalizes and extends current theoretical results in this area, 2.~reveals the relationship between a large number of existing schemes, and 3.~identifies opportunities for innovation, a selection of which are demonstrated in experiments. We briefly summarize the contributions of this paper, focusing on the unified viewpoint and novel algorithmic extensions to support federated and continual learning.

\subsection{Unification}
The main theoretical contributions of the paper, described in \cref{sec:PVI,sec:optim,sec:related-work}, are: to develop Partitioned Variational Inference; clean up, generalize and derive new supporting theory (including PVI fixed-point optimization, mini-batch approximation, hyperparameter learning); and show that PVI subsumes standard global variational inference, (local) variational message passing, and other well-established approaches. In addition, we also show in \cref{sec:related-work} that damped fixed-point optimization and natural gradient methods applied to PVI are equivalent to variationally-limited power EP.  

In \cref{sec:related-work} PVI is used to connect a large literature that has become fragmented with separated strands of related, but mutually uncited work. More specifically we unify work on: online VI \citep{ghahramani:2000,sato:2001,broderick+al:2013,bui+al:2017b,nguyen+al:2018}; global VI \citep{sato:2001,hensman+al:2012,hoffman+al:2013,salimans+knowles:13,sheth+khardon:2016,sheth+al:2015,sheth+khardon:2016b}; local VI \citep{knowles+minka:2011,wand:2014,khan+li:2018}; power EP and related algorithms \citep{minka:2001, minka:2004, li+al:2015,hasenclever+al:2017,gelman+al:2014}; and stochastic mini-batch variants of these algorithms \citep{hoffman+al:2013,li+al:2015,khan+li:2018}. \Cref{fig:pvi_special_cases,fig:past_work} and \cref{table:past-work} present a summary of these relationships in the context of PVI.

\subsection{Probabilistic inference for federated machine learning}

The goal of federated learning is to enable distributed training of machine learning models without centralizing data \citep[see e.g.][]{mcmahan+al:17,zhao+al:18}. This is challenging in practice as: 
\begin{itemize}
    \item modern data sets can often be distributed inhomogeneously and unevenly across many machines, for examples, mobile devices can contain many images which can be used for training a classification model, but accessing such information is often restricted and privacy-sensitive;
    \item computation resources available at terminal machines can be leveraged, but communication between these machines or between them and a central server can be limited and unreliable, for example, communication from and to mobile devices is often costly, and each device can be abruptly disconnected from the training setup or, similarly, a new device can appear;
    \item the inference or prediction step is often needed in an any-time fashion at each machine, i.e.~each machine needs to have access to a high-quality model to make predictions without having to send data to a remote server.
\end{itemize}
These requirements are often not satisfied in the traditional training pipelines, many of which require data to be stored in a single machine, or in a data center where it is typically distributed among many machines in a homogeneous and balanced fashion \citep[see e.g.][]{dean+al:12,zhang+al:15,chen+al:16}. Federated learning attempts to bridge this gap by tackling the aforementioned constraints. Additionally, this type of learning is arguably less privacy-sensitive as compared to centralized learning approaches, as it does not require local data to be collected and sent to a central server. It can also be further improved by employing encrypted aggregation steps \citep{bonawitz+al:17} or differentially-private mechanisms \citep{dwork+roth:14}.


Distributed inference is also an active research area in the Bayesian statistics and machine learning literature. For example, parallel Markov chain Monte Carlo approaches typically run multiple independent Markov chains on different partitions of the data set, but require heuristics to aggregate, reweight and average the samples at test time \citep[see e.g.][]{wang+dunson:13,scott+al:16}. The closest to our work is the distributed EP algorithms of \cite{gelman+al:2014} and \cite{hasenclever+al:2017}, which employ (approximate) MCMC for data partitions and EP for communication between workers. However, it is not clear these distributed approaches will work well in the federated settings described above. In \cref{sec:federated}, we demonstrate that PVI can naturally and flexibly address the above challenges, and thus be used for federated learning with efficient synchronous or lock-free asynchronous communication. The proposed approach can be combined with recent advances in Monte Carlo VI for neural networks, enabling fast and communication-efficient training of Bayesian neural networks on non-iid federated data. We provide an extensive experiment comparing to alternative approaches in \cref{sec:experiments}.

\subsection{Probabilistic inference for continual learning}

Continual learning (also termed online learning or life-long learning or incremental learning) is the ability to learn continually and adapt quickly to new experiences without catastrophically forgetting previously seen experiences \citep{schlimmer+fisher:86,mccloskey+cohen:89,sutton_whitehead:93,ratcliff:90}. 
Such requirements arise in many practical settings in which data can arrive sequentially or tasks may change over time (e.g.~new classes may be discovered), or entirely new tasks can emerge. 
Batch learning algorithms which deal with the entire data set at once are not applicable in these settings, as (1) data can arrive one point at a time or in batches of a size that is unknown a priori, or in a possibly non i.i.d.~way; and (2) previously seen data may not be directly accessible, which means the continual learning algorithms need to intelligently decide how to best combine prior or current experience with new data while being resistant to under-fitting or over-fitting to new data (i.e.~intransigence vs forgetting).

Continual learning has a rich literature \citep[see e.g.][]{opper:1999,sato:2001,ghahramani:2000,csato+opper:2002,minka:2001,smola+al:04} but is enjoying a resurgence of interest ranging from deepening understanding of transfer learning and catastrophic forgetting \citep{goodfellow+al:13,flesch+al:2018}, to developing learning algorithms for various models and applications \citep{broderick+al:2013,li+hoiem:16,kirkpatrick+al:17,zenke+al:17,seff+al:17,bui+al:2017,nguyen+al:2018,zeno+al:2018,chaudhry+al:2018}, to setting up relevant metrics and benchmarks for evaluation \citep{lomonaco_maltoni:2017,hayes+al:2018}. While the PVI framework enables us to connect and unify much of the literature in this area, it also allows gaps in the literature to be identified and enables the development of new and improved algorithmic solutions. We demonstrate this in \cref{sec:continual} by presenting a new continual learning method for Gaussian process regression and classification that greatly extends earlier work by \cite{csato+opper:2002} and \cite{bui+al:2017}, allowing principled handling of hyperparameters and private pseudo-points for new data.  The new technique is shown to be superior to alternative online learning approaches on various toy and real-world data sets in \cref{sec:experiments}. We also show in \cref{sec:federated} that continual learning can be reframed as a special case of federated learning.

\section{Partitioned Variational Inference}
\label{sec:PVI}
In this section, we introduce Partitioned Variational Inference, a framework that encompasses many approaches to variational inference. We begin by framing PVI in terms of a series of local variational free-energy optimization problems, proving several key properties of the algorithm that reveal the relationship to global VI. 
In order to keep the development clear, we have separated most of the discussion of related work into section \ref{sec:related-work}. 

Consider a parametric probabilistic model defined by the prior $p(\theta|\epsilon)$ over parameters $\theta$ and the likelihood function $p(\vect{y}|\theta,\epsilon) = \prod_{m=1}^{M}p(\vect{y}_m|\theta,\epsilon)$, where $\{ \vect{y}_1, \ldots, \vect{y}_M \}$ is a partition of $\vect{y}$ into $M$ groups of data points. Depending on the context, a data group $\vect{y}_m$ can be considered to be a mini-batch of $\vect{y}$ which is fixed across epochs, or a data shard. For simplicity, we assume for the moment that the hyperparameters $\epsilon$ are fixed and suppress them to lighten the notation. We will discuss hyperparameter optimization at the end of this section. 

Exact Bayesian inference in this class of model is in general intractable so we resort to variational inference. In particular, we posit a variational approximation of the true posterior as follows,
\begin{align}
    q(\theta) &= p(\theta) \prod_{m=1}^{M} t_m(\theta) 
    \approx \frac{1}{\mathcal{Z}} p(\theta) \prod_{m=1}^{M} p(\vect{y}_m|\theta) = p(\theta | \vect{y}), \label{eq:approx}
\end{align}
where $\mathcal{Z}$ is the normalizing constant of the true posterior, or marginal likelihood. The approximate likelihood $t_m(\theta)$ will be refined by PVI to approximate the effect the likelihood term $p(\vect{y}_m|\theta)$ has on the posterior. 
Note that the form of $q(\theta)$ in \eqref{eq:approx} is similar to that employed by the expectation propagation algorithm \citep{minka:2001}, but with two differences. First, the approximate posterior is not restricted to lie in the exponential family, as is typically the case for EP. Second, the approximate posterior does not include a normalizing constant. Instead, the PVI algorithm will automatically ensure that the product of the prior and approximate likelihood factors in \eqref{eq:approx} is a normalized distribution.  We will show that PVI will return an approximation to the marginal likelihood $\log \mathcal{Z} = \log p(\vect{y})$ in addition to the approximation of the posterior.

Algorithm \ref{alg:vmp} details the PVI algorithm. At each iteration $i$, we select an approximate likelihood to refine according to a schedule $b_i \in \{ 1 \hdots M \}$. The approximate likelihood $t^{(i-1)}_{b_i}(\theta)$ obtained from the previous iteration will be refined and the corresponding data-group is denoted $\vect{y}_{b_i}$. 
The refinement proceeds in two steps. First, we refine the approximate posterior using the local (negative) variational free energy $q^{(i)}(\theta) = \argmax_{q(\theta) \in \mathcal{Q}} \mathcal{F}^{(i)}(q(\theta))$ where the optimization is over a tractable family $\mathcal{Q}$ and
\begin{align}
\label{eq:vfe}
    \mathcal{F}^{(i)}(q(\theta)) = \int \mathrm{d}\theta q(\theta) \log \frac{q^{(i-1)}(\theta) p(\vect{y}_{b_i}|\theta)}{q(\theta) t^{(i-1)}_{b_i}(\theta)}.
\end{align}
Second, the new approximate likelihood is found by division, 
$t^{(i)}_{b_i}(\theta) = \frac{q^{(i)}(\theta)}{q^{(i-1)}(\theta)} t^{(i-1)}_{b_i}(\theta)$.

We will now justify these steps by stating properties, derived in the appendix, that show 1) the local free-energy optimization is equivalent to a variational KL optimization, 2)  the update for the approximate likelihoods is consistent with the normalized density specified in \ref{eq:approx}, and 3) any fixed point of the algorithm is also a local optimum of global VI and at this fixed point the sum of the local free-energies is equal to the global variational free-energy. The following properties apply for general $q(\theta)$, and are not limited to the exponential family.\footnote{However, we will only consider exponential family approximations in the experiments in \cref{sec:experiments}.}

\begin{property}
\label{prop:local-fe-opt}
Maximizing the local free-energy $\mathcal{F}^{(i)}(q(\theta))$ is equivalent to the  KL optimization
\begin{align}
    q^{(i)}(\theta) = \argmin_{q(\theta) \in \mathcal{Q}} \mathrm{KL} \left( q(\theta) ~\|~ \widehat{p}^{(i)}(\theta) \right),
\end{align}
where $\widehat{p}^{(i)}(\theta) = \frac{1}{\widehat{\mathcal{Z}}_i} \frac{q^{(i-1)}(\theta)}{t^{(i-1)}_{b_i}(\theta)} p(\vect{y}_{b_i}|\theta) = \frac{1}{\widehat{\mathcal{Z}}_i} p(\vect{y}_{b_i}|\theta) \prod_{m \ne b_i } t^{(i-1)}_{m}(\theta) $ is known as the tilted distribution in the EP literature and is intractable. 
\end{property}

The proof is straightforward (see \ref{sec:appen:local-FE-KL}). The tilted distribution can be justified as a sensible target as it removes the approximate likelihood $t^{(i-1)}_{b_i}(\theta)$ from the current approximate posterior and replaces it with the true likelihood $p(\vect{y}_{b_i}|\theta)$. In this way, the tilted distribution comprises one true likelihood, $M-1$ approximate likelihoods and the prior. The KL optimization then ensures the new posterior better approximates the true likelihood's effect, in the context of the approximate likelihoods and the prior. 

\begin{property}
\label{prop:valid-prob}
At the end of each iteration $i = 0, 1, \ldots$, $q^{(i)}(\theta) = p(\theta) \prod_{m=1}^{M} t^{(i)}_m(\theta)$.
\end{property}
Again the proof is simple (see \ref{sec:appen:normalizedt}), but it relies on PVI initializing the approximate likelihood factors to unity so that $q^{(0)}(\theta) = p(\theta)$. 

\begin{property}
\label{prop:fixed-point}
\sloppy Let $q^*(\theta) = p(\theta) \prod_{m=1}^{M} t^*_m(\theta)$ be a fixed point of Algorithm \ref{alg:vmp}, $\mathcal{F}_m(q(\theta)) = \int \mathrm{d}\theta q(\theta) \log \frac{q^*(\theta) p(\vect{y}_m|\theta)}{q(\theta) t^*_m(\theta)}$ be the local free-energy w.r.t.~the factor $t_m$, and $\mathcal{F}(q(\theta)) = \int \mathrm{d}\theta q(\theta) \log \frac{p(\theta) \prod_{m=1}^{M} p(\vect{y}_m|\theta)}{q(\theta)}$ be the global free-energy. We have:

(a) $\sum_{m=1}^M \mathcal{F}_m(q^*(\theta)) = \mathcal{F}(q^*(\theta))$, i.e.~the sum of the local free-energies is equal to the global free-energy, i.e.~the PVI fixed point is an optimum of global VI,

(b) If $q^*(\theta) = \argmax_{q(\theta) \in \mathcal{Q}} \mathcal{F}_m(q(\theta))$ for all $m$, then $q^*(\theta) = \argmax_{q(\theta) \in \mathcal{Q}}  \mathcal{F}(q(\theta))$.
\end{property}

These results are more complex to show, but can be derived by computing the derivative and Hessian of the global free-energy and substituting into these expressions the derivatives and Hessians of the local free-energies (see \ref{sec:appen:local-global}). The fact that the fixed point of PVI recovers a global VI solution (both the optimal $q(\theta)$ and the global free-energy at this optimum) is the main theoretical justification for employing PVI. However, we do not believe that there is a Lyapunov function for PVI, indicating that it may oscillate or diverge in general.

\begin{algorithm}[tb]
   \caption{Partitioned Variational Inference}
   \label{alg:vmp}
\begin{algorithmic}
   \STATE {\bfseries Input:} data partition $\{ \vect{y}_1, \ldots, \vect{y}_M \}$, prior $p(\theta)$ \\[5pt]
   \STATE Initialize:
     \begin{align*}
        t^{(0)}_m(\theta) &\defeq 1 \text{ for all } m = 1, 2, \ldots, M. \\
        q^{(0)}(\theta) &\defeq p(\theta).
     \end{align*}
   \FOR{$i=1,2,\ldots$ until convergence}
       \STATE $b_i \defeq$ index of the next approximate likelihood to refine.
       \STATE Compute the new approximate posterior:
       \begin{equation}
%
 q^{(i)}(\theta) \defeq \argmax_{q(\theta) \in \mathcal{Q}} \int \mathrm{d}\theta q(\theta) \log \frac{q^{(i-1)}(\theta) p(\vect{y}_{b_i}|\theta)}{q(\theta) t^{(i-1)}_{b_i}(\theta)} \nonumber
       \end{equation}
       \STATE Update the approximate likelihood:
       \begin{align}
       t^{(i)}_{b_i}(\theta) &\defeq \frac{q^{(i)}(\theta)}{q^{(i-1)}(\theta)} t^{(i-1)}_{b_i}(\theta), \\
       t^{(i)}_m(\theta) &\defeq t^{(i-1)}_m(\theta) \text{ for all } m \neq b_i. \notag
       \end{align}
   \ENDFOR
\end{algorithmic}
\end{algorithm}

Having laid out the general framework for PVI, what remains to be decided is the method used for optimizing the local free-energies. In a moment we consider three choices: analytic updates, off-the-shelf optimization methods and fixed-point iterations, as well as discussing how stochastic approximations can be combined with these approaches. Before turning to these choices, we compare and contrast the algorithmic benefits of the local and global approaches to VI in different settings. This discussion will help shape the development of the optimization choices which follows.

\subsection{When should a local VI approach be employed rather than a global one?}

\begin{figure}[!ht]
    \centering
    \includegraphics[width=\textwidth]{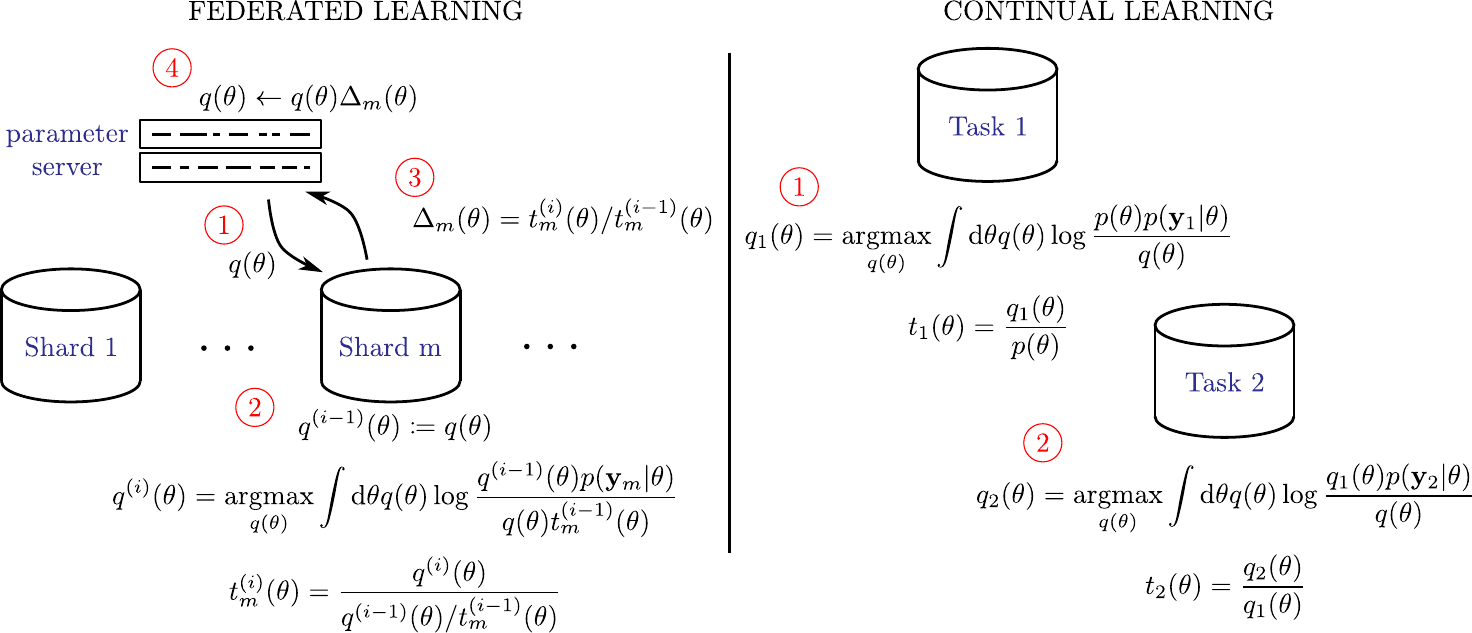}
    \caption{Steps of the PVI algorithm when being used for continual learning [left] and federated learning [right].}
    \label{fig:pvi_illustration}
\end{figure}

We will describe in section \ref{sec:related-work} how the PVI framework unifies a large body of existing literature, thereby providing a useful conceptual scaffold for understanding the relationship between algorithms. However, it is important to ask: What algorithmic and computation benefits, if any, arise from considering a set of local free-energy updates, rather than a single global approximation (possibly leveraging stochastic mini-batch approximation)?

In a nutshell, we will show that if the data set is fixed before inference is performed (batch learning) or arrives in a simple online iid way (simple online learning), and distributed computation is not available, then global VI will typically be simpler to implement, require less memory, and faster to converge than more local versions of PVI (the case of scaling collapsed bounds being a possible exception). However, if the conditions above are not met, the local versions of PVI will be appropriate. We will now unpack important examples of this sort.

The PVI approach is ideally suited to the distributed setting, with simple distributed variants allowing asynchronous distributed updates. One simple approach, similar to that of \cite{hasenclever+al:2017}, uses $M$ workers that are each allocated a data group $\vect{y}_m$. The workers store and refine the associated approximate likelihood $t_{m}(\theta)$. A server maintains and updates the approximate posterior and communicates it to the workers. An idle worker receives the current posterior from the server, optimizes the local free-energy, computes the change in the local approximate likelihood $\Delta_m(\theta) = t_{m}^{(\text{new})}(\theta) / t_{m}^{(\text{old})}(\theta)$, sends this to the server, and repeats. The local workers do not change $q(\theta)$ directly. Instead, the server maintains a queue of approximate likelihood updates and applies these to the approximate posterior $q^{(\text{new})}(\theta) = q^{(\text{old})}(\theta) \Delta_m(\theta)$. This setup supports asynchronous updates of the approximate likelihood factors. See \cref{fig:pvi_illustration} for a pictorial depiction of these steps. 
In contrast, global VI is generally ill-suited to the distributed setting. Although the free-energy optimization can be parallelized over data points,  typically this will only be advantageous for large mini-batches where the extra communication overhead does not dominate. Large mini-batches often result in slow optimization progress (early in learning it is often clear how to improve $q(\theta)$ after seeing only a small number of data points). The special case of global VI employing mini-batch approximations and natural gradient updates can support asynchronous distributed processing if each worker receives statistically identical data and updates with the same frequency. It could not operate successfully when each node contains different amounts or types of data, or if some workers update more frequently than others.

Distributed versions of PVI not only enable VI to be scaled to large problems, but they also allow inference algorithms to be sent to user data, rather than requiring user data to be collected and centralized before performing inference. Consider the situation where workers are personal devices, like mobile phones, containing user data $\vect{y}_m$. Here the local free-energy updates can be performed client-side on the user's devices and only summaries $t_{m}(\theta)$ of the relevant aspects of that information are communicated back to the central server. The frequency with which these messages are sent might be limited to improve security. Such an implementation is arguably more secure than one in which the user data (or associated gradients) are sent back to a central server \citep{royal-soc:2017}. Since the amount and type of data at the nodes is outside of the control of the algorithm designer, mini-batch natural gradient global VI will generally be inappropriate for this setting.

The PVI approach is also well suited to the continual or life-long learning setting. These settings are very general forms of online learning in which new data regularly arrive in a potentially non-iid way, tasks may change over time, and entirely new tasks may emerge. In this situation, the PVI framework can not only be used to continuously update the posterior distribution $q(\theta)$ in light of new data by optimizing the local free-energy for the newly seen data, it can also be used to revisit old data groups (potentially in a judiciously selected way) thereby mitigating problems like catastrophic forgetting. The update steps for this learning scenario are illustrated in \cref{fig:pvi_illustration}.
In contrast, global VI is fundamentally ill-suited to the general online setting. The special case of global VI employing mini-batch approximations with natural gradient updates may be appropriate when the data are iid and only one update is performed for each new task (simple online learning), but it is not generally applicable.

We will return to discuss the key issues raised in this section -- the speed of convergence, memory overhead, online learning, and distributed inference -- in the context of different options for carrying out the optimization of the local free-energies in \cref{sec:optim}.

\subsection{Hyperparameter Learning}

Many probabilistic models depend on a set of hyperparameters $\epsilon$ and it is often necessary to learn suitable settings from data to achieve good performance on a task. One method is to optimize the variational free-energy thereby approximating maximum likelihood learning. The gradient of the global variational free-energy decomposes into a set of local computations, as shown in appendix \ref{sec:hyper-gradients},
\begin{align}
\frac{\mathrm{d}}{ \mathrm{d} \epsilon} \mathcal{F}(\epsilon,q(\theta)) =& \sum_{m=1}^M \mathbb{E}_{q(\theta)} \left [\frac{\mathrm{d}}{ \mathrm{d} \epsilon} \log p(\mathbf{y}_m|\theta,\epsilon) \right] + \mathbb{E}_{q(\theta)} \left[\frac{\mathrm{d}}{ \mathrm{d} \epsilon} \log p(\theta|\epsilon) \right].
\end{align}
This expression holds for general $q(\theta)$ and is valid both for coordinate ascent (updating $\epsilon$ with $q(\theta)$ fixed) and for optimizing the collapsed bound (where the approximate posterior optimizes the global free-energy $q(\theta) = q^*(\theta)$ and therefore depends implicitly on $\epsilon$). Notice that this expression is amenable to stochastic approximation which leads to optimization schemes that use only local information at each step.  When combined with different choices for the optimization of the local free-energies wrt $q(\theta)$, this leads to a wealth of possible hyperparameter optimization schemes.

In cases where a distributional estimate for the hyperparameters is necessary, e.g.~in continual learning, the PVI framework above can be extended to handle the hyperparameters. In particular, the approximate posterior in \cref{eq:approx} can be modified as follows,
\begin{align}
    q(\theta, \epsilon) &= p(\epsilon) p(\theta|\epsilon)  \prod_{m=1}^{M} t_m(\theta, \epsilon) 
    \approx \frac{1}{\mathcal{Z}} p(\epsilon) p(\theta|\epsilon) \prod_{m=1}^{M} p(\vect{y}_m|\theta, \epsilon) = p(\theta, \epsilon | \vect{y}), \label{eq:approx_hyper},
\end{align}
where the approximate likelihood factor $t_m(\theta, \epsilon)$ now involves both the model parameters and the hyperparameters. Similar to \cref{eq:vfe}, the approximate posterior above leads to the following local variational free-energy,
\begin{align}
\label{eq:vfe_hyper}
    \mathcal{F}^{(i)}(q(\theta,\epsilon)) = \int \mathrm{d}\theta\mathrm{d}\epsilon q(\theta,\epsilon) \log \frac{q^{(i-1)}(\theta, \epsilon) p(\vect{y}_{b_i}|\theta, \epsilon)}{q(\theta, \epsilon) t^{(i-1)}_{b_i}(\theta, \epsilon)}.
\end{align}
Note that this approach retains all favourable properties of PVI such as local computation and flexibility in choosing optimization strategies and stochastic approximations.

\section{Approaches for Optimizing the Local Free-energies}
\label{sec:optim}

Having established the general PVI algorithm and its properties, we will now describe different options for performing the optimization of the local free-energies.

\subsection{Analytic Local Free-energy Updates}

Each local free-energy is equivalent in form to a global free-energy with an effective prior $p_{\mathrm{eff}}(\theta) = q^{(i-1)}(\theta)/ t^{(i-1)}_{b_i}(\theta)$. As such, in conjugate exponential family models the KL optimizations will be available in closed form, for example in GP regression, and these updates can be substituted back into the local variational free-energies to yield locally-collapsed bounds, $\mathcal{F}_n(q^{(i)}(\theta))$, that are useful for hyperparameter optimization \citep{bui+al:2017}. One advantage of using local versions of PVI is that this allows collapsed bounds to be leveraged on large data sets where an application to entire data set would be computationally intractable, potentially speeding up convergence over global VI.

\subsection{Off-the-shelf Optimizers for Local Free-energy Optimization}
If analytic updates are not tractable, the local free-energy optimizations can be carried out using standard optimizers. The PVI framework automatically breaks the data set into a series of local free-energy optimization problems and the propagation of uncertainty between the data groups weights the information extracted from each. This means non-stochastic optimizers such as BFGS can now be leveraged in the large data setting. Of course, if a further stochastic approximation like Monte Carlo VI is employed for each local optimization, stochastic optimizers such as RMSProp \citep{tieleman+hinton:2012} or Adam \citep{kingma+ba:2014} might be more appropriate choices. In all cases, since the local free-energy is equivalent in form to a global free-energy with an effective prior $p_{\mathrm{eff}}(\theta) = q^{(i-1)}(\theta)/ t^{(i-1)}_{b_i}(\theta)$, PVI can be implemented via trivial modification to existing code for global VI. This is a key advantage of PVI over previous local VI approaches, such as variational message passing \citep{winn+bishop:2005,winn+minka:2009,knowles+minka:2011}, in which bespoke and closed-form updates are needed for different likelihoods and cavity distributions. 

\subsection{Local Free-energy Fixed Point Updates, Natural Gradient Methods, and Mirror Descent}
An alternative to using off-the-shelf optimizers is to derive fixed-point update equations by zeroing the gradients of the local free-energy. These fixed-point updates have elegant properties for approximate posterior distributions that are in the exponential family.
\begin{property}
\label{prop:fixed-point-equations}
If the prior and approximate likelihood factors are in the un-normalized exponential family $t_m(\theta) = t_m(\theta ; \eta_m) = \exp(\eta_m^\intercal T(\theta))$ so that the variational distribution is in the normalized exponential family $q(\theta) = \exp( \eta_q^\intercal T(\theta) - A(\eta_q) )$, then the stationary point of the local free-energy $\frac{\mathrm{d}\mathcal{F}^{(i)}(q(\theta))}{\mathrm{d}\eta_q}  = 0$ implies
\begin{align}
\label{eq:fixed-point-bi}
\eta^{(i)}_{b_i}  &= \mathbb{C}^{-1} \frac{\mathrm{d}}{\mathrm{d}\eta_q} \mathbb{E}_q (\log p(\vect{y}_{b_i}|\theta)).
\end{align}
where $\mathbb{C} \defeq \frac{\mathrm{d}^2A(\eta_q)}{\mathrm{d}\eta_q\mathrm{d}\eta_q} = \mathrm{cov}_{q(\theta)}[ T(\theta) T^\intercal(\theta)]$ is the Fisher Information. Moreover, the Fisher Information can be written as $\mathbb{C} = \frac{\mathrm{d}\mu_q}{\mathrm{d}\eta_q}$ where $\mu_q = \mathbb{E}_q (T(\theta))$ is the mean parameter of $q(\theta)$. Hence,
\begin{align}
\label{eq:fixed-point-bi}
\eta^{(i)}_{b_i}  &
= \frac{\mathrm{d}}{\mathrm{d}\mu_q} \mathbb{E}_q (\log p(\vect{y}_{b_i}|\theta)).
\end{align}
For some approximate posterior distributions $q(\theta)$, taking derivatives of the average log-likelihood with respect to the mean parameters is analytic (e.g.~Gaussian) and for some it is not (e.g.~gamma).
\end{property}
These conditions, derived in appendix \ref{sec:appen:fp}, can be used as fixed point equations. That is, they can be iterated possibly with damping $\rho$, 
\begin{align}
\label{eq:fixed-point-bi-damp}
\eta^{(i)}_{b_i}  &
= (1-\rho)\eta^{(i-1)}_{b_i} + \rho \frac{\mathrm{d}}{\mathrm{d}\mu_q} \mathbb{E}_q (\log p(\vect{y}_{b_i}|\theta)).
\end{align}
These iterations, which form an inner-loop in PVI, are themselves not guaranteed to converge (there is no Lyapunov function in general and so, for example, the local free-energy will not reduce at every step).

The fixed point updates are the natural gradients of the local free-energy and the damped versions are natural gradient ascent \citep{sato:2001,hoffman+al:2013}. The natural gradients could also be used in other optimization schemes \citep{hensman+al:2012,salimbeni+al:2018}. The damped updates are also equivalent to performing mirror-descent \citep{raskutti+mukherjee:2015, khan+li:2018}, a general form of proximal algorithm \citep{parikh+boyd:2014} that can be interpreted as trust-region methods. For more details about the relationship between these methods, see appendix \ref{sec:nat-grad-mirror-trust}. Additionally, while natural gradients or fixed-point updates have been shown to be effective in the batch global VI settings \citep[see e.g.][]{honkela+al:2010}, we present some result in \cref{sec:app:bnn_opt} showing adaptive first-order methods employing flat gradients such as Adam \citep{kingma+ba:2014} performs as well as natural gradient methods, when stochastic mini-batch approximations are used. 

For these types of updates there is an interesting relationship between PVI and global (batch) VI:

\begin{property} \label{prop:local-global-fp}
PVI methods employing parallel updates result in identical dynamics for $q(\theta)$ given by the following equation, regardless of the partition of the data employed

\begin{align}
\eta_q^{(i)} = \eta_0 + \frac{\mathrm{d}}{\mathrm{d}\mu_{q^{(i-1)}}} \mathbb{E}_q (\log p(\vect{y}|\theta)) = \eta_0 + \sum_{n=1}^N \frac{\mathrm{d}}{\mathrm{d}\mu_{q^{(i-1)}}} \mathbb{E}_{q^{(i-1)}} (\log p(y_{n}|\theta)).
\end{align}
\end{property}

See \ref{sec:local-global-same} for the proof. If parallel fixed-point updates are desired, then it is more memory efficient to employ batch VI $M=1$, since then only one global set of natural parameters needs to be retained. However, as previously discussed, using $M=1$ gives up opportunities for online learning and distributed computation (e.g.~asynchronous updates).  

\subsection{Stochastic mini-batch approximation}
\label{section:mini-batch}

There are two distinct ways to apply stochastic approximations within the PVI scheme. 

\subsubsection{Stochastic Approximation within the Local Free-Energy}
\label{sec:stoc_approx_1}

The first form of stochastic approximation leverages the fact that each local free-energy decomposes into a sum over data points and can, therefore, be approximated by sampling mini-batches within each data group $\vect{y}_m$. In the case where each partition includes a large number of data points, this leads to algorithms that converge more quickly than the batch variants -- since a reasonable update for the approximate posterior can often be determined from just a few data points -- and this faster convergence opens the door to processing larger data sets.

Mini-batch approximation can be employed in the general PVI case, but for simplicity we consider the global VI case here $M=1$. If simplified fixed point updates are used for optimization, then sampling $L$ mini-batches of data from the data distribution $\mathbf{y}_l \stackrel{\text{iid}}{\sim} p_{\text{data}}(y)$ yields the following stochastic approximation to the damped updates,\footnote{We have used a distinct notation for a mini-batch ($\vect{y}_l$) and a data group ($\vect{y}_m$) since the former will be selected iid from the data set and will vary at each epoch, whilst the latter need not be determined in this way and is fixed across epochs.}
\begin{align}
\eta^{(i)}_{q}  &
= (1-\rho)\eta^{(i-1)}_{q} + \rho \left ( \eta_0 + L \frac{\mathrm{d}}{\mathrm{d}\mu_q} \mathbb{E}_q (\log p(\vect{y}_l|\theta)) \right),\\
& = \eta^{(i-1)}_{q} + \rho' \left (   \frac{\mathrm{d}}{\mathrm{d}\mu_q} \mathbb{E}_q (\log p(\vect{y}_l|\theta)) - \eta^{(i-1)}_{\text{like}}/L \right).
\label{eq:stoch-global-main}
\end{align}
Here the first form of the update is stochastic natural gradient ascent and the second form reveals the implied deletion step where $\eta^{(i-1)}_{\text{like}}/L = (\eta^{(i-1)}_{q} - \eta^{(i-1)}_{0})/L$ is the contribution a mini-batch likelihood makes to the posterior natural parameters on average. The rescaled learning rate is $\rho' = L \rho$. These two forms reveals that the mini-batch stochastic natural gradient update resembles an EP update step. See appendix \ref{sec:stochastic-approx} for full details.

\subsubsection{Stochastic Scheduling of Updates Between Local Free-Energies} 
\label{sec:stoc_approx_2}

The second form of stochastic approximation is to randomize the update schedule. For example, using $M=N$ and randomly selecting subsets of data to update in parallel. This can be memory intensive, requiring $N$ local natural parameters to be stored. A more memory efficient approach is to fix the mini-batches across epochs and to visit the data groups $\vect{y}_m$ in a random order \citep{khan+li:2018}. For the simplified fixed point updates, this yields  
\begin{align}
\eta^{(i)}_{m}  & = (1-\rho) \eta^{(i-1)}_{m} + \rho \frac{\mathrm{d}}{\mathrm{d}\mu_{q^{(i-1)}}} \mathbb{E}_{q^{(i-1)}} (\log p(\vect{y}_{m}|\theta)). \label{eq:stoch-local-main}
\end{align}
This approach results in a subtly different update to $q$ that retains a specific approximation to the likelihood of each data partition, rather than a single global approximation
\begin{align}
\eta^{(i)}_{q}  &
= \eta^{(i-1)}_{q} - \rho \left ( \frac{\mathrm{d}}{\mathrm{d}\mu_q} \mathbb{E}_q (\log p(\vect{y}_m|\theta)) - \eta^{(i-1)}_{m} \right).\label{eq:stoch-local-EP-main}
\end{align}
If the first approach in \cref{eq:stoch-local-main} employs learning rates that obey the Robins Munro conditions, the fixed points will be identical to the second approach in \cref{eq:stoch-local-EP-main} and they will correspond to optima of the global free-energy.

\subsubsection{Comparing and Contrasting Stochastic Approaches} 

There are pros and cons to both approaches. The first approach in \cref{sec:stoc_approx_1} has a memory footprint $L$ times smaller than the second approach in \cref{sec:stoc_approx_2} and can converge more quickly. For example, on the first pass through the data, it effectively allows approximate likelihoods for as of yet unseen data to be updated based on those for the data seen so far, which means that larger learning rates can be used $\rho' > \rho$. The second approach is required for continual learning, asynchronous updates, and client-side processing where the assumption that each mini-batch is iid (and a single gradient step is performed on each) is typically incorrect. The second approach also tends to produce less noisy learning curves, with stochasticity only entering via the schedule and not as an approximation to the local free-energy and the gradients thereof. 

These approaches could also be combined, with stochastic scheduling selecting the local free-energy to update next and mini-batch updates employed for each local free-energy optimization. See appendix \ref{sec:stochastic-approx} for a full discussion.

\section{Unification of Previous Work}
\label{sec:related-work}
\begin{figure}[!ht]
\centering
\includegraphics[scale=1.1]{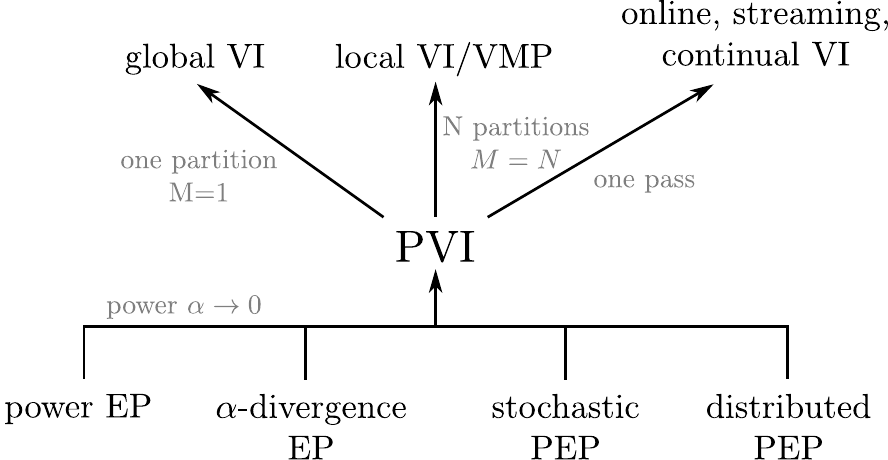}
\caption{Variational inference schemes encompassed by the PVI framework.}
\label{fig:pvi_special_cases}
\end{figure}

\begin{figure}[!ht]
\includegraphics[width=\linewidth]{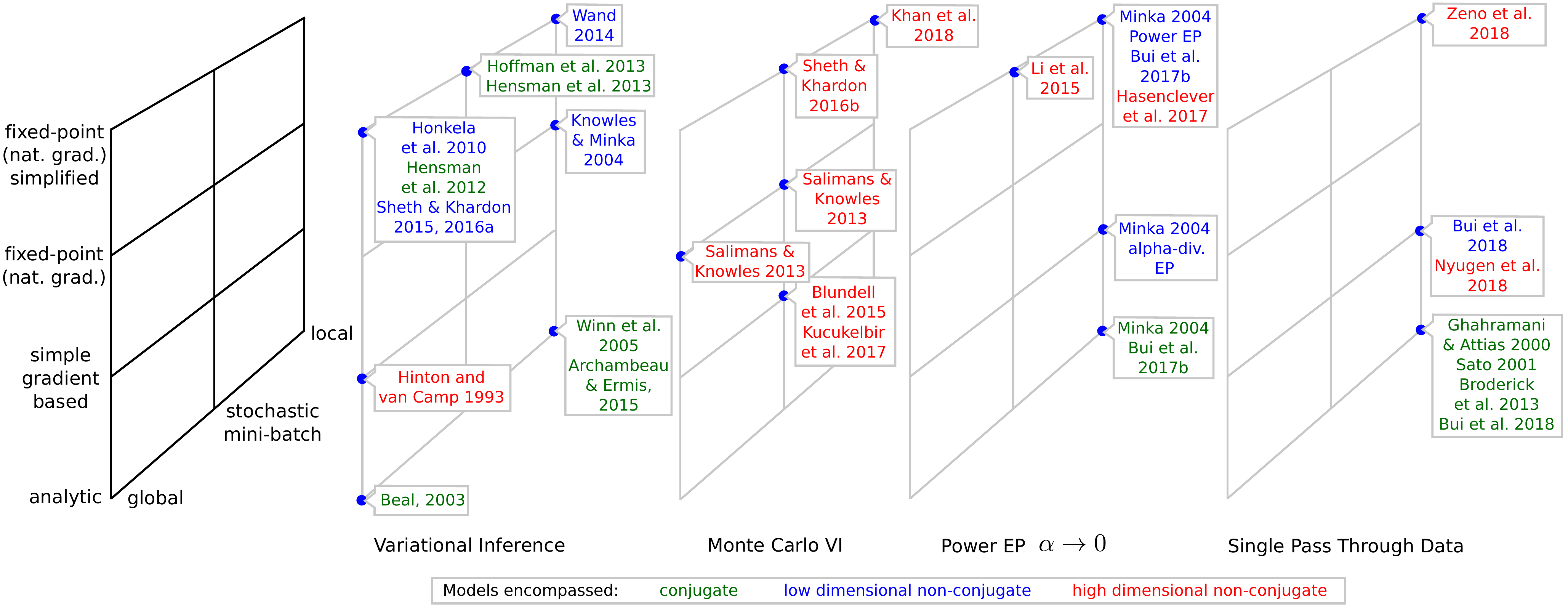}
\caption{The local VI framework unifies prior work. The granularity of the approximation and the optimization method employed are two fundamental algorithmic dimensions that are shown as axes. Fixed-point updates are identical to natural gradient ascent with unit step size. The models encompassed by each paper are indicated by the color. See \ref{table:past-work} for more information.}
\label{fig:past_work}
\end{figure}

The local VI framework described above unifies a large number of existing approaches. These methods include global VI (\cref{sec:global}), local VI (\cref{sec:local}), online VI (\cref{sec:online}) and a number of methods based on power EP (\cref{sec:pep,sec:alpha_ep,sec:stochastic_pep,sec:distributed_pep}). A schematic showing the relationships between these methods at a high level is shown in \cref{fig:pvi_special_cases}. The literature has been organized into in \cref{fig:past_work} and \cref{table:past-work}.

\subsection{Global VI Fixed Point Methods}
\label{sec:global}

There has been a long history of applying the fixed point updates for global VI (PVI where $M=1$). \citet{sato:2001} derived them for conjugate exponential family models, showing they recover the closed form updates for $q(\theta)$, and noting that damped fixed point updates are equivalent to natural gradient ascent with unit step size ($\rho =1$). 
Sato's insight was subsequently built upon by several authors. \cite{honkela+al:2010} considered non-conjugate models, employed a Gaussian variational distribution and used natural gradients to update the variational distribution's mean.
\citet{hensman+al:2012} and \citet{hoffman+al:2013} applied the insight to conjugate models when optimizing collapsed variational free-energies and deriving stochastic natural gradient descent, respectively. \citet{salimans+knowles:13} apply the fixed points to non-conjugate models where the expectations over $q$ are intractable and use Monte Carlo to approximate them, but they explicitly calculate the Fisher information matrix, which is unnecessary for exponential family $q$. 
\citet{sheth+khardon:2016} and \citet{sheth+al:2015} treat non-conjugate models with Gaussian latent variables, employ the cancellation of the Fisher information, and analyze convergence properties. \citet{sheth+khardon:2016b} further extend this to two level-models through Monte Carlo essentially applying the Fisher information cancellation to \citet{salimans+knowles:13}, but they were unaware of this prior work. 

\subsection{Fully Local VI Fixed Point Methods} 
\label{sec:local}

There has also been a long history of applying the fixed point updates in the fully local VI setting (where $M=N$). \citet{knowles+minka:2011} derive them for non-conjugate variational message passing, but explicitly calculate the Fisher information matrix (except in a case where $q$ was univariate Gaussian case where they do employ the cancellation). \citet{wand:2014} simplified VMP by applying the Fisher information cancellation to the case where $q(\theta)$ is multivariate Gaussian. \cite{khan+li:2018} also extend VMP to employ MC approximation and the Fisher information cancellation. They were unaware of \citet{wand:2014}, but extend this work by treating a wider range of approximate posteriors and models, stochastic updates, and a principled approach to damping. The work is closely related to  \cite{salimans+knowles:13} and \cite{sheth+khardon:2016b}, since although these papers use fixed-point updates for global VI, they show that these decompose over data points and thereby derive mini-batch updates that closely resemble fixed-point local VI. This is a result of property \ref{prop:local-global-fp}.


\subsection{Online, Streaming, Incremental and Continual VI as a single pass of PVI}
\label{sec:online}

If PVI makes a single pass through the data, the approximate likelihoods do not need to be explicitly computed or stored as data-groups are not revisited. In this case PVI reduces to initializing the approximate posterior to be the prior, $q^{(0)}(\theta) = p(\theta)$  and then optimizing a sequence of local free-energies
\begin{align}
q^{(i)}(\theta) \defeq \argmax_{q(\theta) \in \mathcal{Q}} \int \mathrm{d}\theta q(\theta) \log \frac{q^{(i-1)}(\theta) p(\vect{y}_{b_i}|\theta)}{q(\theta)} .\nonumber
\end{align}
These have the form of standard variational inference with the prior replaced by the previous variational distribution $q^{(i-1)}(\theta)$. This idea --  combining the likelihood from a new batch of data with the previous approximate posterior and projecting back to a new approximate posterior -- underpins online variational inference \citep{ghahramani:2000,sato:2001}, streaming variational inference \citep{broderick+al:2013,bui+al:2017b}, and variational continual learning \citep{nguyen+al:2018}. Early work on online VI used conjugate models and analytic updates \citep{ghahramani:2000,sato:2001,broderick+al:2013,bui+al:2017b}, this was followed by off-the-shelf optimization approaches for non-conjugate models \citep{bui+al:2017b} and further extended to leverage MC approximations of the local-free energy \citep{nguyen+al:2018}. Recently \citet{zeno+al:2018} use the variational continual learning framework of \citet{nguyen+al:2018}, but employ fixed-point updates instead.

\subsection{Power EP as a Fully Local VI Fixed Point Method} 
\label{sec:pep}

There is also an important relationship between PVI methods employing fixed point updates and power expectation propagation \citep{minka:2004}. Property \ref{prop:pep-fp} below states that the local VI fixed point equations are recovered from the Power EP algorithm as $\alpha \rightarrow 0$. 
\begin{property}
\label{prop:pep-fp}
The damped fixed point equations are precisely those returned by the PEP algorithm, shown in \cref{alg:pep}, in the limit that $\alpha \rightarrow 0$.
\end{property}
\begin{algorithm}[tb]
   \caption{One step of the PEP algorithm at the $i$-th iteration, for the $b_i$-th data partition}
   \label{alg:pep}
\begin{algorithmic}
    \STATE Compute the tilted distribution:
    $\hat{p}^{(i)}_{\alpha}(\theta) =  q^{(i-1)}(\theta) \left ( \frac{p(\mathbf{y}_{b_i} | \theta)}{t_{b_i}^{(i-1)}(\theta)}\right)^{\alpha}$
    \STATE Moment match:
    $q_{\alpha}(\theta) = \mathrm{proj}( \hat{p}^{(i)}_{\alpha}(\theta) ) \; \text{such that} \; \mathbb{E}_{q(\theta)} \left(T(\theta)\right) = \mathbb{E}_{\hat{p}^{(i)}_{\alpha}(\theta)} \left(T(\theta)\right)$
    \STATE Update the posterior distribution with damping $\rho$:
    $q^{(i)}(\theta) = \left ( q^{(i-1)}(\theta) \right )^{1-\rho/\alpha} \left( q_{\alpha}(\theta) \right)^{\rho/\alpha}$
    \STATE Update the approximate likelihood:
    $t^{(i)}_{b_i}(\theta) = \frac{q^{(i)}(\theta)}{q^{(i-1)}(\theta)} t^{(i-1)}_{b_i}(\theta)$ 
\end{algorithmic}
\end{algorithm}
Although we suspect \citet{knowles+minka:2011} knew of this relationship, and it is well known that Power EP has the same fixed points as VI in this case, it does not appear to be widely known that variationally limited Power EP yields exactly the same algorithm as fixed point local VI. See \ref{proof:pep-fp} for the proof.
 
\subsection{Alpha-divergence EP as a Local VI Method with Off-the-shelf Optimization}
\label{sec:alpha_ep}

PVI is intimately related to alpha-divergence EP. If PVI's KL divergence is replaced by an alpha divergence $ \mathrm{D}_{\alpha} [p(\theta) || q(\theta) ] = \frac{1}{\alpha(1-\alpha)} \int \left [ \alpha p(\theta) + (1-\alpha) q(\theta) -  p(\theta)^{\alpha}  q(\theta)^{1-\alpha} \right] \mathrm{d}\theta$ we recover the alpha-divergence formulation of the power-EP algorithm \citep{minka:2004} which encompasses the current case as $\alpha \rightarrow 0$ and EP when $\alpha \rightarrow 1$ \citep{minka:2001}. The updates using this formulation are shown in \cref{alg:pep_alpha}.
\begin{algorithm}[tb]
   \caption{One step of the PEP algorithm, as in \cref{alg:pep}, but with alpha divergence minimization}
   \label{alg:pep_alpha}
\begin{algorithmic}
    \STATE Compute the tilted distribution:
    $\hat{p}^{(i)}(\theta) =  q^{(i-1)}(\theta)  \frac{p(\mathbf{y}_{b_i} | \theta)}{t_{b_i}^{(i-1)}(\theta)}$
    \STATE Find the posterior distribution:
    $q^{(i)}(\theta) \defeq \argmin_{q(\theta) \in \mathcal{Q}} \mathrm{D}_{\alpha} [\hat{p}^{(i)}(\theta) || q(\theta) ]$
    \STATE Update the approximate likelihood:
    $t^{(i)}_{b_i}(\theta) = \frac{q^{(i)}(\theta)}{q^{(i-1)}(\theta)} t^{(i-1)}_{b_i}(\theta)$ 
\end{algorithmic}
\end{algorithm}
The alpha divergence is typically very difficult to compute once more than one non-Gaussian likelihood is included in a data group $\mathbf{y}_m$, meaning that for general alpha it would be appropriate to set $M=N$. The variational KL is the exception as it decomposes over data points. 

\subsection{Stochastic Power EP as a Stochastic Global VI Fixed Point Method}
\label{sec:stochastic_pep}

The stochastic power EP algorithm \citep{li+al:2015} reduces the memory overhead of EP by maintaining a single likelihood approximation that approximates the average effect a likelihood has on the posterior  $q(\theta) = p(\theta) t(\theta)^M $. Taking the variational limit of this algorithm, $\alpha \rightarrow 0$,  we recover global VI $M=1$ with damped simplified fixed-point updates that employ a stochastic (mini-batch) approximation \citep{hoffman+al:2013}.
\begin{property}
\label{prop:sep}
The mini-batch fixed point equations are precisely those returned by the SPEP algorithm, shown in \cref{alg:spep} in the limit that $\alpha \rightarrow 0$. 
\end{property}
\begin{algorithm}[tb]
   \caption{One step of the SPEP algorithm at the $i$-th iteration, for the $b_i$-th data partition}
   \label{alg:spep}
\begin{algorithmic}
    \STATE Compute the tilted distribution:
    $\hat{p}^{(i)}_{\alpha}(\theta) =  q^{(i-1)}(\theta) \left ( \frac{p(\mathbf{y}_{b_i} | \theta)}{t^{(i-1)}(\theta)}\right)^{\alpha}$
    \STATE Moment match:
    $q_{\alpha}(\theta) = \mathrm{proj}( \hat{p}^{(i)}_{\alpha}(\theta) )\; \text{such that} \; \mathbb{E}_{q(\theta)} \left(T(\theta)\right) = \mathbb{E}_{\hat{p}^{(i)}_{\alpha}(\theta)} \left(T(\theta)\right)$
    \STATE Update the posterior distribution with damping $\rho$:
    $q^{(i)}(\theta) = \left ( q^{(i-1)}(\theta) \right )^{1-N \rho /\alpha} \left( q_{\alpha}(\theta) \right)^{N \rho/\alpha}$
    \STATE Update the approximate likelihood:
    $t^{(i)} = \left (\frac{q^{(i)}(\theta)}{p(\theta)} \right)^{1/N}$ 
\end{algorithmic}
\end{algorithm}
In this way the relationship between EP and SEP is the same as the relationship between fixed point PVI and fixed point mini-batch global VI (see section \ref{section:mini-batch} where the two approaches differ by removing either an average natural parameter or a specific one). Similarly, if we altered PVI to maintain a single average likelihood approximation, as SEP does, we would recover mini-batch global VI.

\subsection{Distributed (Power) EP Methods} 
\label{sec:distributed_pep}

The convergent distributed Power EP approach of \cite{hasenclever+al:2017} recovers a version of PVI as $\alpha \rightarrow 0$ with convergence guarantees. The PVI approach is also similar in spirit to \citet{gelman+al:2014,hasenclever+al:2017} who use EP to split up data sets into small parts that are amenable to MCMC. Here we are using PVI to split up data sets so that they are amenable for optimization.

\afterpage{
\begin{landscape}
\begin{table*}[ht]
\centering
\begin{tabular}{p{0.24\linewidth}p{0.10\linewidth}p{0.25\linewidth}p{0.24\linewidth}p{0.1\linewidth}}
\hline
\rule{0pt}{1.1em}\rule[-0.5em]{0pt}{1em}\textbf{Reference} & \textbf{Granularity} & \textbf{Optimization}  & \textbf{Models} & \textbf{Name}\\
%
\hline
\multicolumn{3}{l}{\textbf{Global VI} [PVI $M=1$, see \cref{sec:global}]} & & \\
\citet{beal2003variational} & global & analytic & conjugate & VI \\
\citet{sato:2001} & global & analytic & conjugate (MoG) & 
\\
\cite{hinton+vancamp:1993} & global & gradient ascent & non-conjugate (neural network) & 
\\
\citet{honkela+al:2010} & global & natural gradient (mean only)  & non-conj.~(MoG, NSSM, NFA)&  \\
\citet{hensman+al:2012} & global & CG with natural gradient& conjugate & 
\\
\citet{hensman+al:13} & global & stochastic natural gradient & conjugate & 
\\
\citet{hoffman+al:2013} & global &stochastic natural gradient & conjugate & SVI\\
\citet{kucukelbir+al:2017} & global & stochastic gradient descent & non-conjugate & ADVI \\
\citet{salimans+knowles:2013}   & global &  fixed-point + MC + stochastic& non-conjugate (PR, BB, SV) & 
\\
\citet{sheth+al:2015} & global & simplified fixed point & non-conjugate (GLV) &  \\
\citet{sheth+khardon:2016} & global & simplified fixed point & non-conjugate (GLV) &  \\
\cite{sheth+khardon:2016b} & global & simplified fixed point + MC & non-conjugate (two level) & \\
\hline
%
\multicolumn{3}{l}{\textbf{Fully Local VI} [PVI $M=N$, see \cref{sec:local}]} & & \\
\citet{winn+bishop:2005} & fully local & analytic & conjugate (GM)  & VMP\\
\citet{archambeau+ermis:2015} & fully local  & incremental & conjugate (LDA) & IVI\\
\citet{knowles+minka:2011} & fully local  &  fixed-point & non-conjugate (LR, MR) & NC-VMP\\
\citet{wand:2014} & fully local  &simplified fixed-point & non-conjugate (PMM, HLR)  & SNC-VMP \\
\citet{khan+li:2018}  & local & damped stochastic simplified fixed-point & non-conjugate (LR, GFM, PGMF) & CCVI \\
\hline
%
\multicolumn{3}{l}{\textbf{Online VI} [one pass of PVI, see \cref{sec:online}]} && \\
\citet{ghahramani:2000} & fully local  & analytic & conjugate (MoG)  &\\
\citet{sato:2001} & fully local & analytic & conjugate (MoG) & online VB \\
\citet{broderick+al:2013} & fully local & analytic & conjugate (LDA) & streaming VI \\
\citet{bui+al:2017} & fully local &analytic/LBFGS & conjugate and not (SGPs) &  \\ 
\citet{nguyen+al:2018}  & fully local &Adam & non-conjugate (BNN) & VCL\\
\citet{zeno+al:2018} & fully local & fixed-point & non-conjugate (BNN) & BGD\\
\hline
%
\multicolumn{3}{l}{\textbf{Power EP} [PVI when $\alpha \rightarrow 0$, see \cref{sec:pep,sec:alpha_ep,sec:stochastic_pep,sec:distributed_pep}]} & & \\
\citet{minka:2004}  &  local& series fixed point & non-conjugate (GM) & PEP  \\
\citet{minka:2004}  &  local& optimization &  & ADEP  \\
\citet{bui+al:2017b}  & local & analytic/fixed-point & conjugate / non-conj.~(GPs) & PEP \\
\citet{hasenclever+al:2017} &  local & analytic with MC & non-conjugate (BNN) & CPEP \\
\citet{li+al:2015}  & local & stochastic fixed point & non-conjugate (LR, BNN) & SPEP \\
\hline
\end{tabular}
\setcounter{table}{0}
\caption{Variational inference schemes encompassed by the PVI framework. See previous page for full caption.}
\end{table*}
\end{landscape}}
 
   \begin{table}[!ht]
\setcounter{table}{0}
        \caption[]{Variational inference schemes encompassed by the PVI framework. (See next page.) 
        Selected past work has been organized into four categories: global VI (PVI with $M=1$), fully local PVI ($M=N$), Power EP variants, and online VI. The citation to the work is provided along with the granularity of the method (global indicates $M=1$, fully local $M=N$, local implies general $M$ can be used). 
        The optimization used from the PVI perspective on this work is noted.  Abbreviations used here are: Conjugate Gradient (CG) and Monte Carlo (MC). The model class that the scheme encompasses is noted (conjugate versus non-conjugate) along with the specific models that the scheme was tested on. Model abbreviations are: Non-linear State-space Model (NSSM), Non-linear Factor Analysis (NFA), Latent Dirichlet Allocation (LDA),  Poisson Mixed Model (PMM), Heteroscedastic Linear Regression (HLR),  Sparse Gaussian Processes (SGPs), Graphical Model (GM),  Logistic Regression (LR),  Beta-binomial (BB),  Stochastic Volatility model (SV),  Probit Regression (PR),  Multinomial Regression (MR),  Bayesian Neural Network (BNN),  Gamma factor model (GFM),  Poisson Gamma Matrix Factorization (PGMF),  Mixture of Gaussians (MoG).   Poisson Mixed Model (PMM),  Heteroscedastic Linear Regression (HLR),  Gaussian Latent Variable (GLV).
        If the scheme proposed by the method has a name, this is noted in the final column. Abbreviations of the inference scheme are:  Automatic Differentiation VI (ADVI), Incremental VI (IVI),  Non-conjugate Variational Message Passing (NC-VMP),  Simplified NC-VMP (SNC-VMP),  Conjugate-Computation VI (CCVI),   Power EP (PEP),  Alpha-divergence PEP (ADPEP), Convergent Power EP (CPEP),  Stochastic Power EP (SPEP), Variational Continual Learning (VCL),  Bayesian Gradient Descent (BGD).  }
        \label{table:past-work}
    \end{table}

\section{Improving Federated Learning for Bayesian Neural Networks Using PVI}
\label{sec:federated}
Having connected the previous literature using the unifying framework based on PVI, we will discuss how PVI enables novel and practical algorithms to emerge. In this section, we detail how Partitioned Variational Inference can be used for federated approximate training of Bayesian neural networks, allowing both synchronous or lock-free asynchronous model updates across many machines.\footnote{Neural network models are used throughout this section and the experiment, but other models can be employed in the same fashion as the training framework developed here is general.} As a running example, consider a multiclass classification problem with $C$ classes and assume that the training points are partitioned into $K$ disjoint memory shards (or subsets). In practical federated settings, the allocation of data points to shards is often unknown a priori, for example, the number of data points across various shards may be unbalanced or some classes may be present only on a few memory shards, or on one in the extreme. We first discuss Bayesian neural networks and a global variational approximation for training BNNs, and detail how this approximation can be used at the shard level.

Consider a neural network that models the distribution of a target $y$ given an input $\xvec$, $p(y|\theta, \xvec)$, where $\theta$ include the weights and biases in the network. To complete the model, we assign a prior $p(\theta)$ over the unknown parameters $\theta$. Having specified the probability of everything, we turn the handle of probability theory to obtain the posterior distribution,
\begin{align}
    p(\theta|\xvec, \yvec) = \frac{p(\theta) p(\yvec|\theta, \xvec)}{p(\yvec|\xvec)} =  \frac{p(\theta) \prod_{k=1}^{K} \prod_{n=1}^{N_k} p(y_{k,n}|\theta, \xvec_{k,n})}{p(\yvec|\xvec)}.
\end{align}
The exact posterior above is analytically intractable and thus approximation techniques such as sampling or deterministic methods are needed. There is a long history of research on approximate Bayesian training of neural networks, including extended Kalman filtering \citep{singhal+wu:89}, Laplace's approximation \citep{mackay:2003}, Hamiltonian Monte Carlo \citep{neal:92,neal:12}, variational inference \citep{hinton+vancamp:1993,barber+bishop:98,grave:11,blundell+al:15,gal+ghahramani:16}, sequential Monte Carlo \citep{freitas+al:00}, expectation propagation \citep{hernandez-lobato+adams:15}, and approximate power EP \citep{li+al:2015,hernandez-lobato+li+al:16}. In this section, we focus on Monte Carlo variational inference methods with a mean-field Gaussian variational approximation \citep{grave:11,blundell+al:15}. In detail, a factorized global variational approximation, $q(\theta) = \prod_i \norm(\theta_i; \mu_i, \sigma_i^2)$, is used to lower-bound the log marginal likelihood as follows,
\begin{align}
    \log p(\yvec|\xvec) = \log \int \dd \theta \; p(\theta) p(\yvec|\theta, \xvec) \geq \int \dd \theta \; q(\theta) \log \frac{p(\theta) p(\yvec|\theta, \xvec)}{q(\theta)} = \mathcal{F}_{\mathrm{GVI}}(q(\theta)),
\end{align}
where $\mathcal{F}_{\mathrm{GVI}}(q(\theta))$ is the variational lower bound or the negative variational free-energy. This bound can be expanded as follows,
\begin{align}
    \mathcal{F}_{\mathrm{GVI}}(q(\theta)) = -\mathrm{KL}[q(\theta) || p(\theta)] + \sum_{k=1}^{K} \sum_{n=1}^{N_k} \int \dd \theta \; q(\theta) \log p(y_{k,n}|\theta, \xvec_{k,n}).
    \label{eqn:bnn_gvi_vfe}
\end{align}
When the prior is chosen to be a Gaussian, the KL term in the bound above can be computed analytically. In contrast, the expected log-likelihood term is not analytically tractable. However, it can be approximated by simple Monte Carlo with the (local) reparameterization trick such that low-variance stochastic gradients of the approximate expectation wrt the variational parameters $\{\mu_i, \sigma_i\}$ can be easily obtained \citep{rezende+al:2014,kingma+welling:2014,kingma+al:15}. 

The variational lower-bound above can be optimized using any off-the-shelf stochastic optimizer, and its gradient computation can be trivially distributed across many machines. A possible synchronously distributed schedule when using K compute nodes, each having access to a memory shard, is as follows: (i) a central compute node passes the current $q(\theta)$ to K workers, (ii) each worker then computes the gradients of the expected log-likelihood of (a mini-batch of) its own data and passes the gradients back to the central node, (iii) the central node aggregates these gradients, combines the result with the gradient of the KL term, and performs an optimization step to obtain a new $q(\theta)$. These steps are then repeated for a fixed number of iterations or until convergence. However, notice that this schedule is communication-inefficient, as it requires frequent communication of the gradients and the updated variational approximation between the central node and the K compute workers. We will next discuss an inference scheme based on PVI that allows communication efficient updates between workers that is compatible with various scheduling schemes.

Following the PVI formulation in \cref{sec:PVI}, the approximate posterior can be rewritten using the approximate factors, one for each memory shard, as follows,
\begin{align}
    p(\theta|\xvec, \yvec) \propto p(\theta) \prod_{k=1}^{K} \prod_{n=1}^{N_k} p(y_{k,n}|\theta, \xvec_{k,n}) \approx p(\theta) \prod_{k=1}^{K} t_k(\theta) = q(\theta),
\end{align}
where $t_k(\theta)$ approximates the contribution of data points in the k-th shard to the posterior. As discussed in the previous sections, PVI turns the original global approximate inference task into a collection of approximate inference tasks, i.e.~for the k-th memory shard and k-th compute node, the task is to maximize,
\begin{align}
    \mathcal{F}^{k}_{\mathrm{PVI}}(q(\theta)) = -\mathrm{KL}[q(\theta) || q^{\setminus k}(\theta)] + \sum_{n=1}^{N_k} \int \dd \theta \; q(\theta) \log p(y_{k,n}|\theta, \xvec_{k,n}),
    \label{eqn:bnn_pvi_vfe}
\end{align}
where $q^{\setminus k}(\theta) = q(\theta) / t_k(\theta)$ is the context or effective prior set by data points in other shards. Once a new variational approximation $q(\theta)$ is obtained, a new approximate factor can be computed accordingly, $t_k(\theta) = q(\theta) / q^{\setminus k}(\theta)$. Note that the objective for each compute node is almost identical to the GVI objective, except the prior is now replaced by the context and the data are limited to the compute node's accessible data. This means any global VI implementation available on a compute node (either using optimization, fixed-point updates, or in close-formed) can be trivially modified to handle PVI. A key additional difference to GVI is the communication frequency between the compute nodes and the central parameter server (that holds the latest $q(\theta)$): a worker can decide to pass $t_k(\theta)$ back to the central server after multiple passes through its data, after one epoch, or after just one mini-batch. This leaves room for practitioners to choose a learning schedule that meets communication constraints. More importantly, PVI enables various communication strategies to be deployed, for example:
\begin{itemize}
    \item Sequential PVI with only one pass through the data set: each worker, in turn, runs Global VI, with the previous posterior being the prior/context, for the data points in its memory shard and returns the posterior approximation to the parameter server. This posterior approximation will then be used as the context for the next worker's execution. Note that this is exactly equivalent to Variational Continual Learning \citep{nguyen+al:2018} and can be combined with the {\it multihead} architecture, each head handling one task or one worker, or with episodic memory \citep[see e.g.][]{zenke+al:17,nguyen+al:2018}. This strategy is communication-efficient as only a small number of messages are required --- only one up/down update is needed for each worker.
    \item PVI with synchronous model updates: instead of sequentially updating the context distribution and running only one worker at a time, all workers can be run in parallel. That is, each worker occasionally sends its updated contribution to the posterior back to the parameter server. The parameter server waits for all workers to finish before aggregating the approximate factors and sending the new posterior back to the workers. The workers will then update their own context distributions based on the current state of the central parameters. This process then repeats. By analyzing the homogeneity of the data and updates across workers, heuristics could be used to choose the learning rate for each worker and damping factor for the central parameter server --- we leave this for future work.
    \item PVI with lock-free asynchronous updates: instead of waiting for all workers to finish training locally, the model aggregation and update steps can be performed as soon as any worker has finished. This strategy is particularly useful when communication is done over an unreliable channel, the distribution of the data across different machines is highly unbalanced, or when a machine can be disconnected from the training procedure at any time. However, this strategy is expected to be generally worse compared the synchronous update scheme above, since the context/cavity distribution could be changed while a worker is running and the next parameter update performed by this worker could overwrite the updates made by other workers, i.e.~there is the possibility of stale updates.
\end{itemize}

We demonstrate these communication strategies on a large-scale federated classification task in \cref{sec:exp_federated} and highlight the advantages and potential pitfalls of PVI, GVI and various alternatives for different levels of data homogeneity across memory shards.

\section{Improving Continual Learning for Sparse Gaussian Processes Using PVI}
\label{sec:continual}
Gaussian processes (GPs) are flexible probabilistic distributions over functions that have been used in wide variety of machine learning problems, including supervised learning \citep{rasmussen+williams:05}, unsupervised learning \citep{lawrence:04} and reinforcement learning \citep{deisenroth:10}. 
The application of GPs to more general, large-scale settings is however hindered by analytical and computational intractabilities. 
As a result, a large body of active GP research aims to develop efficient approximation strategies for inference and learning in GP models. 
In this work, we develop an approximation based on partitioned variational inference for GP regression and classification in a continual learning setting. 
In this setting, data arrive sequentially, either one data point at a time or in batches of a size that is unknown a priori.
An efficient strategy to accurately update the model in an online fashion is thus needed and can be used for various applications such as control \citep{nguyen-tuong+al:09} or mapping \citep{ocallaghan+al:12}.

In particular, building on recent work on pseudo-point sparse approximations \citep{titsias:2009a,hensman+al:15,matthews+al:16,bui+al:2017b} and streaming approximations \citep{csato+opper:2002,bui+al:2017}, we develop a streaming variational approximation that approximates the posterior distribution over both the GP latent function {\it and} the hyperparameters for GP regression and classification models. Additionally, the partitioned VI view of this approximation allows just-in-time, dynamic allocation of new pseudo-points specific to a data batch, and more efficient training time and accurate predictions in practice. We will provide a concise review of sparse approximations for Gaussian process regression and classification before summarizing the proposed continual learning approach. For interested readers, see \cite{quinonero+rasmussen:05,bui+al:2017} for more comprehensive reviews of sparse GPs. \Cref{sec:appen:sgp} contains the full derivation of different streaming variational approaches with shared or private pseudo points, and with maximum likelihood or variational learning strategies for the hyperparameters.

\subsection{Variational inference for both latent function and hyperparameters}

Given $N$ input and output pairs $\{\xvec_n, y_n\}_{n=1}^{N}$, a standard GP regression or classification model assumes the outputs $\{y_n\}_{n=1}^{N}$ are generated from the inputs $\{\xvec_n\}_{n=1}^{N}$ according to $y_n = f(\xvec_n) + \xi_n$, where $f$ is an unknown function that is corrupted by observation noise, for example, $\xi \sim \norm (0, \sigma_y^2)$ in the real-valued output regression problem.\footnote{In this section, $f$ stands for the model parameters, as denoted by $\theta$ in the previous sections.} Typically, $f$ is assumed to be drawn from a zero-mean GP prior $f \sim \mathcal{GP}(\mathbf{0}, k(\cdot, \cdot|\epsilon))$ whose covariance function depends on hyperparameters $\epsilon$. We also place a prior over the hyperparameters $\epsilon$ and as such inference involves finding the posterior over both $f$ and $\epsilon$, $p(f, \epsilon| \yvec, \xvec)$, and computing the marginal likelihood $p(\yvec|\xvec)$, where we have collected the inputs and observations into vectors $\xvec= \{\xvec_n\}_{n=1}^N$ and $\yvec = \{ y_n \}_{n=1}^N$ respectively. This is one key difference to the work of \cite{bui+al:2017b}, in which only a point estimate of the hyperparameters is learned via maximum likelihood. The dependence on the inputs of the posterior, marginal likelihood, and other quantities will be suppressed when appropriate to lighten the notation. Exact inference in the model considered here is analytically and computationally intractable, due to the non-linear dependency between $f$ and $\epsilon$, and the need to perform a high dimensional integration when $N$ is large. 

In this work, we focus on the variational free energy approximation scheme \citep{titsias:2009a,matthews+al:16} which is arguably the leading approximation method for many scenarios. This scheme lower bounds the marginal likelihood of the model using a variational distribution $q(f,\epsilon)$ over the latent function and the hyperparameters:
\begin{align}
\log p(\yvec|\xvec) = \log \int \dd f \dd \epsilon\; p(\yvec, f, \epsilon|\xvec) \geq \int \dd f \dd \epsilon\; q(f, \epsilon) \log \frac{p(\yvec|f,\epsilon,\xvec) p(f|\epsilon) p(\epsilon)}{q(f,\epsilon)} = \mathcal{F}(q(f, \epsilon)),\nonumber
\end{align}
where $\mathcal{F}(q(f,\epsilon))$ is the variational surrogate objective and can be maximized to obtain $q(f,\epsilon)$. In order to arrive at a computationally tractable method, the approximate posterior is parameterized via a set of $M_\avec$ pseudo-outputs $\avec$ which are a subset of the function values $f = \{f_{\neq \avec}, \avec \}$. Specifically, the approximate posterior takes the following structure:
\begin{align}
    q(f,\epsilon) = p(f_{\neq \avec}|\avec, \epsilon) q(\avec) q(\epsilon), \label{eqn:gp_global_q}
\end{align}
where $q(\avec)$ and $q(\epsilon)$ are variational distributions over $\avec$ and $\epsilon$ respectively, and $p(f_{\neq \avec}|\avec, \epsilon)$ is the conditional prior distribution of the remaining latent function values. Note that while $\avec$ and $\epsilon$ are assumed to be factorized in the approximate posterior, the dependencies between the remaining latent function values $f_{\neq \avec}$ and the hyperparameters $\epsilon$, and between $f_{\neq \avec}$ themselves are retained due to the conditional prior. This assumption leads to a critical cancellation that results in a computationally tractable lower bound as follows:
\begin{align}
\mathcal{F}(q(\avec), q(\epsilon)) 
	&= \int \dd f \dd \epsilon \; q(f) \log \frac{p(\yvec|f, \epsilon,\xvec) p(\epsilon) p(\avec | \epsilon) ) \bcancel{p (f_{\neq \avec}|\avec, \epsilon)}}{\bcancel{p(f_{\neq \avec}|\avec, \epsilon)} q(\avec) q(\epsilon)} \nonumber \\
	&= - \mathrm{KL}[q(\epsilon)||p(\epsilon)] - \int \dd \epsilon \; q(\epsilon) \mathrm{KL}[q(\avec)||p(\avec|\epsilon)] \nonumber\\&\quad\quad + \sum_{n} \int \dd \epsilon\; \dd \avec\;\dd f_n q(\epsilon) q(\avec) p(f_n|\avec, \epsilon) \log p(y_n|f_n, \epsilon, \xvec_n), \nonumber
\end{align}
where $f_n = f(\xvec_n)$ is the latent function value at $\xvec_n$. Most terms in the variational lower bound above require computation of an expectation wrt the variational approximation $q(\epsilon)$, which is not available in closed-form even when $q(\epsilon)$ takes a simple form such as a diagonal Gaussian. However, these expectations can be approximated by simple Monte Carlo with the {\it reparameterization trick} \citep{kingma+welling:2014,rezende+al:2014}. The remaining expectations can be handled tractably, either in closed-form or by using Gaussian quadrature.

\subsection{Continual learning for streaming data with private pseudo-points}
In continual learning, data arrive sequentially and revisiting previously seen data is prohibitive, and the current variational posterior can be reused to approximate the effect of the previous data on the exact posterior. Let $\{\xvec_1,\yvec_1\}$ and $\{\xvec_2,\yvec_2\}$ denote previous and new data, respectively. We first re-interpret the structured global approximate posterior in \cref{eqn:gp_global_q} as a product of local approximate factors as follows,
\begin{align*}
    p(f,\epsilon|\yvec_1,\xvec_1) & = p(\epsilon) p(f|\epsilon) p(\yvec_1|f, \epsilon, \xvec_1) / \mathcal{Z}_1 \\ &= p(\epsilon) p(f_{\neq \avec}|\avec, \epsilon) \textcolor{red}{ p(\avec | \epsilon) } \textcolor{blue}{ p(\yvec_1|f, \epsilon, \xvec_1) }  / \mathcal{Z}_1 \\ &\approx p(\epsilon) p(f_{\neq \avec}|\avec, \epsilon) \textcolor{red}{ t_1(\avec) t_1(\epsilon) } \textcolor{blue}{ g_1(\avec)  g_1(\epsilon)}\\ &= p(f_{\neq \avec}|\avec, \epsilon) q_1(\avec) q_1(\epsilon),
\end{align*}
where $q_1(\avec) = t_1(\avec) g_1(\avec)$, $q_1(\epsilon) = p(\epsilon) t_1(\epsilon) g_1(\epsilon)$, and $t_1(\cdot)$s and $g_1(\cdot)$s are introduced to approximate the contribution of $p(\avec | \epsilon)$ and $p(\yvec_1|f, \epsilon, \xvec_1)$ to the posterior, respectively. Note that the last equation is identical to \cref{eqn:gp_global_q}, but the factor representation above facilitates inference using PVI and allows more flexible approximations in the streaming setting. In particular, the exact posterior when both old data $\yvec_1$ and newly arrived data $\yvec_2$ are included can be approximated in a similar fashion,
\begin{align*}
    p(f,\epsilon|\yvec_1,\yvec_2,\xvec_1,\xvec_2) & = p(\epsilon) p(f|\epsilon) p(\yvec_1|f, \epsilon, \xvec_1)  p(\yvec_2|f, \epsilon, \xvec_2) / \mathcal{Z}_{12} \\ &= p(\epsilon) p(f_{\neq \avec,\bvec}|\avec, \bvec, \epsilon) \textcolor{DarkOrchid}{ p(\bvec | \avec, \epsilon) } \textcolor{red}{ p(\avec | \epsilon) } \textcolor{blue}{ p(\yvec_1|f, \epsilon, \xvec_1) } \textcolor{ForestGreen}{ p(\yvec_2|f, \epsilon, \xvec_2) } / \mathcal{Z}_{12} \\ &\approx p(\epsilon) p(f_{\neq \avec,\bvec}|\avec, \bvec, \epsilon) \textcolor{DarkOrchid}{ t_2(\bvec | \avec) t_2(\epsilon) } \textcolor{red}{ t_1(\avec) t_1(\epsilon) } \textcolor{blue}{ g_1(\avec)  g_1(\epsilon)} \textcolor{ForestGreen}{ g_2(\bvec)  g_2(\epsilon)}
\end{align*}
where $\bvec$ are new pseudo-outputs, and $t_2(\cdot)$s and $g_2(\cdot)$s are approximate contributions of $p(\bvec | \avec, \epsilon)$ and $p(\yvec_2|f, \epsilon, \xvec_2)$ to the posterior, respectively. As we have reused the approximate factors $t_1(\avec)$ and $g_1(\avec)$, the newly introduced pseudo-points $\bvec$ can be thought of as pseudo-points private to the new data. This is the key difference of this work compared to the approach of \cite{bui+al:2017}, in which both old and new data share a same set of pseudo-points. The advantages of the approach based on private pseudo-points are potentially two-fold: (i) it is conceptually simpler to focus the approximation effort to handle the new data points while keeping the approximation for previous data fixed, as a new data batch may require only a small number of representative pseudo-points, and (ii) the number of parameters (variational parameters and private pseudo-inputs) is much smaller, leading to arguably easier problem to initialize and optimize.

The approximate factors $t_2(\cdot)$ and $g_2(\cdot)$ can be found by employing the PVI algorithm in \cref{sec:PVI}. Alternatively, in the continual learning setting where data points do not need to be revisited, we can convert the factor-based variational approximation above to a global variational approximation,
\begin{align*}
    p(f,\epsilon|\yvec_1,\yvec_2,\xvec_1,\xvec_2) & \approx p(\epsilon) p(f_{\neq \avec,\bvec}|\avec, \bvec, \epsilon) \textcolor{DarkOrchid}{ t_2(\bvec | \avec) t_2(\epsilon) } \textcolor{red}{ t_1(\avec) t_1(\epsilon) } \textcolor{blue}{ g_1(\avec)  g_1(\epsilon)} \textcolor{ForestGreen}{ g_2(\bvec)  g_2(\epsilon)} \\ &= p(f_{\neq \avec,\bvec}|\avec, \bvec, \epsilon) q_2(\bvec|\avec) q_1(\avec) q_2(\epsilon)
\end{align*}
where $q_2(\bvec|\avec) = t_2(\bvec|\avec) g_2(\bvec)$, $q_2(\epsilon) = p(\epsilon) t_1(\epsilon) g_1(\epsilon) t_2(\epsilon) g_2(\epsilon)$, and $q_2(\bvec|\avec)$ and $q_2(\epsilon)$ are parameterized and optimized, along with the location of $\bvec$. While this does not change the fixed-point solution compared to the PVI algorithm, it allows existing sparse global VI implementations such as that in GPflow \citep{matthews+al:17} to be easily extended and deployed.

\section{Experiments}
\label{sec:experiments}
Having discussed the connections to the literature and developed two novel applications of PVI, we validate the proposed methods by running a suite of continual and federated learning experiments on Bayesian neural network and Gaussian process models.   

\subsection{Distributed Federated Variational Inference for Neural Networks}
\label{sec:exp_federated}

In this section, we demonstrate that Partitioned Variational Inference is well-suited for federated approximate training of Bayesian neural networks, allowing both synchronous or lock-free asynchronous model updates across many machines. In particular, we consider the MNIST ten-class classification problem and assume that the training points are partitioned into $K$ disjoint shards. Two levels of data homogeneity across memory shards are considered: homogeneous [or iid, e.g.~each shard has training points of all classes] and inhomogeneous [or non-iid, e.g.~when $K=10$, each shard has training points of only one class]. We evaluate different training methods using a Bayesian neural network with one hidden layer of 200 rectified linear units. We place a diagonal standard Normal prior over the parameters, $p(\theta) = \norm(\theta; 0, \mathrm{I})$, and initialize the mean of the variational approximations as suggested by \cite{glorot+bengio:17}. For distributed training methods, the data set is partitioned into 10 subsets or shards ($K=10$), and 10 compute nodes (workers) with each able to access one memory shard. The implementation of different inference strategies was done in Tensorflow \citep{abadi+al:16} and the communication between workers is managed using Ray \citep{moritz+al:17}. We use Adam \citep{kingma+ba:2014} for the inner loop optimization for partitioned, distributed methods or the outer loop optimization for global VI, and mini-batches of 200 data points. In the next few paragraphs, we briefly detail the methods compared in this section and their results.

\begin{landscape}
\begin{figure}
    \captionsetup{font=footnotesize,labelfont=footnotesize}
    \begin{subfigure}[h]{0.5\linewidth}
        \includegraphics[width=\linewidth]{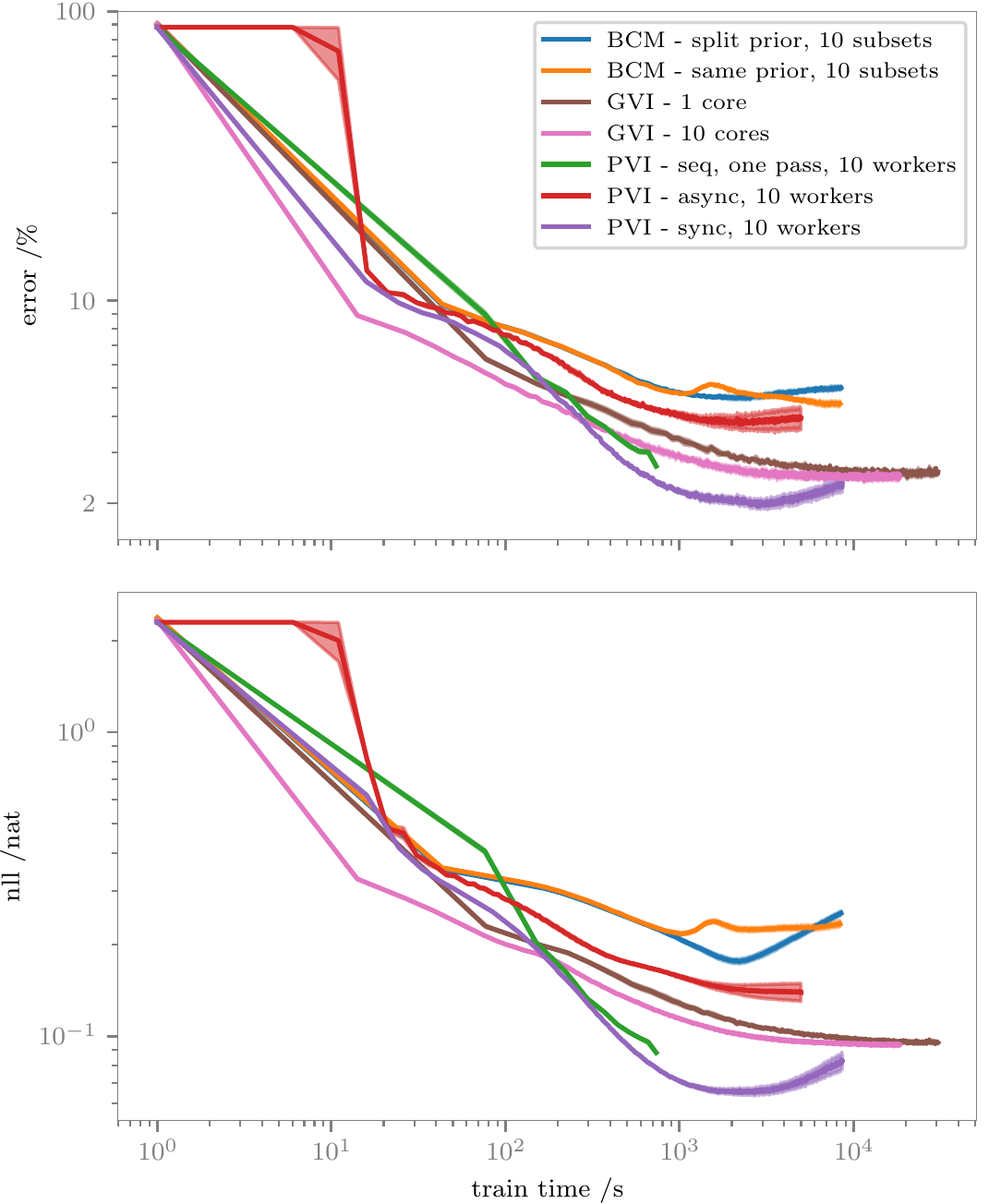}
        \caption{Error and NLL vs train time}
    \end{subfigure}
    \hfill
    \begin{subfigure}[h]{0.5\linewidth}
        \includegraphics[width=\linewidth]{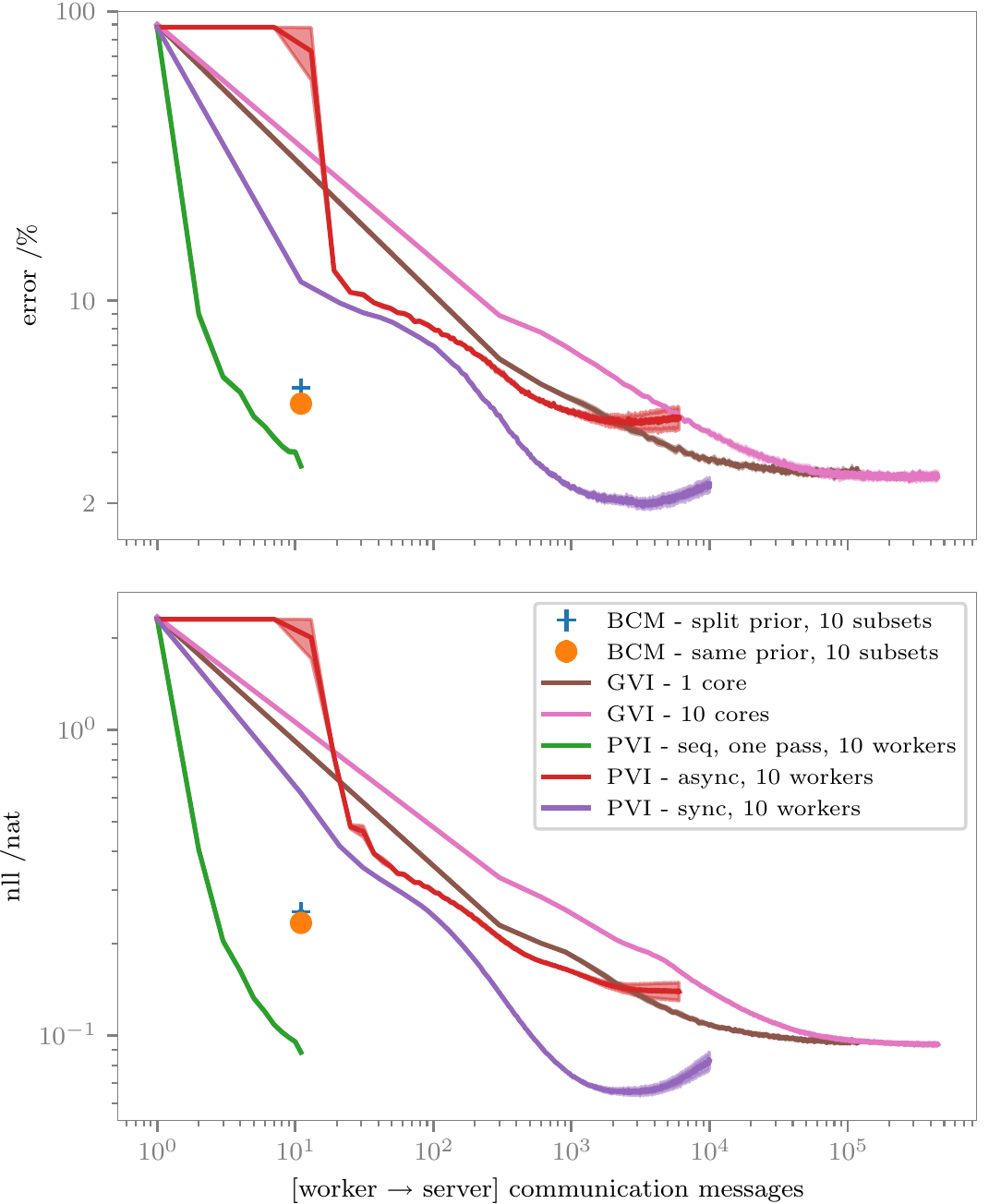}
        \caption{Error and NLL vs communication cost}
    \end{subfigure}%
    \caption{Performance on the test set in the federated MNIST experiment with an iid distribution of training points across ten workers. The test performance is measured using the classification error [error] and the negative log-likelihood [nll], and for both measures, lower is better. All methods are assessed using the performance vs train time and performance vs communication cost plots --- closer to the bottom left of the plots is better. Methods used for benchmarking are: Bayesian Committee Machines (BCM) with the standard Normal prior [same] and with a weakened prior [split], Global VI (GVI) with one and ten compute cores, PVI with sequential updates and only one pass through the data [equivalent to Variational Continual Learning], PVI with lock-free asynchronous updates (PVI - async), and PVI with synchronous updates (PVI - sync). For ease of presentation, the x-axes for the plots start at 1. See text for more details. Best viewed in colour.\label{fig:res_dist_best_iid}}
\end{figure}
\end{landscape}

\begin{landscape}
\begin{figure}
    \captionsetup{font=footnotesize,labelfont=footnotesize}
    \begin{subfigure}[h]{0.5\linewidth}
        \includegraphics[width=\linewidth]{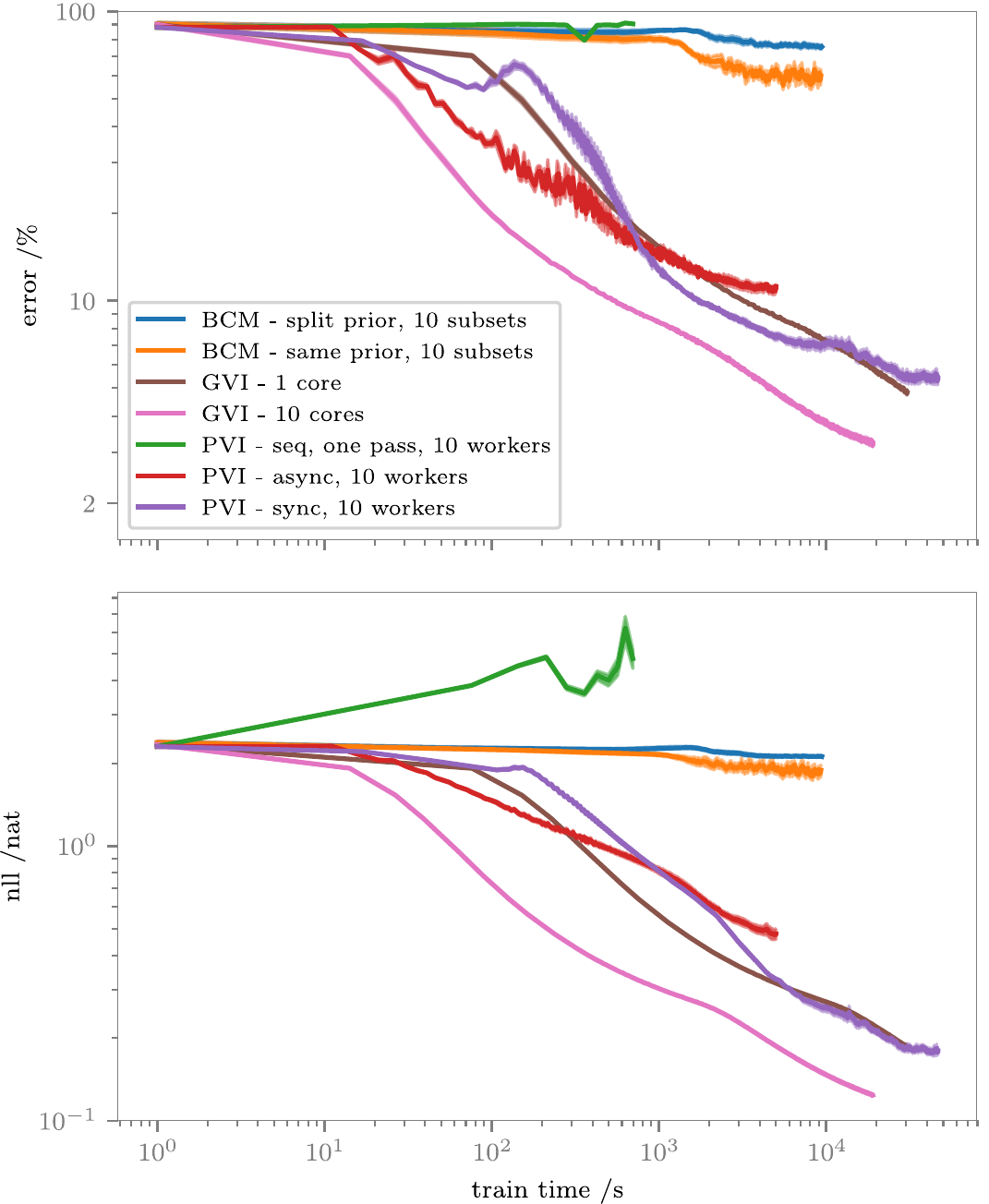}
        \caption{Error and NLL vs train time}
    \end{subfigure}
    \hfill
    \begin{subfigure}[h]{0.5\linewidth}
        \includegraphics[width=\linewidth]{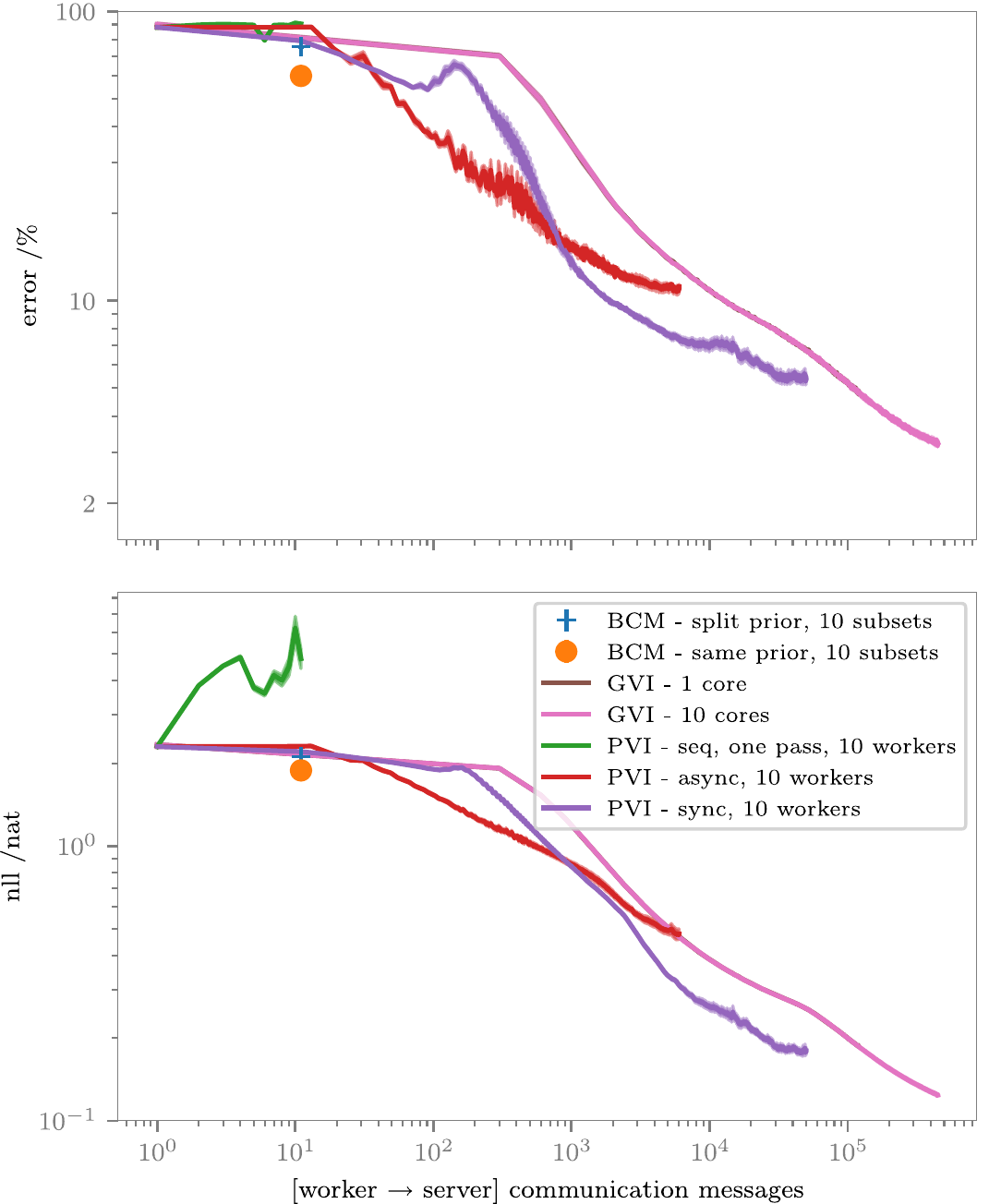}
        \caption{Error and NLL vs communication cost}
    \end{subfigure}%
    \caption{Performance on the test set in the federated MNIST experiment with a non-iid distribution of training points across ten workers, i.e.~each worker has access to digits of only one class. The test performance is measured using the classification error [error] and the negative log-likelihood [nll], and for both measures, lower is better. All methods are assessed using the performance vs train time and performance vs communication cost plots --- closer to the bottom left of the plots is better. Methods used for benchmarking are: Bayesian Committee Machines (BCM) with the standard Normal prior [same] and with a weakened prior [split], Global VI (GVI) with one and ten compute cores, PVI with sequential updates and only one pass through the data [equivalent to Variational Continual Learning], PVI with lock-free asynchronous updates (PVI - async), and PVI with synchronous updates (PVI - sync). For ease of presentation, the x-axes for the plots start at 1. See text for more details. Best viewed in colour.\label{fig:res_dist_best_noniid}}
\end{figure}
\end{landscape}

\paragraph{Global VI}
We first evaluate global VI with a diagonal Gaussian variational approximation for the weights in the neural network. In particular, it is assumed that there is only one compute node (with either one core or ten cores) that can access the entire data set. This compute node maintains a global variational approximation to the exact posterior, and adjusts this approximation using the noisy gradients of the variational free-energy in \cref{eqn:bnn_gvi_vfe}. We simulate the data distribution by sequentially showing mini-batches that can potentially have all ten classes (iid) or that have data of only one class (non-iid). \Cref{fig:app:res_dist_gvi_one,fig:app:res_dist_gvi_ten} in the appendix show the full performance statistics on the test set during training for different learning rates and data homogeneity levels. The performance depends strongly on the learning rate, especially when the mini-batches are non-iid. Faster convergence early in training does not guarantee a better eventual model as measured by test performance, for the iid setting. Note that GVI for the non-iid setting can arrive at a good test error rate, albeit requiring a much smaller learning rate and a substantially larger training time. In addition, global VI is not communication-efficient, as the global parameters are updated as often as data mini-batches are considered. The best performing method for the iid/non-iid settings is selected from all of the learning rates considered and they are shown in \cref{fig:res_dist_best_iid,fig:res_dist_best_noniid}.

\paragraph{Bayesian committee machine} The Bayesian committee machine (BCM) is a simple baseline which is naturally applicable to partitioned data \citep{tresp:00}. The BCM performs (approximate) inference for each data shard independently of other data shards and aggregate the sub-posteriors at the end. In particular, global VI with a diagonal Gaussian variational approximation is applied independently to the data in each shard yielding approximate local posteriors $\{q_k(\theta)\}_{k=1}^{K}$. The aggregation step involves multiplying $K$ Gaussian densities. The only shared information across different members of the committee is the prior. This baseline, therefore, assesses the benefits from coordination between the workers. We consider two prior sharing strategies as follows,
\begin{align*}
&\text{BCM --- same:} \; & p(\theta|\xvec, \yvec) &\propto p(\theta) \prod_{k=1}^{K} p(\yvec_k|\theta, \xvec_k) = \frac{\prod_{k=1}^{K} [p(\theta) p(\yvec_k|\theta, \xvec_k)]}{[p(\theta)]^{K-1}} \approx \frac{\prod_{k=1}^{K} q_k(\theta)}{[p(\theta)]^{K-1}},\\
&\text{BCM --- split:} \; & p(\theta|\xvec, \yvec) &\propto p(\theta) \prod_{k=1}^{K} p(\yvec_k|\theta, \xvec_k) = \prod_{k=1}^{K} [[p(\theta)]^{N_k/N} p(\yvec_k|\theta, \xvec_k)] \approx  \prod_{k=1}^{K} q_k(\theta).
\end{align*} 
BCM is fully parallelizable (one worker independently performs inference for one shard) and is communication-efficient (only one round of communication is required at the end of the training). However, there are several potential disadvantages: (i) it is not clear whether the prior sharing schemes discussed above will over-regularize or under-regularize the network compared to the original batch training scheme, and (ii) since each shard develops independent approximations and the model is unidentifiable, it is unclear if the simple combination rules above are appropriate. For example, different members of the committee might learn equivalent posteriors up to a permutation of the hidden units. Although initializing each approximate posterior $q_k(\theta)$ in the same way can mitigate this effect, the lack of a shared context is likely to be problematic.
We evaluate BCM for both iid and non-iid settings, with different learning rates for the Adam optimizer for each worker and show the full results in \cref{fig:app:res_dist_bcm_iid,fig:app:res_dist_bcm_noniid}. It is perhaps surprising that BCM works well in the iid data setting, although the best error rate of $4\%$ is still much higher than state-of-the-art results on the MNIST classification task ($\sim1\%$). However, the concern above about the potential pitfalls when multiplying different sub-posteriors is validated in the non-iid setting, and in the iid setting when each worker is trained for a long time before performing the aggregation step. The best results in both settings are selected and shown in \cref{fig:res_dist_best_iid,fig:res_dist_best_noniid}.

\paragraph{Partitioned VI}
As discussed in \cref{sec:federated}, PVI is a natural fit to training probabilistic models on federated data and is flexible such that various communication and scheduling strategies can be employed. In this section, we test three approaches discussed in \cref{sec:federated}:

\begin{itemize}
    \item Sequential PVI with only one pass through the data set: The number of training epochs and learning rates for each worker are varied, and the full results are included in \cref{fig:app:res_dist_pvi_one_pass_iid,fig:app:res_dist_pvi_one_pass_noniid}. The results show that while this strategy is effective for the iid setting, it performs poorly in the non-iid setting. This issue is known in the continual learning literature where incremental learning of a single-head network is known to be challenging. Episodic memory \citep[see e.g.][]{zenke+al:17,nguyen+al:2018} or generative replay \citep{shin+al:17} is typically used to address this problem. The performance for the best hyperparameter settings are shown in \cref{fig:res_dist_best_iid,fig:res_dist_best_noniid}. 
    \item PVI with synchronous model updates: In this experiment, each worker runs one epoch of Global VI between message passing steps, and the parameter server waits for all workers to finish before aggregating information and sending it back to the workers. We explore different learning rates for the inner loop optimization and various damping rates for the parameter server, and show the full results in \cref{fig:app:res_dist_pvi_sync_iid,fig:app:res_dist_pvi_sync_noniid}. While the performance on the test set depends strongly on the learning rate and damping factor, if these values are appropriately chosen, this update strategy can achieve competitive performance ($\sim<2\%$). By analyzing the homogeneity of the data and updates across workers, some forms of heuristics could be used to choose the learning rate and damping factor --- we leave this for future work. We pick the best performing runs and compare with other methods in \cref{fig:res_dist_best_iid,fig:res_dist_best_noniid}.
    \item PVI with lock-free asynchronous updates: Similar to the synchronous PVI experiment, we vary the learning rate and damping factor and include the full results in \cref{fig:app:res_dist_pvi_async_iid,fig:app:res_dist_pvi_async_noniid}. The test performance of this method is generally worse compared the synchronous update scheme, since the context/cavity distribution could be changed while a worker is running and the next parameter update performed by this worker could overwrite the updates made by other workers. While we do not simulate conditions that favour this scheduling scheme such that unreliable communication channels or unbalanced data across memory shards, we expect this strategy to perform well compared to other methods in these scenarios. The best hyperparameters are selected and their performance are shown in  \cref{fig:res_dist_best_iid,fig:res_dist_best_noniid}.
\end{itemize}

\paragraph{Discussion} The best performance for each method discussed above are shown in \cref{fig:res_dist_best_iid,fig:res_dist_best_noniid}, demonstrating the accuracy-training time and accuracy-communication cost frontiers. In the iid data setting (\cref{fig:res_dist_best_iid}), distributed training methods can achieve comparable performance in the same training time compared to that of global VI. However, methods based on data partitioning are much more communication-efficient, for example, PVI-sync uses about 10 times fewer messages than GVI when both methods attain a 3$\%$ test error. The results for PVI-seq with one pass through the data demonstrates its efficiency, but highlights the need to revisit data multiple times to obtain better error rate and log-likelihood. BCM shows promising performance, but is outperformed by all other methods, suggesting that communication between workers and setting the right approximate prior (context) for each partition are crucial.

\Cref{fig:res_dist_best_noniid} shows the non-iid data regime is substantially more challenging, as simple training methods including BCM and PVI-seq with one pass perform poorly and other distributed methods require more extreme hyperparameter settings (e.g.~much smaller learning rate and higher damping factor), much longer training time, and higher communication cost to obtain a performance comparable to that in the iid regime. We note that the performance of PVI is significantly better than a recent result by \cite{zhao+al:18}, who achieved a 10\% error rate on the same non-iid data setting. Moreover, unlike this work, we use a fully-connected neural network (rather than a convolutional one) and do not communicate data between the workers (no data synchronization). As in the iid setting, the performance of PVI-async is hindered by stale updates, compared to PVI-sync, despite early faster convergence. While GVI with 10 cores is the best performer in terms of predictive performance, it is the least communication-efficient due to the need to frequently pass gradients between the central parameter server and compute nodes. This, however, suggests that the performance of PVI could be further improved by more frequent updates between workers, essentially trading off the communication cost for more accurate prediction.

\subsection{Improving Continual Learning for Sparse Gaussian Processes}
We evaluate the performance of the continual learning method for sparse Gaussian process models discussed in \cref{sec:continual} on a toy classification problem and a real-world regression problem. The different inference strategies were implemented in Tensorflow \citep{abadi+al:16} and GPflow \citep{matthews+al:17}.

\subsubsection{A comparison on a toy data set}
\begin{figure}[htb]
\centering
\includegraphics[width=\textwidth]{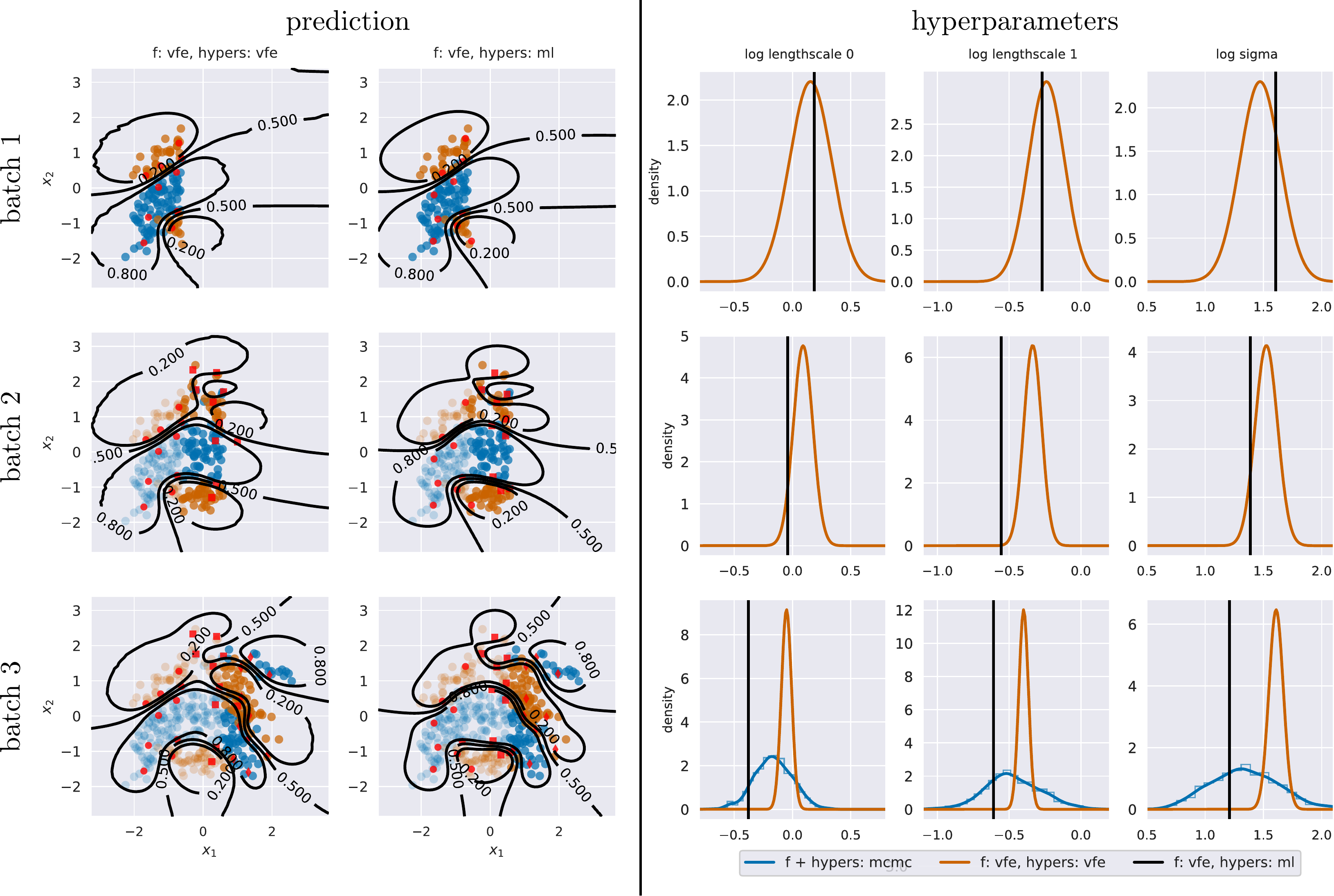}
\caption{Experimental results on a toy streaming data set: predictions after sequentially observing different data batches [left] and corresponding hyperparameter estimates [right]. Three methods were considered for this task: MCMC for both the latent function ($f$) and hyperparameters (hypers) with no sparsification, variational inference for both $f$ and hypers with inducing points, and sparse variational inference for $f$ and maximum likelihood estimation for hypers. Best viewed in colour.}
\label{fig:gp_continual_toy}
\end{figure}

A toy streaming data set was created using the {\it banana} data set, which comprises 400 two-dimensional inputs and corresponding binary targets. The data set is first ordered using one input dimension and then split into three equal batches. We consider two sparse variational approaches for inferring the latent function with 10 pseudo-points for each batch: (i) maximum-likelihood estimation for the hyperparameters --- this is similar to the method of \cite{bui+al:2017} but using private pseudo-points, as described in \cref{sec:appen:sgp}, and (ii) online variational inference for the hyperparameters, as discussed in \cref{sec:continual}. The key results are shown in \cref{fig:gp_continual_toy}, which includes the predictions after observing each data batch and the (distributional) hyperparameter estimates for both methods. We also include the histograms of the hyperparameter samples obtained by running MCMC for both the latent function and hyperparameters on the whole data set. As expected, the sparse variational methods underestimate the width of the distributions over the hyperparameters. The maximum-likelihood estimates for the hyperparameters tend to be smaller and to change faster when moving from one batch to another compared the VI estimates. Consequently, the prediction using the ML hyperparameters tends to have sharper decision boundaries. We include in \cref{sec:appen:sgp} a failure case of the ML approach where the hyperparameter values are {\it overfit} to a data batch, while the VI approach is more robust and maintains better prediction quality.  

\subsubsection{Learning inverse dynamics of a robot arm}
We next test the proposed method on learning inverse dynamics of a robot arm. The data set is generated using a Barrett WAM arm with seven degrees-of-freedom --- seven joints to which control torques can be applied and the joints' angles, speeds and accelerations can be recorded accordingly \citep{nguyen-tuong+al:09}. The aim is to learn the inverse dynamics of the arm, i.e.~to accurately predict the forces used at different joints given the joints' current characteristics. We treat this task as a regression problem with 7 independent outputs and 21 inputs. The data set consists of 12,000 points for training and 3,000 points for prediction. 

\begin{figure}[htb]
\centering
\includegraphics[width=\textwidth]{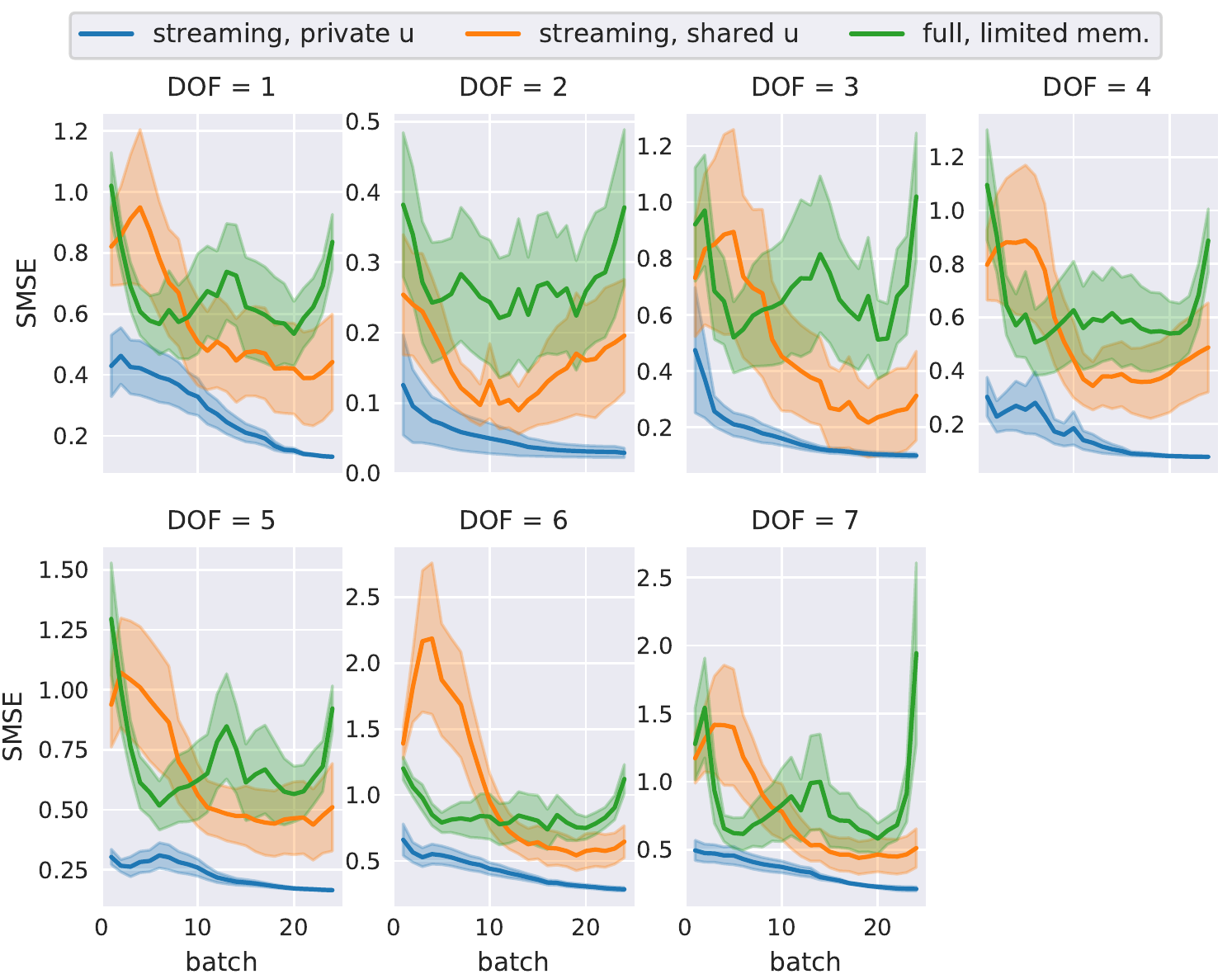}
\caption{Experimental results of learning inverse dynamics of a robot arm, i.e.~predicting the forces applied to the joints of the arm given the locations, speeds, and accelerations of the joints. Three methods were considered for this task: streaming sparse variational GP with private pseudo-points, streaming sparse variational GP with shared pseudo-points, and full GP with limited memory. Full details are included in the text. Best viewed in colour.\label{fig:gp_continual_robot}}
\end{figure}

To simulate the streaming setting, we first sort the data points using a joint's location and form 24 batches of 500 points each. As there are seven joints in the arm, seven streaming data sets were created, each corresponding to one joint being used for sorting. For the proposed method in \cref{sec:continual}, 10 pseudo-points are allocated and optimized for each batch, and as such, there are 240 pseudo-points in total at the end of training. We predict on the test set after sequentially showing a data batch to the model and compute the standardized mean squared errors (SMSEs). The results, averaged over multiple runs corresponding to different inputs used for sorting, are shown in \cref{fig:gp_continual_robot}. Two additional methods were considered: (i) full GP with limited memory of 1500 data points, retrained from scratch after seeing each data batch, and (ii) streaming sparse GPs using variational inference for both latent function and hyperparameters with 240 pseudo-points being shared over all batches and re-optimized after every batch. For both sparse GP methods, online variational inference is used for the hyperparameters. The results demonstrate the proposed method with private pseudo-points is most effective and stable during training among all methods considered. For example, for the third degree-of-freedom, the final SMSEs were $0.100\mypm0.004$ for the proposed method with private pseudo-points, $0.313\mypm0.079$ for the method with global pseudo-points, and $1.021\mypm0.113$ for exact GP with limited memory. While the method with shared pseudo-points has more pseudo-points for earlier batches and is expected to perform better theoretically, this experiment shows its inferior performance due to the need to reinitialize and optimize all pseudo-points at every batch. The full GP with limited memory approach performs poorly and exhibits forgetting as more batches arrived and old data are excluded from the memory. In addition, we also tried the streaming inference scheme of \cite{bui+al:2017}, which only retains a point estimate of the hyperparameters after each batch, but this did not perform well, demonstrating that being distributional over the hyperparameters is crucial.

\section{Conclusion}
This paper provided a general and unifying view of variational inference, Partitioned Variational Inference, for probabilistic models. 
We showed that the PVI framework flexibly subsumes many existing variational inference methods as special cases and allows a wealth of techniques to be connected. We also demonstrated how PVI allows novel algorithmic developments and practical applications. This is illustrated through the development of a streaming variational inference scheme for Gaussian process models and a communication efficient algorithm for training Bayesian neural networks on federated data.

One of the key contributions of this work is the connection of deterministic local message passing methods with global optimization-based schemes. Each of these different branches of approximate inference has been arguably developed exclusively of each other, for example, existing probabilistic programming toolkits tend to work primarily with one approach but not both \citep[see e.g.][]{tran+al:2017,minka+al:2018}. This paper suggests these methods are inter-related and practitioners could benefit from a unified framework, i.e.~there are ways to expand the existing probabilistic programming packages to gracefully handle both approaches. Additionally, the PVI framework could be used to automatically choose a granularity level and an optimization scheme that potentially offer a better inference method for the task or the model at hand. It is, however, unclear how flexible variational approximations such as mixtures of exponential family distributions \cite[see e.g.][]{sudderth+al:2010} or normalizing flows \citep{rezende+mohamed:2015} can be efficiently and tractably accommodated in the PVI framework. We leave these directions as future work.

The experiments in \cref{sec:exp_federated} demonstrated that PVI is well-suited to learning with decentralized data. Deployment of PVI in this setting is practical as its implementation only requires a straightforward modification of existing global VI implementations. We have also explored how this algorithm allows data parallelism --- each local worker stores a complete copy of the model --- and communication efficient, uncertainty-aware updates between workers. A potential future extension of the proposed approach is model parallelism. That is, in addition to decentralizing the data and computation across the workers, the model itself is partitioned. As commonly done in many deep learning training algorithms, model parallelism could be achieved by assigning the parameters (and computation) of different layers of the network to different devices. Another potential research direction is coordinator-free, peer-to-peer only communication between workers. This could be achieved by a worker passing each update to several randomly selected other workers, who then apply the changes, rather than to a central parameter server.

\bibliographystyle{plainnat}
\bibliography{vmp}

\newpage

\appendix

\section{Proofs}

\subsection{Relating Local KL minimization to Local Free-energy Minimization: Proof of Property \ref{prop:local-fe-opt}}
\label{sec:appen:local-FE-KL}
Substituting the tilted distribution $\widehat{p}^{(i)}(\theta)$ into the KL divergence yields,
\begin{align*}
\mathrm{KL} \left( q(\theta) \| \widehat{p}^{(i)}(\theta) \right)  &= \int \mathrm{d}\theta q(\theta) \log  \frac{  q(\theta) \widehat{\mathcal{Z}}_i t^{(i-1)}_{b_i}(\theta) }{ q^{(i-1)}(\theta) p(\vect{y}_{b_i}|\theta)}
= \log \widehat{\mathcal{Z}}_i - \int \mathrm{d}\theta q(\theta) \log  \frac{ q^{(i-1)}(\theta) p(\vect{y}_{b_i}|\theta) }{q(\theta)t^{(i-1)}_{b_i}(\theta)}.
\end{align*}
Hence $\mathcal{F}^{(i)}(q(\theta)) = \log \widehat{\mathcal{Z}}_i - \mathrm{KL} \left( q(\theta) \| \widehat{p}^{(i)}(\theta) \right)$.

\subsection{Showing Valid normalization of $q(\theta)$: Proof of Property \ref{prop:valid-prob}}
 \label{sec:appen:normalizedt}
This property holds for $i = 0$. By induction, assuming $q^{(i-1)}(\theta) = p(\theta) \prod_m t^{(i-1)}_m(\theta)$, we have:
\begin{align*}
p(\theta) \prod_m t^{(i)}_m(\theta) &= p(\theta) t^{(i)}_{b_i}(\theta) \prod_{m \neq b_i} t^{(i)}_m(\theta)
= p(\theta) \frac{q^{(i)}(\theta)}{q^{(i-1)}(\theta)} t^{(i-1)}_{b_i}(\theta) \prod_{m \neq b_i} t^{(i-1)}_m(\theta) \\
&= \frac{q^{(i)}(\theta)}{q^{(i-1)}(\theta)} p(\theta) \prod_m t^{(i-1)}_m(\theta)
= q^{(i)}(\theta).
\end{align*}
Hence, the property holds for all $i$.

\subsection{Maximization of local free-energies implies maximization of global free-energy: Proof of Property \ref{prop:fixed-point}}
 \label{sec:appen:local-global}

(a) We have:
\begin{align*}
\sum_m \mathcal{F}_m(q^*(\theta)) = \sum_m \int \mathrm{d}\theta q^*(\theta) \log \frac{p(\vect{y}_m|\theta)}{t^*_m(\theta)} = \int \mathrm{d}\theta q^*(\theta) \log \frac{p(\theta) \prod_m p(\vect{y}_n|\theta)}{p(\theta) \prod_m t^*_m(\theta)} = \mathcal{F}(q^*(\theta)).
\end{align*}
(b) Let $\eta_q$ and $\eta^*_q$ be the variational parameters of $q(\theta)$ and $q^*(\theta)$ respectively. We can write $q(\theta) = q(\theta; \eta_q)$ and $q^*(\theta) = q(\theta; \eta^*_q) = p(\theta) \prod_m t_m(\theta; \eta^*_q)$. First, we show that at convergence, the derivative of the global free-energies equals the sum of the derivatives of the local free-energies. The derivative of the local free-energy $\mathcal{F}_m(q(\theta))$ w.r.t. $\eta_q$ is:
\begin{align*}
\frac{\mathrm{d} \mathcal{F}_m(q(\theta))}{\mathrm{d} \eta_q} &= \frac{\mathrm{d}}{\mathrm{d} \eta_q} \int \mathrm{d}\theta q(\theta;\eta_q) \log \frac{q(\theta; \eta^*_q) p(\vect{y}_m|\theta)}{q(\theta;\eta_q) t_m(\theta; \eta^*_q)}\\
&= \frac{\mathrm{d}}{\mathrm{d} \eta_q}  \int \mathrm{d}\theta q(\theta;\eta_q) \log \frac{p(\vect{y}_m|\theta)}{t_m(\theta; \eta^*_q)} + \frac{\mathrm{d}}{\mathrm{d} \eta_q} \int \mathrm{d}\theta q(\theta;\eta_q) \log \frac{q(\theta; \eta^*_q)}{q(\theta;\eta_q)}\\
&= \frac{\mathrm{d}}{\mathrm{d} \eta_q} \int \mathrm{d}\theta q(\theta;\eta_q) \log \frac{p(\vect{y}_m|\theta)}{t_m(\theta; \eta^*_q)} + \int \mathrm{d}\theta \frac{\mathrm{d} q(\theta;\eta_q)}{\mathrm{d} \eta_q} \log \frac{q(\theta; \eta^*_q)}{q(\theta;\eta_q)} - \cancelto{0}{\int \mathrm{d}\theta \frac{\mathrm{d} q(\theta;\eta_q)}{\mathrm{d} \eta_q}}.
\end{align*}
Thus, at convergence when $\eta_q = \eta^*_q$,
\begin{align*}
\frac{\mathrm{d} \mathcal{F}_m(q(\theta))}{\mathrm{d} \eta_q} \biggr|_{\eta_q = \eta^*_q} = \frac{\mathrm{d}}{\mathrm{d} \eta_q} \int \mathrm{d}\theta q(\theta;\eta_q) \log \frac{p(\vect{y}_m|\theta)}{t_m(\theta; \eta^*_q)} \biggr|_{\eta_q = \eta^*_q}.
\end{align*}
Summing both sides over all $m$,
\begin{align*}
\sum_m \frac{\mathrm{d} \mathcal{F}_m(q(\theta))}{\mathrm{d} \eta_q} \biggr|_{\eta_q = \eta^*_q} &= \frac{\mathrm{d}}{\mathrm{d} \eta_q} \int \mathrm{d}\theta q(\theta;\eta_q) \log \frac{\prod_m p(\vect{y}_m|\theta)}{\prod_m t_m(\theta; \eta^*_q)} \biggr|_{\eta_q = \eta^*_q} \\
& = \frac{\mathrm{d}}{\mathrm{d} \eta_q} \int \mathrm{d}\theta q(\theta;\eta_q) \log \frac{p(\theta) \prod_m p(\vect{y}_m|\theta)}{q(\theta;\eta^*_q)} \biggr|_{\eta_q = \eta^*_q}.
\end{align*}
Now consider the derivative of the global free-energy $\mathcal{F}(q(\theta))$:
\begin{align*}
\frac{\mathrm{d} \mathcal{F}(q(\theta))}{\mathrm{d} \eta_q} &= \frac{\mathrm{d}}{\mathrm{d} \eta_q} \int \mathrm{d}\theta q(\theta; \eta_q) \log \frac{p(\theta) \prod_m p(\vect{y}_m|\theta)}{q(\theta; \eta_q)} \\
&= \int \mathrm{d}\theta \frac{\mathrm{d} q(\theta; \eta_q)}{\mathrm{d} \eta_q} \log \frac{p(\theta) \prod_m p(\vect{y}_m|\theta)}{q(\theta; \eta_q)} - \cancelto{0}{\int \mathrm{d}\theta \frac{\mathrm{d} q(\theta;\eta_q)}{\mathrm{d} \eta_q}}.
\end{align*}
Hence,
\begin{align*}
\frac{\mathrm{d} \mathcal{F}(q(\theta))}{\mathrm{d} \eta_q} \biggr|_{\eta_q = \eta^*_q} = \int \mathrm{d}\theta \frac{\mathrm{d} q(\theta; \eta_q)}{\mathrm{d} \eta_q} \log \frac{p(\theta) \prod_m p(\vect{y}_m|\theta)}{q(\theta; \eta^*_q)} \biggr|_{\eta_q = \eta^*_q} = \sum_m \frac{\mathrm{d} \mathcal{F}_m(q(\theta))}{\mathrm{d} \eta_q} \biggr|_{\eta_q = \eta^*_q}.
\end{align*}
For all $m$, since $q^*(\theta) = \argmax_{q(\theta) \in \mathcal{Q}} \mathcal{F}_m(q(\theta))$, we have $\frac{\mathrm{d} \mathcal{F}_m(q(\theta))}{\mathrm{d} \eta_q} \big|_{\eta_q = \eta^*_q} = 0$, which implies:
\begin{align*}
\frac{\mathrm{d} \mathcal{F}(q(\theta))}{\mathrm{d} \eta_q} \biggr|_{\eta_q = \eta^*_q} = \sum_m \frac{\mathrm{d} \mathcal{F}_m(q(\theta))}{\mathrm{d} \eta_q} \biggr|_{\eta_q = \eta^*_q} = 0.
\end{align*}
Thus, $q^*(\theta)$ is an extremum of $\mathcal{F}(q(\theta))$.

Now we show that $q^*(\theta)$ is a maximum of $\mathcal{F}(q(\theta))$ by considering the Hessian $\frac{\mathrm{d}^2 \mathcal{F}(q(\theta))}{\mathrm{d} \eta_q \mathrm{d} \eta^\intercal_q}$ of the global free-energy at convergence. Similar to the derivative case, we now show that the Hessian of the global free-energies equals the sum of the Hessians of the local free-energies. The Hessian of the local free-energy $\mathcal{F}_m(q(\theta))$ w.r.t. $\eta_q$ is:
\begin{align*}
\frac{\mathrm{d}^2 \mathcal{F}_m(q(\theta))}{\mathrm{d} \eta_q \mathrm{d} \eta^\intercal_q} &= \frac{\mathrm{d}^2}{\mathrm{d} \eta_q \mathrm{d} \eta^\intercal_q} \int \mathrm{d}\theta q(\theta;\eta_q) \log \frac{q(\theta; \eta^*_q) p(\vect{y}_m|\theta)}{q(\theta;\eta_q) t_m(\theta; \eta^*_q)} \\
&= \frac{\mathrm{d}^2}{\mathrm{d} \eta_q \mathrm{d} \eta^\intercal_q}  \int \mathrm{d}\theta q(\theta;\eta_q) \log \frac{p(\vect{y}_m|\theta)}{t_m(\theta; \eta^*_q)} + \frac{\mathrm{d}^2}{\mathrm{d} \eta_q \mathrm{d} \eta^\intercal_q} \int \mathrm{d}\theta q(\theta;\eta_q) \log \frac{q(\theta; \eta^*_q)}{q(\theta;\eta_q)} \\
&= \frac{\mathrm{d}^2}{\mathrm{d} \eta_q \mathrm{d} \eta^\intercal_q} \int \mathrm{d}\theta q(\theta;\eta_q) \log \frac{p(\vect{y}_m|\theta)}{t_m(\theta; \eta^*_q)} + \int \mathrm{d}\theta \frac{\mathrm{d}^2 q(\theta;\eta_q)}{\mathrm{d} \eta_q \mathrm{d} \eta^\intercal_q} \log \frac{q(\theta; \eta^*_q)}{q(\theta;\eta_q)} \\
& \qquad \qquad \qquad \qquad \qquad \qquad \qquad \qquad - \cancelto{0}{\frac{\mathrm{d}}{\mathrm{d} \eta_q} \int \mathrm{d}\theta \frac{\mathrm{d} q(\theta;\eta_q)}{\mathrm{d} \eta^\intercal_q}}.
\end{align*}
At convergence when $\eta_q = \eta^*_q$,
\begin{align*}
\frac{\mathrm{d}^2 \mathcal{F}_m(q(\theta))}{\mathrm{d} \eta_q \mathrm{d} \eta^\intercal_q} \biggr|_{\eta_q = \eta^*_q} 
= \frac{\mathrm{d}^2}{\mathrm{d} \eta_q \mathrm{d} \eta^\intercal_q} \int \mathrm{d}\theta q(\theta;\eta_q) \log \frac{p(\vect{y}_m|\theta)}{t_m(\theta; \eta^*_q)} \biggr|_{\eta_q = \eta^*_q}.
\end{align*}
Summing both sides over all $m$,
\begin{align*}
\sum_m \frac{\mathrm{d}^2 \mathcal{F}_m(q(\theta))}{\mathrm{d} \eta_q \mathrm{d} \eta^\intercal_q} \biggr|_{\eta_q = \eta^*_q} 
&= \frac{\mathrm{d}^2}{\mathrm{d} \eta_q \mathrm{d} \eta^\intercal_q} \int \mathrm{d}\theta q(\theta;\eta_q) \log \frac{\prod_m p(\vect{y}_m|\theta)}{\prod_m t_m(\theta; \eta^*_q)} \biggr|_{\eta_q = \eta^*_q} \\
&= \frac{\mathrm{d}^2}{\mathrm{d} \eta_q \mathrm{d} \eta^\intercal_q} \int \mathrm{d}\theta q(\theta;\eta_q) \log \frac{p(\theta) \prod_m p(\vect{y}_m|\theta)}{q(\theta;\eta^*_q)} \biggr|_{\eta_q = \eta^*_q}.
\end{align*}
Now consider the Hessian of the global free-energy $\mathcal{F}(q(\theta))$:
\begin{align*}
\frac{\mathrm{d}^2 \mathcal{F}(q(\theta))}{\mathrm{d} \eta_q \mathrm{d} \eta^\intercal_q} &= \frac{\mathrm{d}^2}{\mathrm{d} \eta_q \mathrm{d} \eta^\intercal_q} \int \mathrm{d}\theta q(\theta; \eta_q) \log \frac{p(\theta) \prod_m p(\vect{y}_m|\theta)}{q(\theta; \eta_q)} \\
&= \int \mathrm{d}\theta \frac{\mathrm{d}^2 q(\theta; \eta_q)}{\mathrm{d} \eta_q \mathrm{d} \eta^\intercal_q} \log \frac{p(\theta) \prod_m p(\vect{y}_m|\theta)}{q(\theta; \eta_q)} - \cancelto{0}{\frac{\mathrm{d}}{\mathrm{d} \eta_q} \int \mathrm{d}\theta \frac{\mathrm{d} q(\theta;\eta_q)}{\mathrm{d} \eta^\intercal_q}}.
\end{align*}
Hence,
\begin{align*}
\frac{\mathrm{d}^2 \mathcal{F}(q(\theta))}{\mathrm{d} \eta_q \mathrm{d} \eta^\intercal_q} \biggr|_{\eta_q = \eta^*_q} 
= \int \mathrm{d}\theta \frac{\mathrm{d}^2 q(\theta; \eta_q)}{\mathrm{d} \eta_q \mathrm{d} \eta^\intercal_q} \log \frac{p(\theta) \prod_m p(\vect{y}_m|\theta)}{q(\theta; \eta_q)} \biggr|_{\eta_q = \eta^*_q}
= \sum_m \frac{\mathrm{d}^2 \mathcal{F}_m(q(\theta))}{\mathrm{d} \eta_q \mathrm{d} \eta^\intercal_q} \biggr|_{\eta_q = \eta^*_q}.
\end{align*}
For all $m$, since $q^*(\theta) = \argmax_{q(\theta) \in \mathcal{Q}} \mathcal{F}_m(q(\theta))$, the Hessian $\frac{\mathrm{d}^2 \mathcal{F}_m(q(\theta))}{\mathrm{d} \eta_q \mathrm{d} \eta^\intercal_q} \big|_{\eta_q = \eta^*_q}$ is negative definite, which implies that the Hessian $\frac{\mathrm{d}^2 \mathcal{F}(q(\theta))}{\mathrm{d} \eta_q \mathrm{d} \eta^\intercal_q} \big|_{\eta_q = \eta^*_q}$ of the global free-energy is also negative definite. Therefore, $q^*(\theta)$ is a maximum of $\mathcal{F}(q(\theta))$.

\subsection{Derivation of fixed point equations (property \ref{prop:fixed-point-equations})}
\label{sec:appen:fp}
Assume that the approximate likelihoods $\{t_m(\theta)\}_{m=1}^M$ are in the un-normalized exponential family. That is, $t_m(\theta) = \exp(\eta_m^\intercal T(\theta) + c_m)$ for some constant $c_m$. In this section we  absorb the constant into the natural parameters $\eta^\intercal_m \leftarrow [c_m, \eta^\intercal_m]$ and add a corresponding unit element into the sufficient statistics $T(\theta)^\intercal \leftarrow [1,T(\theta)^\intercal]$. To lighten notation we still denote the sufficient statistic vector as $T(\theta)$ so , 
\begin{align}
\label{eq:exp-family}
t_m(\theta) = t_m(\theta ; \eta_m) = \exp(\eta_m^\intercal T(\theta)),
\end{align}
where $\eta_m$ is the natural parameters and $T(\theta)$ is the augmented sufficient statistics.
We also assume that the prior $p(\theta)$ and the variational distributions $q^{(i)}(\theta)$ are normalized exponential family distributions.

To derive the fixed point updates for the local variational inference algorithm, we consider maximizing the local variational free energy $\mathcal{F}^{(i)}(q(\theta))$ in \eqref{eq:vfe}. Assuming that at iteration $i$, the variational distribution $q^{(i-1)}(\theta)$ obtained from the previous iteration has the form:
\begin{align}
q^{(i-1)}(\theta) = \exp( \eta_{\textrm{prev}}^\intercal T(\theta) - A(\eta_{\textrm{prev}})),
\end{align}
and the target distribution $q(\theta)$ that we optimize has the form:
\begin{align}
q(\theta) = \exp( \eta_q^\intercal T(\theta) - A(\eta_q) ),
\end{align}
where $A(\cdot)$ is the log-partition function. Let ${ \eta = \eta_{\textrm{prev}} - \eta_q - \eta^{(i-1)}_{b_i} }$, we can write $\mathcal{F}^{(i)}(q(\theta))$ as:
\begin{align*}
\mathcal{F}^{(i)}(q(\theta)) &= A(\eta_q) - A(\eta_{\textrm{prev}}) + \mathbb{E}_q (\log p(\vect{y}_{b_i}|\theta)) + \eta^\intercal \mathbb{E}_q (T(\theta)).
\end{align*}
Take the derivative of $\mathcal{F}^{(i)}(q(\theta))$ w.r.t.~$\eta_q$ and note that $\frac{\mathrm{d}A(\eta_q)}{\mathrm{d}\eta_q} = \mathrm{E}_q (T(\theta))$, we have:
\begin{align*}
\frac{\mathrm{d}\mathcal{F}^{(i)}(q(\theta))}{\mathrm{d}\eta_q}  &= \frac{\mathrm{d}}{\mathrm{d}\eta_q} \mathbb{E}_q (\log p(\vect{y}_{b_i}|\theta)) + \frac{\mathrm{d}^2A(\eta_q)}{\mathrm{d}\eta_q\mathrm{d}\eta_q} \eta.
\end{align*}
Set this derivative to zero, we obtain:
\begin{align}
\label{eq:fixed-point-q}
\eta_q  &= \mathbb{C}^{-1} \frac{\mathrm{d}}{\mathrm{d}\eta_q} \mathbb{E}_q (\log p(\vect{y}_{b_i}|\theta)) + \eta_{\textrm{prev}} - \eta^{(i-1)}_{b_i},
\end{align}
where $\mathbb{C} \defeq \frac{\mathrm{d}^2A(\eta_q)}{\mathrm{d}\eta_q\mathrm{d}\eta_q} = \mathrm{cov}_{q(\theta)}[ T(\theta) T^\intercal(\theta)]$ is the Fisher Information.
Note that from Property \ref{prop:valid-prob}, 
$\eta_q = \eta_0 + \sum_{m \neq b_i} \eta^{(i-1)}_m + \eta^{(i)}_{b_i}$ and ${ \eta_{\textrm{prev}} = \eta_0 + \sum_m \eta^{(i-1)}_m }$, where $\eta_0$ is the natural parameters for the prior $p(\theta)$.
Hence, from \eqref{eq:fixed-point-q}, a fixed point update for Algorithm \ref{alg:vmp} only needs to update $\eta^{(i)}_{b_i}$ locally using:
\begin{align}
\eta^{(i)}_{b_i}  &= \mathbb{C}^{-1} \frac{\mathrm{d}}{\mathrm{d}\eta_q} \mathbb{E}_q (\log p(\vect{y}_{b_i}|\theta)).
\end{align}
The Fisher Information can be written as $\mathbb{C} = \frac{\mathrm{d}\mu_q}{\mathrm{d}\eta_q}$ where $\mu_q = \mathbb{E}_q (T(\theta))$ is the mean parameter of $q(\theta)$. This leads to a cancellation of the Fisher information,
\begin{align}
\eta^{(i)}_{b_i}  &
= \frac{\mathrm{d}}{\mathrm{d}\mu_q} \mathbb{E}_q (\log p(\vect{y}_{b_i}|\theta)).
\end{align}

\subsection{Equivalence of local and global fixed-points: Proof of property \ref{prop:local-global-fp}}
\label{sec:local-global-same}
Here we show that running parallel fixed-point updates of local-VI with $M$ data groups has equivalent dynamics for $q(\theta)$ as running the fixed points for batch VI ($M=1$). The local-VI fixed point updates (property \ref{prop:fixed-point-equations}) are given by,
\begin{align}
\label{eq:fixed-point-bi}
\eta^{(i)}_{b_i}  &
= \frac{\mathrm{d}}{\mathrm{d}\mu_{q^{(i-1)}}} \mathbb{E}_{q^{(i-1)}} (\log p(\vect{y}_{b_i}|\theta)).
\end{align}
Here we have explicitly denoted the dependence of the approximate posterior on the iteration number $i$ as the dynamics of this is the focus. When $M=N$ and a parallel fixed point update of all the natural parameters, $\{ \eta^{(i)}_{n} \}_{n=1}^N$, is performed, the natural parameters of $q(\theta)$ are updated as follows,
\begin{align}
\eta_q^{(i)} & = \eta_0 + \sum_{m=1}^M \eta^{(i)}_n = \eta_0 + \sum_{m=1}^M \frac{\mathrm{d}}{\mathrm{d}\mu_{q^{(i-1)}}} \mathbb{E}_{q^{(i-1)}} (\log p(\vect{y}_{b_i}|\theta)) \nonumber\\ 
& = 
\eta_0 + \sum_{n=1}^N \frac{\mathrm{d}}{\mathrm{d}\mu_{q^{(i-1)}}} \mathbb{E}_{q^{(i-1)}} (\log p(y_{n}|\theta)).
\end{align}
Here, in the last line, we have used the fact that the data are independent conditioned on $\theta$. Now consider application of the fixed-point updates to the batch case ($M=1$) 
\begin{align}
\eta_q^{(i)} = \eta_0 +  \eta = \eta_0 + \frac{\mathrm{d}}{\mathrm{d}\mu_{q^{(i-1)}}} \mathbb{E}_q (\log p(\vect{y}|\theta)) = \eta_0 + \sum_{n=1}^N \frac{\mathrm{d}}{\mathrm{d}\mu_{q^{(i-1)}}} \mathbb{E}_{q^{(i-1)}} (\log p(\vect{y}_{n}|\theta))
\end{align}
Therefore the updates for $q(\theta)$ are identical in the two cases. Na\"{i}ve implementation of local-VI requires $M$ sets of natural parameters to be maintained, whereas parallel updating means this is unnecessary, but equivalent to fixed-point global VI. 

\subsection{Relations between methods employing stochastic approximation of fixed-points / natural gradient updates} \label{sec:stochastic-approx}

The damped simplified fixed-point updates for global VI are,
\begin{align}
\label{eq:fixed-point-bi}
\eta^{(i)}_{q}  &
= (1-\rho)\eta^{(i-1)}_{q} + \rho \left ( \eta_0 + \frac{\mathrm{d}}{\mathrm{d}\mu_q} \mathbb{E}_q (\log p(\vect{y}|\theta)) \right).
\end{align}
\citet{hoffman+al:2013,sheth+khardon:2016b} employ stochastic mini-batch approximation of $\mathbb{E}_q (\log p(\vect{y}|\theta)) = \sum_{n} \mathbb{E}_{q(\theta)} (\log p(y_n|\theta)) = N \mathbb{E}_{p_{\text{data}}(y),q(\theta)} (\log p(y_n|\theta)) $ by sub-sampling the data distribution $\mathbf{y}_l \stackrel{\text{iid}}{\sim} p_{\text{data}}(y)$ where $p_{\text{data}}(y) = \frac{1}{N} \sum_{n=1}^N \delta(y - y_n)$. This approximation yields,
\begin{align}
\eta^{(i)}_{q}  &
= (1-\rho)\eta^{(i-1)}_{q} + \rho \left ( \eta_0 + L \frac{\mathrm{d}}{\mathrm{d}\mu_q} \mathbb{E}_q (\log p(\vect{y}_l|\theta)) \right). \label{eq:stoch-global}
\end{align}
Where $\vect{y}_l$ are a mini-batch containing $N/L$ data points.  \citet{li+al:2015} show that their stochastic power EP algorithm recovers precisely these same updates when $\alpha \rightarrow 0$ (this is related to property \ref{prop:pep-fp}). The relation to EP-like algorithms can be made more explicit by writing  \ref{eq:stoch-global} as
\begin{align}
\eta^{(i)}_{q}  &
= \eta^{(i-1)}_{q} + \rho' \left (   \frac{\mathrm{d}}{\mathrm{d}\mu_q} \mathbb{E}_q (\log p(\vect{y}_l|\theta)) - \eta^{(i-1)}_{\text{like}}/L \right). \label{eq:stoch-global-EP}
\end{align}
Where $\rho' = \rho L$ is a rescaled learning rate and $\eta^{(i-1)}_{\text{like}} = \eta^{(i-1)}_{q} - \eta^{(i-1)}_{0}$ is the portion of the approximate posterior natural parameters that approximates the likelihoods. As such, $\eta^{(i-1)}_{\text{like}}/L$ is the contribution a mini-batch likelihood makes on average to the posterior. 
So, interpreting the update in these terms, the new approximate posterior is equal to the old approximate posterior with $\rho'$ of the average mini-batch likelihood approximation removed and $\rho'$ of the approximation from the current mini-batch, $\frac{\mathrm{d}}{\mathrm{d}\mu_q} \mathbb{E}_q (\log p(\vect{y}_m|\theta))$, added in place of it.

\citet{khan+li:2018} take a different approach employing damped simplified fixed-point updates for local VI $M=N$ and then using an update schedule that selects a mini-batch at random and then updates the local natural parameters for these data-points in parallel. That is for all data points in the mini-batch they apply
\begin{align}
\eta^{(i)}_{n}  & = (1-\rho) \eta^{(i-1)}_{n} + \rho \frac{\mathrm{d}}{\mathrm{d}\mu_{q}} \mathbb{E}_{q} (\log p(y_{n}|\theta)). \label{eq:stoch-fully-local}
\end{align}
This local update incurs a  memory overhead that is $N$ times larger due to the need maintain $N$ sets of local parameters rather than just one. However, if the mini-batch partition is fixed across epochs, then it is sufficient to maintain $M$ natural parameters instead, corresponding to one for each mini-batch,
\begin{align}
\eta^{(i)}_{m}  & = (1-\rho) \eta^{(i-1)}_{m} + \rho \frac{\mathrm{d}}{\mathrm{d}\mu_{q}} \mathbb{E}_{q} (\log p(\vect{y}_{m}|\theta)). \label{eq:stoch-local}
\end{align}
Interestingly, both of these local updates (\ref{eq:stoch-fully-local} and \ref{eq:stoch-local}) result in a subtly different update to $q$ as the stochastic global update above (\ref{eq:stoch-global-EP}),
\begin{align}
\eta^{(i)}_{q}  &
= \eta^{(i-1)}_{q} - \rho \left ( \frac{\mathrm{d}}{\mathrm{d}\mu_q} \mathbb{E}_q (\log p(\vect{y}_m|\theta)) - \eta^{(i-1)}_{m} \right).\label{eq:stoch-local-EP}
\end{align}

Here the deletion step is explicitly revealed again, but now it involves removing the natural parameters of the $m$th approximate likelihood $\eta^{(i-1)}_{m}$ rather than the average mini-batch likelihood approximation $\eta^{(i-1)}_{\text{like}}/L$.

\begin{figure}[!ht]
\begin{center}
\includegraphics[width=10cm]{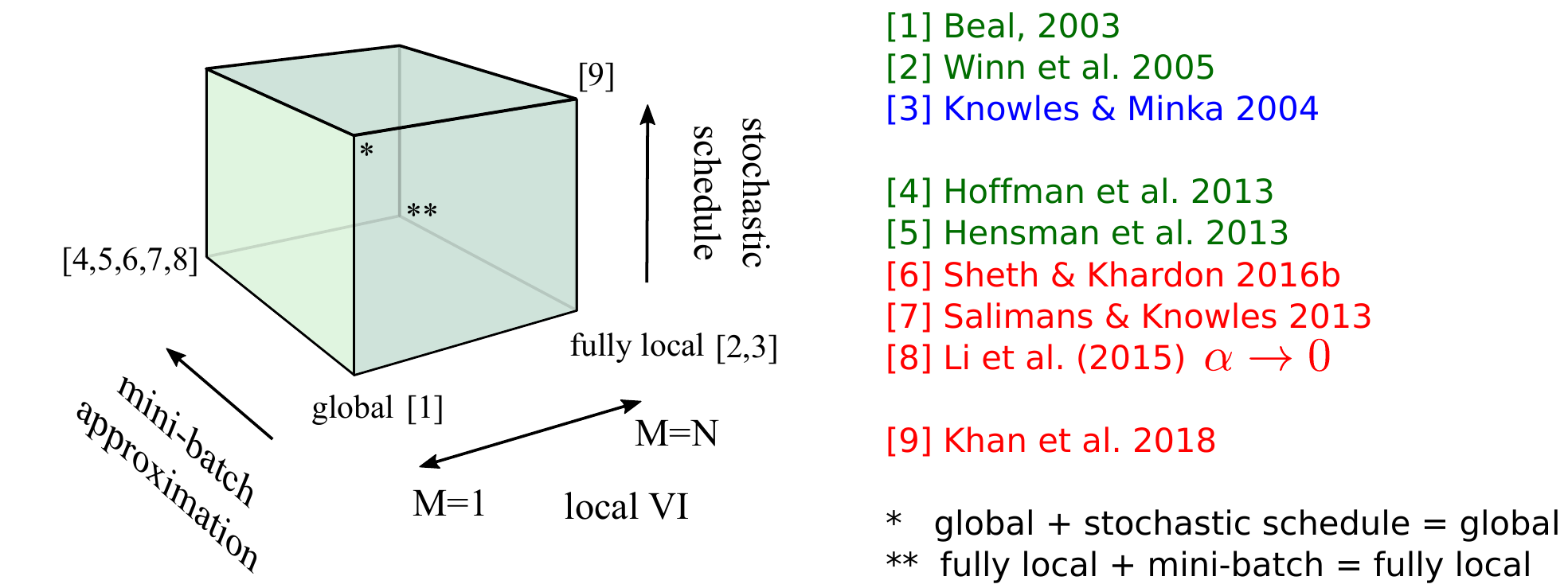}
\end{center}
\caption{Forms of stochastic approximation in the local VI framework and the relationship to previous work. The granularity of the approximation of the approximation is controlled though M. Mini-batch approximation may be used inside each local variational free-energy. A stochastic schedule can be used to randomize the order in which groups of data points are visited. All algorithms have the same fixed-points (the mini-batch approximation learning rate $\rho$ has to obey the Robbin's Munro conditions).}
\label{fig-stochastic}
\end{figure}

A summary of work employing these two types of stochastic approximation is provided in figure \ref{fig-stochastic}.

Each variety of update has its own pros and cons (compare \ref{eq:stoch-global-EP} to \ref{eq:stoch-local-EP}). The stochastic global update \ref{eq:stoch-global-EP} does not support general online learning and distributed asynchronous updates, but it is more memory efficient and faster to converge in the batch setting. 

Consider applying both methods to online learning where $\vect{y}_l$ or $\vect{y}_m$ correspond to the new data seen at each stage and $\rho'$ and $\rho$ are user-determined learning rates for the two algorithms. General online learning poses two challenges for the stochastic global update \ref{eq:stoch-global-EP}. First, the data are typically not iid (due to covariate or data set shift over time). Second, general online learning does not allow all old data points to be revisited, demanding incremental updates instead. This means that when a new batch of data is received we must iterate on just these data to refine the approximate posterior, before moving on and potentially never returning. Iterating \ref{eq:stoch-global-EP} on this new data is possible, but it would have disastrous consequences as it again breaks the iid mini-batch assumption and would just result in $q$ fitting the new data and forgetting the old previously seen data. A single update could be made, but this will normally mean that the approach is data-inefficient and slow to learn. Iterating the local updates, on the other hand, \ref{eq:stoch-fully-local} or  \ref{eq:stoch-local} works nicely as these naturally incorporate a deletion step that removes just the contribution from the current mini-batch and can, therefore, be iterated to our heart's content.

Similar arguments can be made about the distributed setting too. Stochastic global updates \emph{could} be used in the distributed setting, with each worker querying a server for the current natural parameters $\eta_{q}$, computing $\Delta \eta_l = \frac{\mathrm{d}}{\mathrm{d}\mu_q} \mathbb{E}_q (\log p(\vect{y}_l|\theta)) - \eta_{\text{like}}/L$ and communicating this to a server. The server's role is to update the global parameters $\eta_{q}^{(\text{new})} =  \eta_{q}^{(\text{old})} + \Delta \eta_l$ and to send these new parameters to the workers. The difficulty is that this setup must obey the iid assumption so i) the data must be distributed across the workers in an iid way, and ii) the updates must be returned by each worker with the same regularity. In contrast, the stochastic local updates can be used in a very similar way without these restrictions.    

The stochastic global update \ref{eq:stoch-global} does have two important advantages over the local updates \ref{eq:stoch-local-EP}. First, the memory footprint is $L$ times smaller only requiring a single set of natural parameters to be maintained rather than $M$ of them. Second, it can be faster to converge when the mini-batches are iid. Contrast what happens when new data are seen for the first time in the two approaches. For simplicity assume $\rho' = \rho =1$. In the second approach, $\eta^{(i-1)}_{m} = 0$ as the data have not been seen before, but the first approach effectively uses $\eta^{(i-1)}_{m} = \eta^{(i-1)}_{\text{like}}/L$. That is, the first approach effectively estimates the approximate likelihood for new data, based on those for previously seen data. This is a sensible approximation for homogeneous mini-batches. A consequence of this is that the learning rate $\rho'$ can be much larger than $\rho$ (potentially greater than unity) resulting in faster convergence of the approximate posterior. 
It would be interesting to consider modifications of the local updates (\ref{eq:stoch-local-EP}) that estimate the $m$th approximate likelihood based on information from all other data partitions. For example, in the first pass through the data, the approximate likelihoods for unprocessed mini-batches could be updated to be equal to the last approximate likelihood or to the geometric average of previous approximate likelihoods. Alternatively, ideas from inference networks could be employed for this purpose.

\subsection{The relationship between natural gradient, mirror-descent,  trust-region and proximal methods}
\label{sec:nat-grad-mirror-trust}

Each step of gradient ascent of parameters $\eta$ on a cost $\mathcal{C}$ can be interpreted as the result of an optimization problem derived from linearizing the cost function around the old parameter estimate, $\mathcal{C}(\eta) \approx \mathcal{C}(\eta^{(i-1)}) + \nabla_{\eta} \mathcal{C}(\eta^{(i-1)}) (\eta - \eta^{(i-1)}) $ where $\nabla_{\eta} \mathcal{C}(\eta^{(i-1)}) = \frac{\mathrm{d}\mathcal{C}(\eta)}{\mathrm{d}\eta} |_{\eta = \eta^{(i-1)}}$ and adding a soft constraint on the norm of the parameters: 
\begin{align}
\eta^{(i)} = \eta^{(i-1)} + \rho \frac{\mathrm{d}\mathcal{C}(\eta)}{\mathrm{d}\eta} \;\; \Leftrightarrow \;\; \eta^{(i)} = \argmax_{ \eta }  \nabla_{\eta} \mathcal{C}(\eta^{(i-1)}) \eta - \frac{1}{2 \rho} || \eta - \eta^{(i-1)} ||_2^2 \nonumber
\end{align}
Here the terms in the linearized cost that do not depend on $\eta$ have been dropped as they do not effect the solution of the optimization problem.  The linearization of the cost ensures that there is an analytic solution to the optimization and the soft constraint ensures that we do not move too far from the previous setting of the parameters into a region where the linearization is inaccurate.  

This reframing of gradient ascent reveals that it is making an Euclidean assumption about the geometry parameter space and suggests generalizations of the procedure that are suitable for different geometries. In our case, the parameters are natural parameters of a distribution and measures of proximity that employ the KL divergence are natural. 

The main result of this section is that the following optimization problems:
\begin{align}
&\text{\it KL proximal method:} \; \; \eta_q^{(i)} = \argmax_{ \eta_q }  \;  \nabla_{\eta} \mathcal{F}^{(i)}(q^{(i-1)}(\theta)) \eta_q - \frac{1}{\rho} \mathrm{KL} \left( q(\theta) ~\|~ q^{(i-1)}(\theta) \right) \label{eq-opt1}\\
&\text{\it KL trust region:} \; \; \eta_q^{(i)} = \argmax_{ \eta_q }  \;  \nabla_{\eta} \mathcal{F}^{(i)}(q^{(i-1)}(\theta)) \eta_q  \; \text{s.t.} \; \mathrm{KL} \left( q(\theta) ~\|~ q^{(i-1)}(\theta) \right) \le \gamma \label{eq-opt2}\\
&\text{\it KL$^s$ trust region:} \; \; \eta_q^{(i)} = \argmax_{ \eta_q }  \;  \nabla_{\eta} \mathcal{F}^{(i)}(q^{(i-1)}(\theta)) \eta_q  \; \text{s.t.} \; \mathrm{KL}^{\text{s}} \left( q(\theta) ~\|~ q^{(i-1)}(\theta) \right) \le \gamma \label{eq-opt3}\\
&\text{\it mirror descent:} \; \; \mu_q^{(i)} = \argmax_{ \mu_q }  \; \nabla_{\mu} \mathcal{F}^{(i)}(q^{(i-1)}(\theta)) \mu_q  - \frac{1}{\rho} \mathrm{KL} \left( q^{(i-1)}(\theta)  ~\|~  q(\theta)\right) \label{eq-opt4}
\end{align}
All yield the same updates as the damped fixed point equations / natural gradient ascent:
\begin{align}
\eta^{(i)}_{b_i}  &
= (1-\rho)\eta^{(i-1)}_{b_i} + \rho \frac{\mathrm{d}}{\mathrm{d}\mu^{(i-1)}_q} \mathbb{E}_{q^{(i-1)}} (\log p(\vect{y}_{b_i}|\theta))\\
& = (1-\rho)\eta^{(i-1)}_{b_i} + \rho \left [ \frac{\mathrm{d} \mu^{(i-1)}_q}{\mathrm{d}\eta^{(i-1)}_q} \right ]^{-1} \frac{\mathrm{d}}{\mathrm{d}\eta^{(i-1)}_q} \mathbb{E}_{q^{(i-1)}} (\log p(\vect{y}_{b_i}|\theta)).
\end{align}
In the first three cases (\ref{eq-opt1} - \ref{eq-opt3}), this equivalence only holds exactly in the general case if the parameter changes $\Delta \eta_q^{(i)} =  \eta_q^{(i)}-\eta_q^{(i-1)}$ are small.

Here the {\it KL proximal method} is the straightforward generalization of the gradient ascent example that replaces the Euclidean norm by the exclusive KL divergence. The {\it KL trust region method}  uses a hard constraint on the same KL instead, but rewriting this as a Lagrangian recovers the KL proximal method with $1/\rho$ being the Lagrange multiplier. The {\it KL$^s$ trust region method},  often used to justify natural gradient ascent,  employs the symmetrized KL divergence instead of the exclusive KL. The symmetrized KL is the average of the exclusive and inclusive KLs, $\mathrm{KL}^{\text{s}} \left( q(\theta) ~\|~ q^{(i-1)}(\theta) \right) = \frac{1}{2} \mathrm{KL} \left( q(\theta) ~\|~ q^{(i-1)}(\theta) \right) +  \frac{1}{2} \mathrm{KL}\left( q^{(i-1)}(\theta)  ~\|~  q(\theta)\right)$.  {\it Mirror descent}, in its most general form, uses a Bregman divergence to control the extent to which the parameters change rather than a KL divergence. However, when applied to exponential families, the Bregman divergence becomes the inclusive KL divergence yielding the form above \citep{raskutti+mukherjee:2015,khan+al:2016,khan+li:2018}. Note that this last method operates in the mean parameter space and the equivalence is attained by mapping the mean parameter updates back to the natural parameters. Mirror descent has the advantage of not relying on the small parameter change assumption to recover natural gradient ascent. Having explained the rationale behind these approaches we will now sketch how they yield the fixed-point updates.

The equivalence of the {\it KL proximal method} can be shown by differentiating the cost wrt $ \eta_q $  and  substituting in the following expressions:
\begin{align*}
\nabla_{\eta} \mathcal{F}^{(i)}(q^{(i-1)}(\theta)) &= \frac{\mathrm{d}}{\mathrm{d}\eta_{q^{(i-1)}}} \mathbb{E}_{q^{(i-1)}} 
(\log p(\vect{y}_{b_i}|\theta)) - \frac{\mathrm{d} \mu_{q^{(i-1)}}}{\mathrm{d}\eta_{q^{(i-1)}}} \eta^{(i-1)}_{b_i},\\
\frac{\mathrm{d}\mathrm{KL} \left( q(\theta) ~\|~ q^{(i-1)}(\theta) \right) }{\mathrm{d}\eta_q}  &= \frac{\mathrm{d} \mu_{q}}{\mathrm{d}\eta_{q}} \left ( \eta_{q} - \eta_{q}^{(i-1)} \right) \approx  \frac{\mathrm{d} \mu_{q^{(i-1)}}}{\mathrm{d}\eta_{q^{(i-1)}}} \left ( \eta_{q} - \eta_{q}^{(i-1)} \right) .
\end{align*}
In the second line above the approximation results from the assumption of small parameter change $\Delta \eta_q^{(i)}$ (or alternatively local constancy of the Fisher information). Equating the derivatives to zero and rearranging recovers the fixed point equations.

The equivalence of the {\it KL trust region method} is now simple to show as the associated Lagrangian, $\mathcal{L}(\eta_q) = \nabla_{\eta} \mathcal{F}^{(i)}(q^{(i-1)}(\theta)) \eta_q  - \frac{1}{\rho} \left( \mathrm{KL} \left( q(\theta) ~\|~ q^{(i-1)}(\theta) \right) - \gamma \right)$, is the {\it proximal method} up to an additive constant.

The {\it KL$^s$ trust region method} can also be rewritten as a Lagrangian $\mathcal{L}(\eta_q) = \nabla_{\eta} \mathcal{F}^{(i)}(q^{(i-1)}(\theta)) \eta_q  - \frac{1}{\rho} \left( \mathrm{KL}^{\text{s}} \left( q(\theta) ~\|~ q^{(i-1)}(\theta) \right) - \gamma \right)$. For small changes in the approximate posterior natural parameters $\Delta \eta_q^{(i)} =  \eta_q^{(i)}-\eta_q^{(i-1)}$, the symmetrized KL can be approximated using a second order Taylor expansion, 
\begin{align}
\mathrm{KL}^{\text{s}} \left( q(\theta) ~\|~ q^{(i-1)}(\theta) \right) \approx \frac{1}{2} \left ( \eta_{q} - \eta_{q}^{(i-1)} \right)^{\top} \frac{\mathrm{d} \mu_{q^{(i-1)}}}{\mathrm{d}\eta_{q^{(i-1)}}} \left ( \eta_{q} - \eta_{q}^{(i-1)} \right).
\end{align}
This is the same form as the exclusive KL takes, the inclusive and exclusive KL divergences being locally identical around their optima. Taking derivatives of the Lagrangian and setting them to zero recovers the fixed point equations again. 

The {\it mirror descent} method can be shown to yield the fixed points by noting that 
\begin{align*}
\nabla_{\mu} \mathcal{F}^{(i)}(q^{(i-1)}(\theta))  &= \frac{\mathrm{d}}{\mathrm{d}\mu_{q^{(i-1)}}} \mathbb{E}_{q^{(i-1)}} (\log p(\vect{y}_{b_i}|\theta)) -  \eta_{b_i},\\
\frac{\mathrm{d}\mathrm{KL} \left( q^{(i-1)}(\theta) ~\|~ q(\theta) \right) }{\mathrm{d}\mu_q}  &=  \eta_{q} - \eta_{q}^{(i-1)} .
\end{align*}
Differentiating the mirror descent objective and substituting these results in recovers usual update. The last result above can be found using convex duality. For a full derivation and more information on the relationship between mirror descent and natural gradients see \cite{raskutti+mukherjee:2015}.

It is also possible to define optimization approaches analogous to the above that do not linearize the free-energy term and instead perform potentially multiple updates of the nested non-linear optimization problems \citep{theis+hoffman:2015, khan+al:2015, khan+al:2016}.

\subsection{The equivalence of the simplified fixed-point updates and Power EP $\alpha \rightarrow 0$: Proof of property \ref{prop:pep-fp}}
\label{proof:pep-fp}
Assume PEP is using unnormalized distributions throughout:
\begin{enumerate}
\item $p^{(i)}_{\alpha}(\theta) = q^{(i-1)}(\theta) \left( \frac{p(\mathbf{y}_{b_i} | \theta)}{t_{b_i}^{(i-1)}(\theta)}\right)^{\alpha} $ \textrm{\% form tilted}
\item $q_{\alpha}(\theta) = \mathrm{proj}( p^{(i)}_{\alpha}(\theta) )$  \% moment match 
\item $q^{(i)}(\theta) = \left ( q^{(i-1)}(\theta) \right )^{1-1/\alpha} \left( q_{\alpha}(\theta) \right)^{1/\alpha}$ \% update posterior
\item $t^{(i)}_{b_i}(\theta) = \frac{q^{(i)}(\theta)}{q^{(i-1)}(\theta)} t^{(i-1)}_{b_i}(\theta) $ \% update approximate likelihood
\end{enumerate}
From (2,3,4), we have:
\begin{align}
\log t^{(i)}_{b_i}(\theta) = \frac{1}{\alpha} \left( \log \mathrm{proj}( p^{(i)}_{\alpha}(\theta) ) - \log q^{(i-1)}(\theta) \right) + \log t^{(i-1)}_{b_i} (\theta). \label{eq:log-ti-1}
\end{align}
Using the augmented sufficient statistics $T(\theta)$ as in \ref{sec:appen:fp} and the fact that $q_{\alpha}(\theta)$ is unnormalized, we can write:
\begin{align}
\log q_{\alpha}(\theta) = \log \mathrm{proj}( p^{(i)}_{\alpha}(\theta) ) = \eta^\intercal_{q_{\alpha}} T(\theta).
\end{align}
Let $\mu_{q_{\alpha}} = \int q_{\alpha}(\theta) T(\theta) \mathrm{d}\theta$ be the mean parameters of $q_{\alpha}(\theta)$. From the moment matching projection, $\mu_{q_{\alpha}} = \int p^{(i)}_{\alpha}(\theta) T(\theta) \mathrm{d}\theta$.
Using Taylor series to expand $\log \mathrm{proj}( p^{(i)}_{\alpha}(\theta) )$ as a function of $\alpha$ about $\alpha = 0$:
\begin{align}
\log \mathrm{proj}( p^{(i)}_{\alpha}(\theta) ) = \log \mathrm{proj}( p^{(i)}_{\alpha}(\theta) ) \big|_{\alpha=0} + \alpha \left( \frac{\mathrm{d}}{\mathrm{d} \alpha} \log \mathrm{proj}( p^{(i)}_{\alpha}(\theta) ) \big|_{\alpha=0} \right) + F(\alpha),
\end{align}
where $F(\alpha) = \sum_{t=2}^{\infty} \frac{\alpha^t}{t!} ( \frac{\mathrm{d}^t}{\mathrm{d} \alpha^t} \log \mathrm{proj}( p^{(i)}_{\alpha}(\theta) ) \big|_{\alpha=0} )$ collects all the high order terms in the expansion.
Since $\log \mathrm{proj}( p^{(i)}_{\alpha}(\theta) ) \big|_{\alpha=0} = \log q^{(i-1)}(\theta)$, the above equation becomes:
\begin{align}
\log \mathrm{proj}( p^{(i)}_{\alpha}(\theta) ) = \log q^{(i-1)}(\theta) + \alpha \left( \frac{\mathrm{d}}{\mathrm{d} \alpha} \log \mathrm{proj}( p^{(i)}_{\alpha}(\theta) ) \big|_{\alpha=0} \right) + F(\alpha). \label{eq:logproj}
\end{align}
Now consider $\frac{\mathrm{d}}{\mathrm{d} \alpha} \log \mathrm{proj}( p^{(i)}_{\alpha}(\theta) )$, we have:
\begin{align}
\frac{\mathrm{d}}{\mathrm{d} \alpha} \log \mathrm{proj}( p^{(i)}_{\alpha}(\theta) ) 
&= T(\theta)^\intercal \frac{\mathrm{d} \eta_{q_{\alpha}}}{\mathrm{d} \alpha} = T(\theta)^\intercal \frac{\mathrm{d} \eta_{q_{\alpha}}}{\mathrm{d} \mu_{q_{\alpha}}} \frac{\mathrm{d} \mu_{q_{\alpha}}}{\mathrm{d} \alpha} \\
&= T(\theta)^\intercal \frac{\mathrm{d} \eta_{q_{\alpha}}}{\mathrm{d} \mu_{q_{\alpha}}} \frac{\mathrm{d}}{\mathrm{d} \alpha} \int p^{(i)}_{\alpha}(\theta) T(\theta) \mathrm{d}\theta \\
&= T(\theta)^\intercal \frac{\mathrm{d} \eta_{q_{\alpha}}}{\mathrm{d} \mu_{q_{\alpha}}} \frac{\mathrm{d}}{\mathrm{d} \alpha} \int q^{(i-1)}(\theta) \left( \frac{p(\mathbf{y}_{b_i} | \theta)}{t_{b_i}^{(i-1)}(\theta)}\right)^{\alpha} T(\theta) \mathrm{d}\theta \\
&= T(\theta)^\intercal \frac{\mathrm{d} \eta_{q_{\alpha}}}{\mathrm{d} \mu_{q_{\alpha}}} \int q^{(i-1)}(\theta) \left( \frac{p(\mathbf{y}_{b_i} | \theta)}{t_{b_i}^{(i-1)}(\theta)} \right)^{\alpha} \log \frac{p(\mathbf{y}_{b_i} | \theta)}{t_{b_i}^{(i-1)}(\theta)} T(\theta) \mathrm{d}\theta \\
&= T(\theta)^\intercal \frac{\mathrm{d} \eta_{q_{\alpha}}}{\mathrm{d} \mu_{q_{\alpha}}} \frac{\mathrm{d}}{\mathrm{d} \eta_{q^{(i-1)}}} \int q^{(i-1)}(\theta) \left( \frac{p(\mathbf{y}_{b_i} | \theta)}{t_{b_i}^{(i-1)}(\theta)} \right)^{\alpha} \log \frac{p(\mathbf{y}_{b_i} | \theta)}{t_{b_i}^{(i-1)}(\theta)} \mathrm{d}\theta,
\end{align}
where $\eta_{q^{(i-1)}}$ is the natural parameters of $q^{(i-1)}$.

Thus,
\begin{align}
\frac{\mathrm{d}}{\mathrm{d} \alpha} \log \mathrm{proj}( p^{(i)}_{\alpha}(\theta) ) \big|_{\alpha=0} 
&= T(\theta)^\intercal \frac{\cancel{\mathrm{d} \eta_{q^{(i-1)}}}}{\mathrm{d} \mu_{q^{(i-1)}}} \frac{\mathrm{d}}{\cancel{\mathrm{d} \eta_{q^{(i-1)}}}} \int q^{(i-1)}(\theta) \log \frac{p(\mathbf{y}_{b_i} | \theta)}{t_{b_i}^{(i-1)}(\theta)} \mathrm{d}\theta \\
&= T(\theta)^\intercal \frac{\mathrm{d}}{\mathrm{d} \mu_{q^{(i-1)}}} \int q^{(i-1)}(\theta) \log \frac{p(\mathbf{y}_{b_i} | \theta)}{t_{b_i}^{(i-1)}(\theta)} \mathrm{d}\theta.
\end{align}
Plug this into \eqref{eq:logproj}, we obtain:
\begin{align}
\log \mathrm{proj}( p^{(i)}_{\alpha}(\theta) ) = \log q^{(i-1)}(\theta) + \alpha T(\theta)^\intercal \frac{\mathrm{d}}{\mathrm{d} \mu_{q^{(i-1)}}} \int q^{(i-1)}(\theta) \log \frac{p(\mathbf{y}_{b_i} | \theta)}{t_{b_i}^{(i-1)}(\theta)} \mathrm{d}\theta + F(\alpha).
\end{align}
From this equation and \eqref{eq:log-ti-1},
\begin{align}
\log t^{(i)}_{b_i}(\theta) 
&= \frac{1}{\alpha} \left( \alpha T(\theta)^\intercal \frac{\mathrm{d}}{\mathrm{d} \mu_{q^{(i-1)}}} \int q^{(i-1)}(\theta) \log \frac{p(\mathbf{y}_{b_i} | \theta)}{t_{b_i}^{(i-1)}(\theta)} \mathrm{d}\theta + F(\alpha) \right) + \log t^{(i-1)}_{b_i} (\theta) \\
&= T(\theta)^\intercal \frac{\mathrm{d}}{\mathrm{d} \mu_{q^{(i-1)}}} \int q^{(i-1)}(\theta) \log \frac{p(\mathbf{y}_{b_i} | \theta)}{t_{b_i}^{(i-1)}(\theta)} \mathrm{d}\theta + \frac{F(\alpha)}{\alpha} + \log t^{(i-1)}_{b_i} (\theta) \\
&= T(\theta)^\intercal \frac{\mathrm{d}}{\mathrm{d} \mu_{q^{(i-1)}}} \int q^{(i-1)}(\theta) \log p(\mathbf{y}_{b_i} | \theta) \mathrm{d}\theta \\
&{\hskip 5mm} - T(\theta)^\intercal \frac{\mathrm{d}}{\mathrm{d} \mu_{q^{(i-1)}}} \int q^{(i-1)}(\theta) \log t_{b_i}^{(i-1)}(\theta) \mathrm{d}\theta + \frac{F(\alpha)}{\alpha} + \log t^{(i-1)}_{b_i} (\theta). \label{eq:log-ti-2}
\end{align}
Note that:
\begin{align}
T(\theta)^\intercal \frac{\mathrm{d}}{\mathrm{d} \mu_{q^{(i-1)}}} \int q^{(i-1)}(\theta) \log t_{b_i}^{(i-1)}(\theta) \mathrm{d}\theta 
&= \log t_{b_i}^{(i-1)}(\theta) \frac{\mathrm{d}}{\mathrm{d} \mu_{q^{(i-1)}}} \int T(\theta)^\intercal q^{(i-1)}(\theta) \mathrm{d}\theta \\
&= \log t_{b_i}^{(i-1)}(\theta) \frac{\mathrm{d} \mu_{q^{(i-1)}}}{\mathrm{d} \mu_{q^{(i-1)}}} \\
&= \log t_{b_i}^{(i-1)}(\theta).
\end{align}
Hence, \eqref{eq:log-ti-2} becomes:
\begin{align}
T(\theta)^\intercal \eta^{(i)}_{b_i}
&= T(\theta)^\intercal \frac{\mathrm{d}}{\mathrm{d} \mu_{q^{(i-1)}}} \int q^{(i-1)}(\theta) \log p(\mathbf{y}_{b_i} | \theta) \mathrm{d}\theta + \frac{F(\alpha)}{\alpha},
\end{align}
which is equivalent to:
\begin{align}
\label{eq:pep-update}
T(\theta)^\intercal \eta^{(i)}_{b_i}
&= T(\theta)^\intercal \frac{\mathrm{d}}{\mathrm{d} \mu_q} \int q(\theta) \log p(\mathbf{y}_{b_i} | \theta) \mathrm{d}\theta + \frac{F(\alpha)}{\alpha}.
\end{align}
Let $\bar{q}(\theta)$ be the normalized distribution of $q(\theta)$, i.e. $q(\theta) = Z_q \bar{q}(\theta)$. The fixed point update for local VI is:
\begin{align}
\eta^{(i)}_{b_i}  &
= \frac{\mathrm{d}}{\mathrm{d}\mu_{\bar{q}}} \int \bar{q}(\theta) \log p(\vect{y}_{b_i}|\theta) \mathrm{d}\theta,
\end{align}
where $\mu_{\bar{q}} = \int T(\theta)^\intercal \bar{q}(\theta) \mathrm{d}\theta = \mu_q/Z_q$.
We have:
\begin{align}
\frac{\mathrm{d}}{\mathrm{d}\mu_{\bar{q}}} \int \bar{q}(\theta) \log p(\vect{y}_{b_i}|\theta) \mathrm{d}\theta
&= \frac{\mathrm{d}\mu_q}{\mathrm{d}\mu_{\bar{q}}} \frac{\mathrm{d}}{\mathrm{d}\mu_q} \int \frac{1}{Z_q} q(\theta) \log p(\vect{y}_{b_i}|\theta) \mathrm{d}\theta \\
&= (\cancel{Z_q} \mathbb{I}) \frac{1}{\cancel{Z_q}} \frac{\mathrm{d}}{\mathrm{d}\mu_q} \int q(\theta) \log p(\vect{y}_{b_i}|\theta) \mathrm{d}\theta \\
&= \frac{\mathrm{d}}{\mathrm{d}\mu_q} \int q(\theta) \log p(\vect{y}_{b_i}|\theta) \mathrm{d}\theta.
\end{align}
From this equation and the fact that $\frac{F(\alpha)}{\alpha} \rightarrow 0$ when $\alpha \rightarrow 0$, the fixed point update for local VI satisfies the Power-EP update in \eqref{eq:pep-update} when $\alpha \rightarrow 0$.

\section{Gradients of the free-energy with respect to hyperparameters}
\label{sec:hyper-gradients}
In this section, we derive the gradients of the global free energy wrt the hyperparameters when the local VI procedure has converged to the optimal approximate posterior (either through analytic, off-the- shelf or fixed-point optimization). We provide two derivations: First, the standard one which is specific to approximate posteriors that are in the exponential family. Second, a derivation that applies for general $q(\theta)$ which also provides more insight. 

\subsection{The standard derivation}
 As shown in the previous sections, the global free energy is as follows,
\begin{align*}
\mathcal{F}(q(\theta)) 
&= \int \mathrm{d}\theta q(\theta; \eta_q) \log \frac{p(\theta) \prod_m p(\vect{y}_m|\theta)}{q(\theta; \eta_q)} \\
&= \underbrace{\int \mathrm{d}\theta q(\theta; \eta_q) [\log p(\theta) - \log q(\theta; \eta_q)]}_{\mathcal{F}_1} + \underbrace{\sum_m \int \mathrm{d}\theta q(\theta; \eta_q) \log p(\vect{y}_m|\theta)}_{\mathcal{F}_2}.
\end{align*}
Note that,
\begin{align*}
q(\theta; \eta_q) &= \exp(\eta_q^\intercal T(\theta) - A(\eta_q)), \\
p(\theta; \eta_0) &= \exp(\eta_0^\intercal T(\theta) - A(\eta_0)), \\
\eta_q &= \eta_0 + \sum_m \eta_m.
\end{align*}
Hence,
\begin{align*}
\mathcal{F}_1 = A(\eta_q) - A(\eta_0) -  \sum_m \eta_n^\intercal \mathbb{E}_q(\theta)[T(\theta)] =  A(\eta_0) - A(\eta_q) -  \sum_m \eta_n^\intercal\frac{\mathrm{d} A(\eta_q)}{\mathrm{d} \eta_q}.
\end{align*}
Differentiating $\mathcal{F}_1$ and $\mathcal{F}_2$ wrt a hyperparameter $\epsilon$ of the model, noting that the natural gradients of the local factors and the global variational approximation both depend on $\epsilon$, gives,
\begin{align*}
\frac{\mathrm{d} \mathcal{F}_1} {\mathrm{d} \epsilon} &= \left( \frac{\mathrm{d} A(\eta_q)} {\mathrm{d} \eta_q} \right)^\intercal \frac{\mathrm{d} \eta_q} {\mathrm{d} \epsilon} - \left( \frac{\mathrm{d} A(\eta_0)} {\mathrm{d} \eta_0} \right)^\intercal \frac{\mathrm{d} \eta_0} {\mathrm{d} \epsilon} - \sum_m \left[ \left(\frac{\mathrm{d}A(\eta_q)}{\mathrm{d}\eta_q}\right)^\intercal \frac{\mathrm{d} \eta_m} {\mathrm{d} \epsilon} - \eta_m^\intercal \frac{\mathrm{d}^2A(\eta_q)}{\mathrm{d}\eta_q\mathrm{d}\eta_q} \frac{\mathrm{d} \eta_q}{\mathrm{d} \epsilon}\right]\\
\frac{\mathrm{d} \mathcal{F}_{2}} {\mathrm{d} \epsilon} &= \sum_m \left[ \frac{\partial \mathcal{F}_{2m}}{\partial\epsilon} + \left(\frac{\partial \mathcal{F}_{2m}} {\partial \eta_q}\right)^\intercal \frac{\mathrm{d} \eta_q} {\mathrm{d} \epsilon} \right]
\end{align*}
Note that 
$\eta_q = \eta_0 + \sum_{m} \eta_m$, leading to
\begin{align}
-\sum_{m} \frac{\mathrm{d} \eta_m} {\mathrm{d} \epsilon} = - \frac{\mathrm{d} \eta_q} {\mathrm{d} \epsilon} + \frac{\mathrm{d} \eta_0} {\mathrm{d} \epsilon},
\end{align}
Here,
\begin{align}
\frac{\partial \mathcal{F}_{2m}} {\partial \eta_q} = \frac{\partial} {\partial \eta_q}  \mathbb{E}_q (\log p(\vect{y}_{b_i}|\theta)).
\end{align}

and that at convergence \ref{eq:fixed-point-bi},
\begin{align}
\frac{\partial \mathcal{F}_{2m}} {\partial \eta_q} = \frac{\mathrm{d} \mu_q}{\mathrm{d} \eta_q} \eta_q = \frac{\mathrm{d}^2A(\eta_q)}{\mathrm{d}\eta_q\mathrm{d}\eta_q} \eta_m.
\end{align}
Therefore,
\begin{align}
\frac{\mathrm{d} \mathcal{F}_{\mathrm{VFE}}} {\mathrm{d} \epsilon} 
&=  \left(\frac{\mathrm{d} A(\eta_q)} {\mathrm{d} \eta_q} - \frac{\mathrm{d} A(\eta_0)} {\mathrm{d} \eta_0} \right)^\intercal \frac{\mathrm{d} \eta_0} {\mathrm{d} \epsilon} + \sum_m \frac{\partial \mathcal{F}_{2m} }{\partial\epsilon} \\
&= \left(\mu_q - \mu_0 \right)^\intercal \frac{\mathrm{d} \eta_0} {\mathrm{d} \epsilon} + \sum_m \frac{\partial \mathcal{F}_{2m} }{\partial\epsilon},
\end{align}
where $\mu_q = \frac{\mathrm{d}A(\eta_q)}{\mathrm{d}\eta_q} = \mathbb{E}_{q(\theta)}[ T(\theta)]$ are the {\it mean} parameters.

\subsection{A more general derivation}

Consider the general case where the approximate posterior comprises a product of terms that approximate the likelihoods and one that approximates the prior,
\begin{align}
q(\theta; \psi) = \prod_{n=0}^N t_n(\theta;\psi).
\end{align}
Here $\psi$ are the variational parameters (these may correspond to natural parameters. Again the scale of the approximate terms $t_n(\theta;\psi)$ will be set such that $q(\theta; \psi)$ is normalized. Note this is a more general form of approximate posterior that allows the prior to be approximated if it lies outside of the variational family $\mathcal{Q}$. If the prior lies within the variational family, the local updates will automatically set it equal to the prior recovering the treatment in the rest of the paper and meaning that the results presented here will still hold.

The global free-energy depends on the model hyperparameters through the joint distribution $p(\mathbf{y},\theta|\epsilon)$,
\begin{align}
\mathcal{F}( \epsilon, q(\theta; \psi)) = \int \mathrm{d}\theta \; q(\theta; \psi) \log \frac{p(\mathbf{y},\theta|\epsilon)}{q(\theta; \psi)} 
\end{align}
Now consider the optimal variational approximation for a fixed setting of the hyperparameters,
\begin{align}
\psi^{\text{opt}}(\epsilon) = \argmax_{\psi} \int \mathrm{d}\theta \; q(\theta; \psi) \log \frac{p(\mathbf{y},\theta|\epsilon)}{q(\theta; \psi)}.
\end{align}
The collapsed variational bound can therefore be denoted, $\mathcal{F}(\epsilon,q(\theta; \psi^{\text{opt}}(\epsilon)))$ and it is this that we will optimize to find the hyperparameters. Before we do so, note that we have been careful to represent the two distinct ways that the free-energy depends on the hyperparameters, i) through the log-joint's dependence, ii) through the optimal approximate posterior's implicit dependence via $\psi^{\text{opt}}(\epsilon))$. In fact we can decouple these two contributions and consider evaluating the free-energy when the hyperparameters differ, $\mathcal{F}(\epsilon, q(\theta; \psi^{\text{opt}}(\epsilon')))$, the collapsed bound being recovered when $\epsilon' =  \epsilon$.

We are now in a position to compute derivatives of the collapsed free-energy using the insight above to split this into two terms,

\begin{align}
\frac{\mathrm{d}}{ \mathrm{d} \epsilon} \mathcal{F}(\epsilon,q(\theta;  \psi^{\text{opt}}(\epsilon))) =& 
\frac{\mathrm{d}}{ \mathrm{d} \epsilon} \mathcal{F}(\epsilon, q(\theta;  \psi^{\text{opt}}(\epsilon'))) \biggr\vert_{\epsilon'=\epsilon} 
+ \frac{\mathrm{d}}{ \mathrm{d} \epsilon'} \mathcal{F}(\epsilon, q(\theta; \psi^{\text{opt}}(\epsilon'))) \biggr\vert_{\epsilon'=\epsilon}.
\end{align}
We now consider these two terms: First, the dependence through the log-joint distribution,
\begin{align}
& \frac{\mathrm{d}}{ \mathrm{d} \epsilon} \mathcal{F}(\epsilon, q(\theta; \psi^{\text{opt}}(\epsilon'))) \biggr\vert_{\epsilon'=\epsilon}  = \frac{\mathrm{d}}{ \mathrm{d} \epsilon} \int \mathrm{d}\theta \; q(\theta;\psi(\epsilon')) \log p(\mathbf{y},\theta|\epsilon) \biggr\vert_{\epsilon'=\epsilon} \nonumber \\
& \quad = \sum_{m=1}^M \frac{\mathrm{d}}{ \mathrm{d} \epsilon} \int \mathrm{d}\theta \; q(\theta; \psi(\epsilon')) \log p(\mathbf{y}_m|\theta,\epsilon) \biggr\vert_{\epsilon'=\epsilon} +   \int \mathrm{d}\theta \; q(\theta; \psi(\epsilon')) \frac{\mathrm{d}}{ \mathrm{d} \epsilon} \log p(\theta|\epsilon) \biggr\vert_{\epsilon'=\epsilon}\nonumber\\
&= \sum_{m=1}^M \mathbb{E}_{q(\theta)} \left [\frac{\mathrm{d}}{ \mathrm{d} \epsilon} \log p(\mathbf{y}_m|\theta,\epsilon) \right] + \mathbb{E}_{q(\theta)} \left[\frac{\mathrm{d}}{ \mathrm{d} \epsilon} \log p(\theta|\epsilon) \right].
\end{align}
Second, the dependence through the optimal approximate posterior's implicit dependence on $\epsilon$
\begin{align}
\frac{\mathrm{d}}{ \mathrm{d} \epsilon'} \mathcal{F}(\epsilon, q(\theta;  \psi^{\text{opt}}(\epsilon'))) \biggr\vert_{\epsilon'=\epsilon} = \frac{\mathrm{d} \psi^{\text{opt}}(\epsilon')}{\mathrm{d} \epsilon'} \frac{\mathrm{d}}{ \mathrm{d} \psi}  \mathcal{F}(\epsilon, q(\theta; \psi) \biggr\vert_{\epsilon'=\epsilon, \psi = \psi^{\text{opt}}(\epsilon')} = 0 .
\end{align}
Here we have substituted in the fact that were are at the collapsed bound and so the derivative wrt $\psi$ is zero.

So the term that arises from the dependence of the approximate posterior on the hyperparameters (terms 2) vanishes meaning the only contribution comes from the first term. This is precisely the same term that would remain if we were to perform coordinate ascent (since then when updating the hyperparameters the approximate posterior would have been fixed). 
\begin{align}
\frac{\mathrm{d}}{ \mathrm{d} \epsilon} \mathcal{F}(\epsilon,q(\theta;  \psi^{\text{opt}}(\epsilon))) =& \sum_{m=1}^M \mathbb{E}_{q(\theta;  \psi^{\text{opt}}(\epsilon))} \left [\frac{\mathrm{d}}{ \mathrm{d} \epsilon} \log p(\mathbf{y}_m|\theta,\epsilon) \right] + \mathbb{E}_{q(\theta;  \psi^{\text{opt}}(\epsilon))} \left[\frac{\mathrm{d}}{ \mathrm{d} \epsilon} \log p(\theta|\epsilon) \right].
\end{align}
When the prior distribution is in the exponential family, the second term above becomes
\begin{align}
\mathbb{E}_{q(\theta; \psi^{\text{opt}}(\epsilon))} \left[\frac{\mathrm{d}}{ \mathrm{d} \epsilon} \log p(\theta|\epsilon) \right] = \left(\mu_q - \mu_0 \right)^\intercal \frac{\mathrm{d} \eta_0} {\mathrm{d} \epsilon}.
\end{align}
This recovers the expression in the previous section, although we have not assumed the approximate posterior is in the exponential family (here $\mu_q $ and $\mu_0$ are the average of the prior's sufficient statistics under the approximate posterior and the prior respectively).

\begin{figure}[!ht]
\begin{center}
\includegraphics[width=8cm]{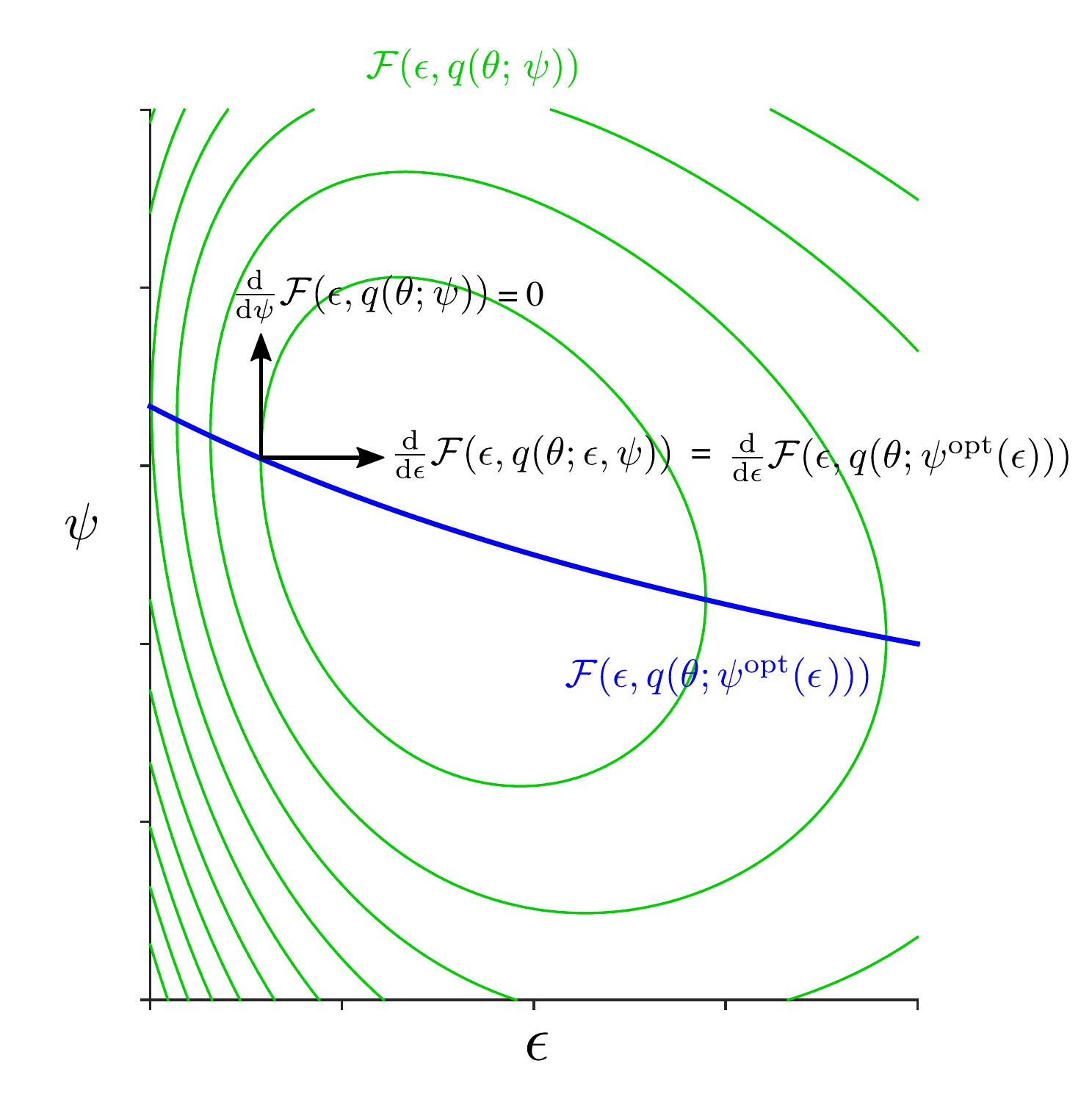}
\end{center}
\caption{Contours of the free-energy $\mathcal{F}( \epsilon, q(\theta;\psi))$ are shown in green as a function of the hyperparameters $\epsilon$ and the variational parameters of the approximate posterior $\phi$. The collapsed bound $\mathcal{F}(\epsilon, q(\theta; \psi^{\text{opt}}(\epsilon)))$  is shown in blue. The gradients of the free-energy with respect to the variational parameters are zero along the collapsed bound $\frac{\mathrm{d}}{ \mathrm{d} \psi} \mathcal{F}( \epsilon, q(\theta; \psi)) \vert_{\psi = \psi^{\text{opt}}} = 0$, by definition. This means that the gradients of the collapsed free-energy as a function of the hyperparameters are equal to those of the free-energy itself, $\frac{\mathrm{d}}{ \mathrm{d} \epsilon} \mathcal{F}(\epsilon,q(\theta;  \psi)) = \frac{\mathrm{d}}{ \mathrm{d} \epsilon} \mathcal{F}(\epsilon,q(\theta;  \psi^{\text{opt}}(\epsilon)))$.
 }\label{fig:free-energy-schem}
\end{figure}

Figure \ref{fig:free-energy-schem} provides some intuition for these results. Note that in the case where the approximating family includes the true posterior distribution, the collapsed bound is equal to the log-likelihood of the hyperparameters. So, the result shows that the gradient of the log-likelihood wrt the hyperparameters is equal to the gradient of the free-energy wrt the hyperparameters, treating $q$ as fixed. Often this is computed in the M-step of variational EM, but it is used in coordinate ascent, which can be slow to converge. Instead, this gradient can be passed to an optimizer to perform direct gradient-based optimization of the log-likelihood.

\section{Streaming Gaussian process regression and classification\label{sec:appen:sgp}}
\subsection{Online variational free-energy approach using shared pseudo-points with approximate maximum likelihood learning for the hyperparameters\label{sec:online_vfe_gp_shared}}
We consider a variational inference scheme with an approximation posterior based on pseudo-points and that all streaming batches or groups of data points touch the same set of pseudo-points, that is, 
\begin{align}
    p(f | \yvec) \propto p(f) \prod_{r=1}^{R} p(\yvec_r|f)  \approx p(f) \prod_{r=1}^{R} t_r(\uvec) \propto p(f_{\ne \uvec} | \uvec) q(\uvec) = q(f),
\end{align}
where $R$ is the number of batches considered and $t_r(\uvec)$ is the approximate contribution of the $r$-th batch to the posterior. This is the standard set up considered for sparse GPs in the literature, see e.g.~\cite{hensman+al:13,bui+al:2017b}. We next detail the specifics for the streaming settings \citep{bui+al:2017}, when we allow the pseudo-points to move and adjust the hyperparameters as new data arrive.

Let $\avec = f(\zvec_{\mathrm{old}})$ and $\bvec = f(\zvec_{\mathrm{new}})$ be the pseudo-outputs or inducing points before and after seeing new data, where $\zvec_{\mathrm{old}}$ and $\zvec_{\mathrm{new}}$ are the pseudo-inputs accordingly. Note that extra pseudo-points can be added or conversely, old pseudo-points can be removed, i.e.~the cardinalities of $\avec$ and $\bvec$ do not need to be the same.
The previous posterior, $q_\mathrm{old}(f) = p(f_{\ne \avec}|\avec, \theta_\mathrm{old}) q(\avec)$, can be used to find the approximate likelihood given by old observations as follows,
\begin{align}
	p(\yvec_\mathrm{old}|f) \approx \frac{q_\mathrm{old}(f) p(\yvec_\mathrm{old}|\theta_\mathrm{old})}{p(f|\theta_\mathrm{old})} \quad\mathrm{as}\quad q_\mathrm{old}(f) \approx \frac{p(f|\theta_\mathrm{old}) p(\yvec_\mathrm{old}|f)} {p(\yvec_\mathrm{old}|\theta_\mathrm{old})}.\label{eqn:approx_lik}
\end{align}
Note that we have made the dependence of the hyperparameters explicit, as these will be optimized, together with the variational parameters, using the variational free-energy. Substituting the approximate likelihood above into the posterior that we want to target gives us:
\begin{align}
p(f|\yvec_\mathrm{old},\yvec_\mathrm{new}) = \frac{p(f|\theta_\mathrm{new}) p(\yvec_\mathrm{old}|f) p(\yvec_\mathrm{new}|f)} {p(\yvec_\mathrm{new}, \yvec_\mathrm{old}|\theta_\mathrm{new})} \approx \frac{p(f|\theta_\mathrm{new}) q_\mathrm{old}(f) p(\yvec_\mathrm{old}|\theta_\mathrm{old}) p(\yvec_\mathrm{new}|f)} {p(f|\theta_\mathrm{old}) p(\yvec_\mathrm{new}, \yvec_\mathrm{old}|\theta_\mathrm{new})}. \label{eqn:running_posterior} 
\end{align}

The new posterior approximation takes the same form as with the previous posterior, but with the new pseudo-points and new hyperparameters: $q_\mathrm{new}(f) = p(f_{\ne \bvec}|\bvec, \theta_\mathrm{new}) q(\bvec)$. This approximate posterior can be obtained by minimizing the KL divergence,
\begin{align}
\mathrm{KL} \lbrack q_\mathrm{new}(f) || \hat{p}(f|\yvec_{\mathrm{old}, \mathrm{new}}) \rbrack 
&= \int \dd f q_\mathrm{new}(f) \log \frac{p(f_{\ne \bvec}|\bvec, \theta_\mathrm{new}) q_\mathrm{new}(\bvec)}{\frac{p(\yvec_\mathrm{old}|\theta_\mathrm{old})}{p(\yvec_\mathrm{new}, \yvec_\mathrm{old}|\theta_\mathrm{new})} p(f|\theta_\mathrm{new}) p(\yvec_\mathrm{new}|f) \frac{q_\mathrm{old}(f)}{p(f|\theta_\mathrm{old})}} \label{eqn:kl}\\
&= \log\frac{\mathcal{Z}_2(\theta_\mathrm{new})}{\mathcal{Z}_1(\theta_\mathrm{old})} + \int \dd f q_\mathrm{new}(f) \left[ \log \frac{p(\avec|\theta_\mathrm{old})q_\mathrm{new}(\bvec)}{p(\bvec|\theta_\mathrm{new}) q_\mathrm{old}(\avec) p(\yvec_\mathrm{new}|f)} \right]. \label{eq:kl-simplified}
\end{align}
The last equation above is obtained by noting that $p(f|\theta_\mathrm{new})/p(f_{\ne \bvec}|\bvec, \theta_\mathrm{new}) = p(\bvec|\theta_\mathrm{new})$ and
\begin{align}
\frac{q_\mathrm{old}(f)}{p(f|\theta_\mathrm{old})} = \frac{\bcancel{p(f_{\ne \avec}|\avec, \theta_\mathrm{old})} q_\mathrm{old}(\avec)}{\bcancel{p(f_{\ne \avec}|\avec, \theta_\mathrm{old})} p(\avec|\theta_\mathrm{old})} = \frac{q_\mathrm{old}(\avec)}{p(\avec|\theta_\mathrm{old})}.\nonumber
\end{align}

Since the KL divergence is non-negative, the second term in \eqref{eq:kl-simplified} is the negative lower bound of the approximate online log marginal likelihood\footnote{Note that this is only an approximation, as the hyperparameters are adjusted as new data arrive.}, or the variational free energy, $\mathcal{F}(q_\mathrm{new}(f))$. We can decompose the bound as follows,
\begin{align}
\mathcal{F}(q_\mathrm{new}(f), \theta_\mathrm{new}) 
	&= \int \dd f q_\mathrm{new}(f) \left[ \log \frac{p(\avec|\theta_\mathrm{old})q_\mathrm{new}(\bvec)}{p(\bvec|\theta_\mathrm{new}) q_\mathrm{old}(\avec) p(\yvec_\mathrm{new}|f)} \right] \nonumber \\
	&= \mathrm{KL}(q(\bvec) || p(\bvec|\theta_\mathrm{new})) - \int \dd f q_\mathrm{new}(f) \log p(\yvec_\mathrm{new}|f) \nonumber \\ &\quad\quad - \int \dd \avec q_\mathrm{new}(\avec) \log \frac{q_\mathrm{old}(\avec)}{p(\avec|\theta_\mathrm{old})} \nonumber.
\end{align}
The first two terms form the variational free-energy as if the current batch is the whole training data, and the last term constrains the posterior to take into account the old likelihood (through the approximate posterior and the prior).

\subsection{Online variational free-energy approach using private pseudo-points with approximate maximum likelihood learning for the hyperparameters\label{sec:appen:sgp_ml_private}}
Instead of using a common set of pseudo-points for all data points or streaming batches, we can assign separate pseudo-points to each batch of data points as follows,
\begin{align}
    p(f | \yvec) \propto p(f) \prod_{r=1}^{R} p(\yvec_r|f)  \approx p(f) \prod_{r=1}^{R} t_r(\uvec_r) \propto p(f_{\ne \uvec} | \uvec) q(\uvec) = q(f),
\end{align}
where $\uvec_r$ are the pseudo-points private to the $r$-th batch. As new data arrives, new pseudo-points will be added to summarize the new data, and the old pseudo-points, corresponding to the previously seen batches, will remain unchanged. This means we only need to add and adjust new pseudo-points and the new likelihood approximation for the new data points, as opposed to all pseudo-points and all corresponding likelihood approximations as done in the previous section.

Similar to the online learning the previous section, we will try to approximate the running posterior in \cref{eqn:running_posterior},
\begin{align}
p(f|\yvec_\mathrm{old},\yvec_\mathrm{new}) \approx \frac{p(f|\theta_\mathrm{new}) q_\mathrm{old}(f)  p(\yvec_\mathrm{new}|f)} {p(f|\theta_\mathrm{old}) } \frac{p(\yvec_\mathrm{old}|\theta_\mathrm{old})}{p(\yvec_\mathrm{new}, \yvec_\mathrm{old}|\theta_\mathrm{new})}, \label{eqn:running_post_b}
\end{align}
where 
\begin{align} 
    p(f|\theta_\mathrm{old}) &= p(f_{\ne \avec}|\avec, \theta_{\mathrm{old}}) p(\avec|\theta_\mathrm{old}),\nonumber\\
    q_\mathrm{old}(f) &= p(f_{\ne \avec}|\avec, \theta_{\mathrm{old}}) q(\avec),\nonumber
\end{align}
and $\avec$ represents all pseudo-points used for previous batches. Let $\bvec$ be the new pseudo-points for the new data and $t_b(\bvec)$ be the contribution of the new data points $\yvec_{\mathrm{new}}$ towards the posterior. The new approximate posterior is assumed to take the following form,
\begin{align}
    q_\mathrm{new}(f) 
        &\propto p(f_{\ne \avec}|\avec, \theta_{\mathrm{new}}) q(\avec) t_b(\bvec) \nonumber \\
        &= p(f_{\ne \avec, \bvec }|\avec, \bvec, \theta_{\mathrm{new}}) q(\avec) p(\bvec|\avec, \theta_{\mathrm{new}}) t_b(\bvec),\nonumber\\
        &= p(f_{\ne \avec, \bvec }|\avec, \bvec, \theta_{\mathrm{new}}) q(\avec) q(\bvec|\avec), \label{eqn:approx_post_private}
\end{align}
where we have chosen $q(\bvec|\avec) \propto p(\bvec|\avec, \theta_{\mathrm{new}}) t_b(\bvec)$ and made the dependence on the hyperparameters $\theta_{\mathrm{new}}$ implicit. Note that $q(a)$ is the variational distribution over the previous pseudo-points, and such, we only need to parameterize and learn the conditional distribution $q(\bvec|\avec)$.

Similar to the previous section, writing down the KL divergence from the running posterior in \cref{eqn:running_post_b} to the approximate posterior in \cref{eqn:approx_post_private}, and ignoring constant terms result in the online variational free-energy as follows,
\begin{align}
\mathcal{F}( q_\mathrm{new}(f), \theta_\mathrm{new})
&= \int \dd f q_\mathrm{new}(f) \log \frac{p(f|\theta_\mathrm{new}) q_\mathrm{old}(f)  p(\yvec_\mathrm{new}|f)} {p(f|\theta_\mathrm{old}) q_\mathrm{new}(f)}.
\end{align}
Note that,
\begin{align}
    \frac{q_\mathrm{old}(f) } {p(f|\theta_\mathrm{old})} &= \frac{ \bcancel{p(f_{\ne \avec}|\avec, \theta_{\mathrm{old}})} q(\avec) }{ \bcancel{p(f_{\ne \avec}|\avec, \theta_{\mathrm{old}})} p(\avec|\theta_\mathrm{old})} = \frac{ q(\avec) }{ p(\avec|\theta_\mathrm{old})},\\
    \frac{p(f|\theta_\mathrm{new})} {q_\mathrm{new}(f)} &= \frac{\bcancel{p(f_{\ne \avec}|\avec, \theta_{\mathrm{old}}) p(\avec, \bvec | \theta_\mathrm{new})}} {\bcancel{p(f_{\ne \avec}|\avec, \theta_{\mathrm{old}})} q(\avec, \bvec)} = \frac{p(\avec, \bvec | \theta_\mathrm{new})}{q(\avec, \bvec)}.
\end{align}
This leads to,
\begin{align}
\mathcal{F}(q_\mathrm{new}(f), \theta_\mathrm{new})
&= - \mathrm{KL}[q(\avec, \bvec) || p(\avec, \bvec | \theta_\mathrm{new})] + \int \dd f q_\mathrm{new}(f) \log p(\yvec_\mathrm{new}|f) \nonumber \\ &\quad\quad- \mathrm{H}[q(\avec))] + \int \dd \avec q(\avec) \log p(\avec|\theta_\mathrm{\theta}).
\end{align}
Note again that we are only optimizing the variational parameters of $q(\bvec|\avec)$ and the hyperparameters, and keeping $q(\avec)$ fixed.

\subsection{Online variational free-energy approach for both hyperparameters and the latent function with shared pseudo-points}
The variational approaches above, while maintaining an online distributional approximation for the latent function, only retain a point estimate of the hyperparameters. Imagine having observed the first batch of data points in a regression task and trained the model on this batch, and that the second batch contains only one data point. In this case, maximum likelihood learning of the hyperparameters will tend to give very large observation noise, i.e.~the noise is used to solely explain the new data and the latent function is largely ignored. Using the new model with the newly obtained hyperparameters will thus result in poor predictions on previously seen data points.

We attempt to address this issue by maintaining a distributional approximation for the hyperparameters, as well as one for the latent function, and adjusting these approximations using variational inference as new data arrive. In particular, extending \cref{sec:online_vfe_gp_shared} by introducing a variational approximation over the hyperparameters gives,
\begin{align}
    &\text{old approx.~posterior:} \quad q_\mathrm{old}(f, \theta) = p(f_{\ne \avec}|\avec, \theta) q(\avec) q_\mathrm{old}(\theta), \nonumber\\
    &\text{new approx.~posterior:} \quad q_\mathrm{new}(f, \theta) = p(f_{\ne \bvec}|\bvec, \theta) q(\bvec) q_\mathrm{new}(\theta).\nonumber
\end{align}
The likelihood of previously seen data points can be approximated via the approximate posterior as follows,
\begin{align}
    p(\yvec_\mathrm{old}|f, \theta) \approx \frac{q_\mathrm{old}(f, \theta) p(\yvec_\mathrm{old})}{p(f|\theta) p(\theta)}.\nonumber
\end{align}
Similar to the previous section, the online variational free-energy can be obtained by applying Jensen's inequality to the online log marginal likelihood, or by writing down the KL divergence as follows,
\begin{align}
    \mathrm{KL} \lbrack q_\mathrm{new}(f, \theta) || p(f, \theta|\yvec_{\mathrm{all}}) \rbrack 
    &= \int \dd f \dd \theta q_\mathrm{new}(f, \theta) \log \frac{p(f_{\ne \bvec}|\bvec, \theta) q_\mathrm{new}(\bvec) q_\mathrm{new}(\theta) p(\yvec_\mathrm{new}, \yvec_\mathrm{old}) }{ p(f|\theta) p(\theta) p(\yvec_\mathrm{new}|f, \theta) p(\yvec_\mathrm{old}|f, \theta)} \nonumber\\
    &= \log p(\yvec_\mathrm{new} | \yvec_\mathrm{old}) + \mathcal{F}(q_\mathrm{new}(f, \theta))\nonumber
\end{align}
where the online variational free-energy is,
\begin{align} 
    \mathcal{F}(q_\mathrm{new}(f, \theta)) &= \int \dd f \dd \theta q_\mathrm{new}(f, \theta) \log \frac{p(f_{\ne \bvec}|\bvec, \theta) q(\bvec) q_\mathrm{new}(\theta) }{ p(f_{\ne \avec}|\avec, \theta) q(\avec) q_\mathrm{old}(\theta) p(\yvec_\mathrm{new}|f, \theta)}\nonumber\\
    &= \int \dd f \dd \theta q_\mathrm{new}(f, \theta) \log \frac{p(\avec, \theta) q(\bvec) q_\mathrm{new}(\theta) }{ p(\bvec, \theta) q(\avec) q_\mathrm{old}(\theta) p(\yvec_\mathrm{new}|f, \theta)}\nonumber\\
    &= \mathrm{KL}[q_\mathrm{new}(\theta) || q_\mathrm{old}(\theta)] + \int \dd \theta q(\theta) \left( \mathrm{KL}[q(\bvec) || p(\bvec|\theta)] \right) \nonumber\\ & + \int \dd \theta q(\theta) q_\mathrm{new}(\avec|\theta) \log \frac{p(\avec|\theta)}{q(\avec)} - \int \dd f \dd \theta q_\mathrm{new}(f, \theta) \log p(\yvec_\mathrm{new}|f, \theta)\nonumber.
\end{align}
Most terms in the variational free-energy above requires computing an expectation wrt the variational approximation $q(\theta)$, which is not available in closed-form even when $q(\theta)$ takes a simple form such as a diagonal Gaussian. However, these expectations can be approximated by simple Monte Carlo with the {\it reparameterization trick} \citep{kingma+welling:2014,rezende+al:2014}. As in previous section, all other expectations can be handled tractably, either in closed-form or by using Gaussian quadrature. 

\subsection{Online variational free-energy approach for both hyperparameters and the latent function with private pseudo-points}

As in \cref{sec:appen:sgp_ml_private}, new pseudo-points can be allocated to new data as they arrive, and the current pseudo-points and their marginal variational approximation will remaine fixed. The corresponding variational approximation for both the latent function and hyperparameters are:
\begin{align}
    &\text{old approx.~posterior:} \quad q_\mathrm{old}(f, \theta) = p(f_{\ne \avec}|\avec, \theta) q(\avec) q_\mathrm{old}(\theta), \nonumber\\
    &\text{new approx.~posterior:} \quad q_\mathrm{new}(f, \theta) = p(f_{\ne \avec, \bvec}|\avec, \bvec, \theta) q(\avec) q(\bvec|\avec) q_\mathrm{new}(\theta).\nonumber
\end{align}

The new approximate posterior above can be derived by approximating the likelihood factor of the new data in the running posterior as follows,
\begin{align}
     \hat{p}(f, \theta | \yvec) 
        &\propto q_\mathrm{old}(f, \theta) p(\yvec_{\mathrm{new}}|f, \theta) \nonumber \\ 
        &= p(f_{\ne \avec}|\avec, \theta) q(\avec) q_\mathrm{old}(\theta) p(\yvec_{\mathrm{new}}|f, \theta) \nonumber\\
        &= p(f_{\ne \avec, \bvec}|\avec, \theta) q(\avec) q_\mathrm{old}(\theta) \textcolor{red}{p(\bvec|\avec, \bvec, \theta)}  \textcolor{blue}{p(\yvec_{\mathrm{new}}|f, \theta)} \nonumber\\
        &\approx p(f_{\ne \avec, \bvec}|\avec, \bvec, \theta) q(\avec) q_\mathrm{old}(\theta) \textcolor{red}{t_1(\bvec|\avec) t_2(\theta)}  \textcolor{blue}{t_3(\bvec) t_4(\theta)} \nonumber,
\end{align}
where $\{t_i\}_{i=1}^4$ are the approximate factors representing the contribution of the conditional prior and the likelihood to the running posterior. In other words, $q(\bvec|\avec) \propto \textcolor{red}{t_1(\bvec|\avec)}  \textcolor{blue}{t_3(\bvec)}$ and $q_\mathrm{new}(\theta) \propto q_\mathrm{old}(\theta) \textcolor{red}{t_2(\theta)} \textcolor{blue}{t_4(\theta)}$. Substituting the above variational approximation to the online variational free-energy gives us,
\begin{align} 
    \mathcal{F}(q_\mathrm{new}(f, \theta)) &= \int \dd f \dd \theta q_\mathrm{new}(f, \theta) \log \frac{p(f_{\ne \avec, \bvec}|\avec, \bvec, \theta) q(\avec) q(\bvec|\avec) q_\mathrm{new}(\theta) }{ p(f_{\ne \avec}|\avec, \theta) q(\avec) q_\mathrm{old}(\theta) p(\yvec_\mathrm{new}|f, \theta)}\nonumber\\
    &= \mathrm{KL}[q_\mathrm{new}(\theta) || q_\mathrm{old}(\theta)] + \int \dd \theta q(\theta) \left( \mathrm{KL}[q(\avec, \bvec) || p(\avec, \bvec|\theta)] \right) \nonumber\\ & - \int \dd \theta q(\theta) \left( \mathrm{KL}[q(\avec) || p(\avec|\theta)] \right) - \int \dd f \dd \theta q_\mathrm{new}(f, \theta) \log p(\yvec_\mathrm{new}|f, \theta).
\end{align}
Similar to the previous section, all terms the free-energy above can be tractably handled in closed-form or by using simple Monte Carlo with the {\it reparameterization trick} \citep{kingma+welling:2014,rezende+al:2014}.

\section{Extra results for streaming Gaussian process regression and classification experiments}

\subsection{Binary classification on a toy 2D data set}
\begin{figure}[!htb]
	\centering
	\includegraphics[width=\textwidth]{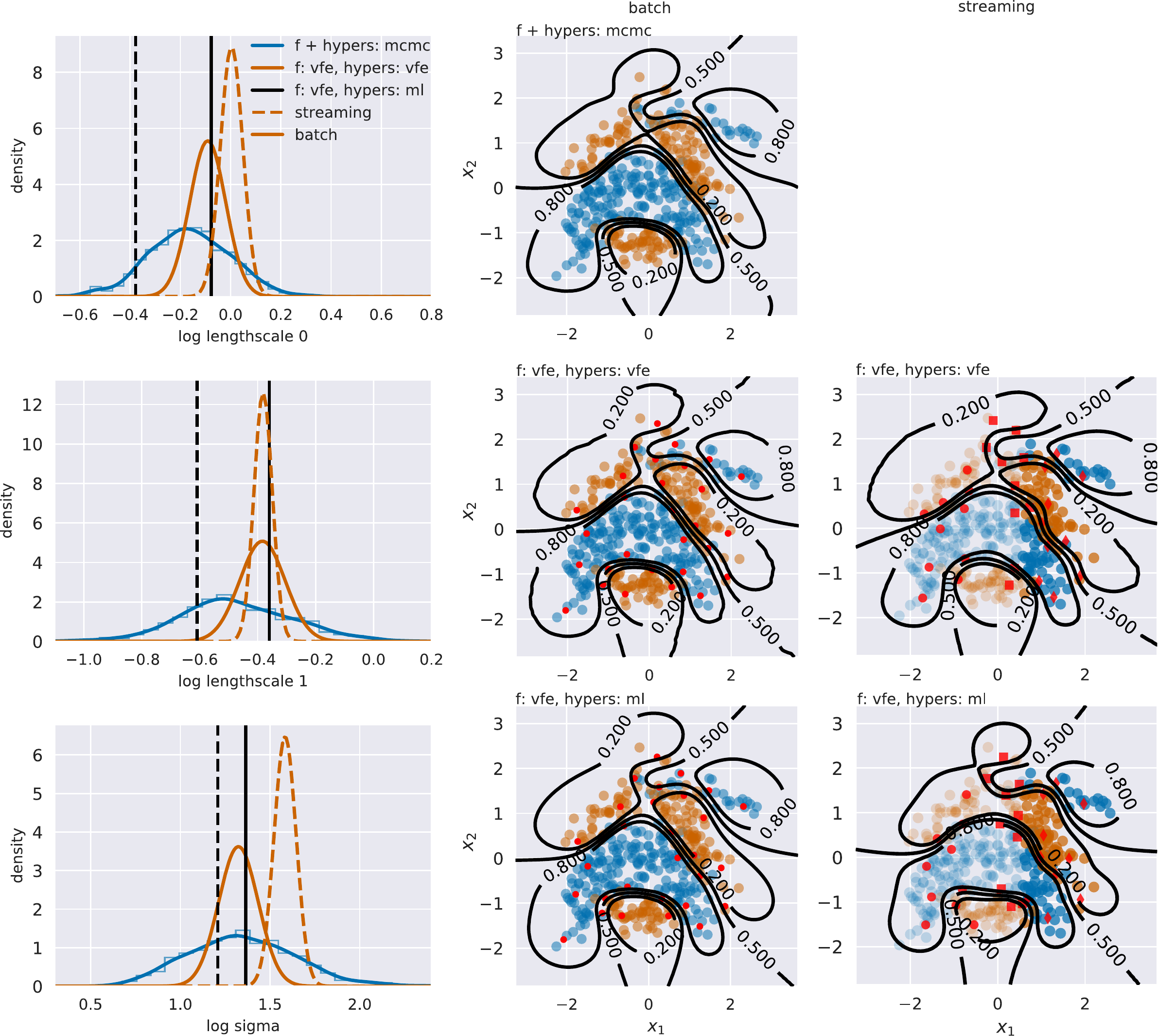}
	\caption{Results of the streaming GP experiment on a toy classification data set: the performance of several batch and streaming methods after seeing all training points. In the batch case, we consider three inference methods: MCMC for both the latent variable and the hyperparameters, VI for both the latent function and the hyperparameters, and VI for the latent function and approximate maximum likelihood learning for the hyperparameters. The two latter methods are also tested in the streaming settings. We show the predictions made by the methods after training in the batch case, and after seeing all three batches in the streaming case. The (distributional) hyperparameter estimates are also included. Best viewed in colour.\label{fig:app:gp_toy_batch_vs_streaming}}
\end{figure}

\begin{figure}[!htb]
\centering
\includegraphics[width=\textwidth]{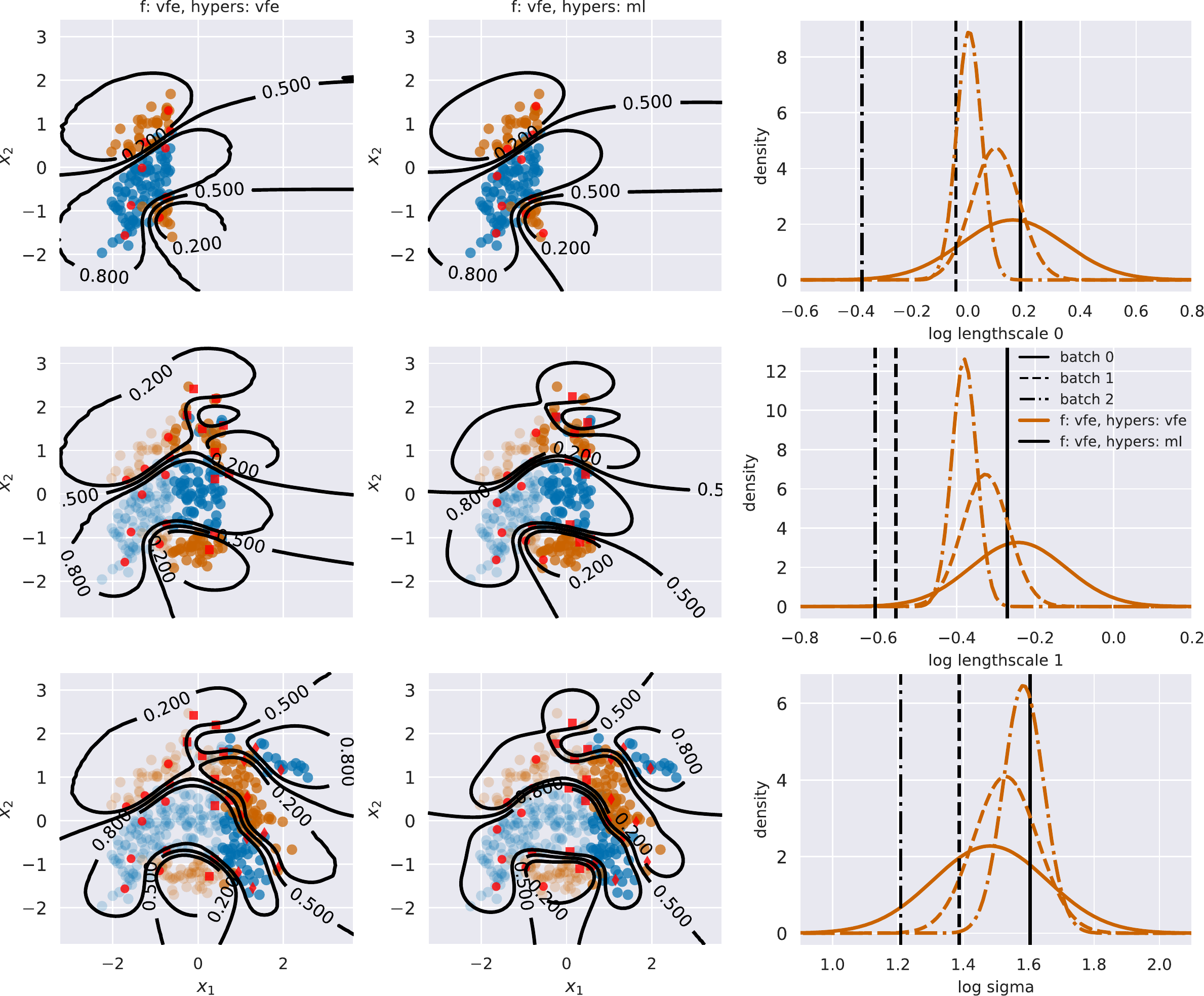}
\caption{Results of the streaming GP experiment on a toy classification data set: the performance of the streaming methods after seeing each data batch. Two methods were considered: VI for both the latent function and the hyperparameters, and VI for the latent function and approximate maximum likelihood learning for the hyperparameters. We show the predictions made by the methods after seeing each data batch and the corresponding (distributional) hyperparameter estimates. Best viewed in colour. \label{fig:app:gp_toy_streaming_1}}
\end{figure}

\begin{figure}[!htb]
\centering
\includegraphics[width=\textwidth]{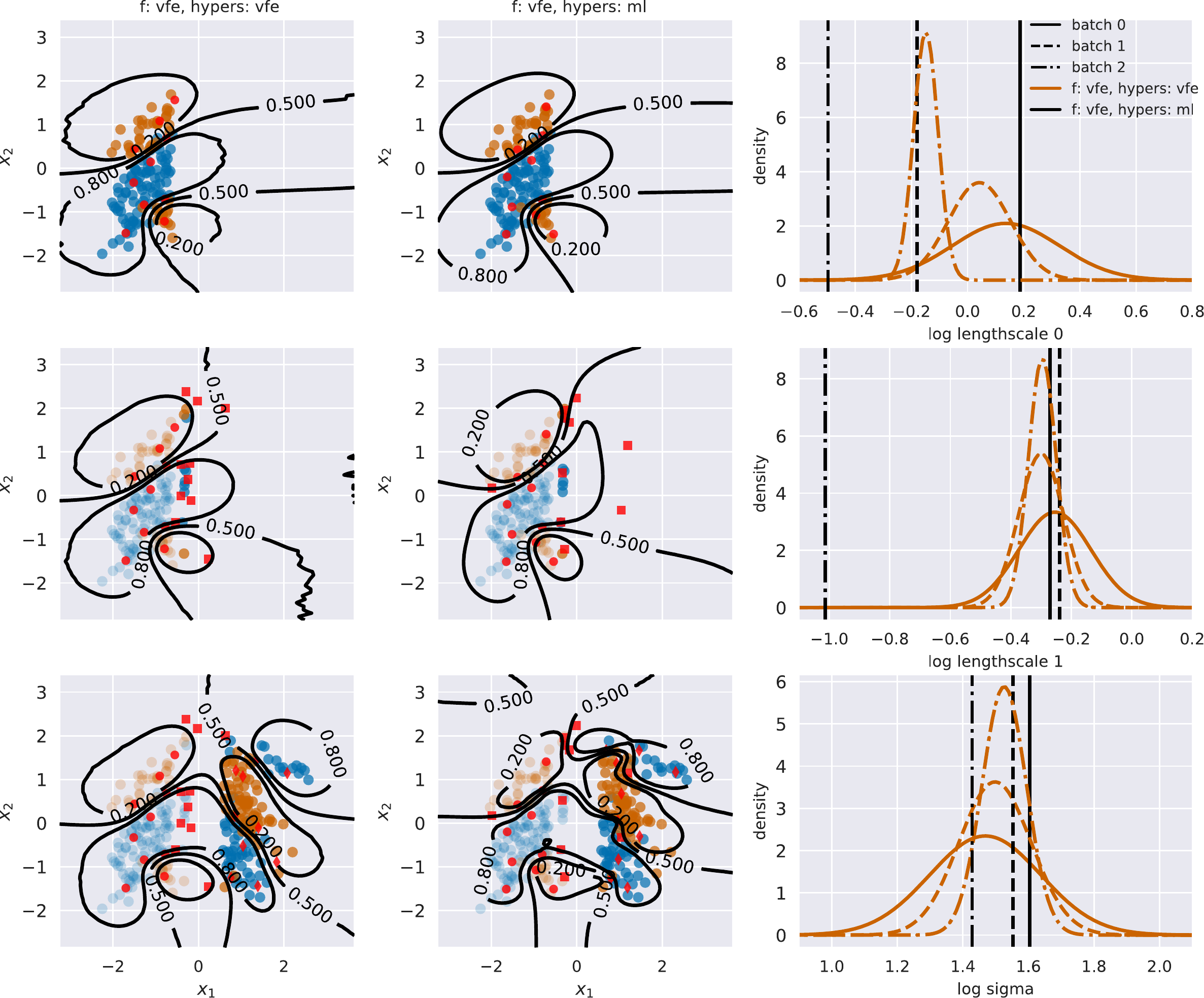}
\caption{Results of the streaming GP experiment on a toy classification data set: a failure case of maximum likelihood learning for the hyperparameters. Two methods were considered: VI for both the latent function and the hyperparameters, and VI for the latent function and approximate maximum likelihood learning for the hyperparameters. We show the predictions made by the methods after seeing each data batch and the corresponding (distributional) hyperparameter estimates. Best viewed in colour. \label{fig:app:gp_toy_streaming_2}}
\end{figure}

In this section, we include several extra results on the toy 2D experiment presented in the main text. We use a Gaussian process prior with a zero mean function and an ARD exponentiated quadratic covariance function and thus there are three kernel hyperparameters to be tuned including the kernel variance and two lengthscale parameters. Several inference methods were considered: (i) MCMC for both the latent function and the hyperparameters, without any sparsification, (ii) variational inference for the latent function and approximate maximum likelihood learning for the hyperparameters, and (iii) variational inference for both the latent function and the hyperparameters. We first consider the batch, static setting, i.e.~inference using the whole data set, and then the streaming setting with three equal batches for the variational methods. \Cref{fig:app:gp_toy_batch_vs_streaming} shows the predictions and the hyperparameter estimates for all methods once all training points are observed. The predictions seem qualitatively similar, though, for the approximate methods, only point estimates or overconfident distributional estimates of the hyperparameters are obtained. The predictions made by the variational methods after sequentially observing the data batches, together with the hyperparameter estimates, are shown in \cref{fig:app:gp_toy_streaming_1}. We also include a failure case when uncertainty over the hyperparameters are not retained and propagated, in \cref{fig:app:gp_toy_streaming_2}. In this case, only ten data points were included in the second batch. One of the lengthscale hyperparameters was severely under-estimated when approximate maximum likelihood learning was used.

\section{Full results of the federated learning experiments}
In this section, we include the full results of the federated learning experiment to train Bayesian neural networks on federated, decentralized data. As a reminder, Bayesian neural networks are an important model in modern machine learning toolkit, fusing the capacity of neural networks with the flexibility of Bayesian inference. The goal of Bayesian learning for neural networks is to obtain the posterior of the network parameters given the training points and to use this for prediction as follows,
\begin{align*}
&\text{posterior:} \;\; & p(\theta|\xvec, \yvec) &= 
\frac{p(\theta) \prod_{n=1}^{N} p(y_n|\theta, x_n)}{p(\yvec|\xvec)} = \frac{p(\theta) \prod_{k=1}^{K} \prod_{n=1}^{N_k} p(y_{n, k}|\theta, x_{n, k})}{p(\yvec|\xvec)},\\
&\text{prediction:} \;\; & p(y^*|\yvec, \xvec, x^*) &= \int p(\theta|\xvec, \yvec) p(y^*|\theta, x^*)  \dd \theta,
\end{align*}
where $\{x_n, y_n\}_{n=1}^{N}$ are the training points, $\theta$ is the network weights and biases. However, getting the exact posterior is analytically intractable and as such approximation methods are needed. In this section, we discuss several approximation strategies for training a Bayesian neural network on the standard MNIST ten-way classification data set. In particular, we focus on a case where data are decentralized on different machines, that is we further assume that $N$ training points are partitioned into $K=10$ disjoint memory shards. Furthermore, two levels of data homogeneity across memory shards are considered: homogeneous [or iid, that is each shard has training points of all classes] and inhomogeneous [or non-iid, i.e.~each shard has training points of only one class]. 

We place a diagonal standard Normal prior over the parameters, $p(\theta) = \norm(\theta; 0, \mathrm{I})$, and initialize the mean of the variational approximations as suggested by \cite{glorot+bengio:17}. For distributed training methods, the data set is partitioned into 10 subsets or shards, and 10 compute nodes (workers) with each able to access one memory shard. The implementation of different inference strategies was done in Tensorflow \citep{abadi+al:16}) and the workload between workers is managed using Ray \citep{moritz+al:17}.

\subsection{Global VI}
\begin{landscape}
\begin{figure}
    \begin{subfigure}[h]{0.5\linewidth}
        \includegraphics[width=\linewidth]{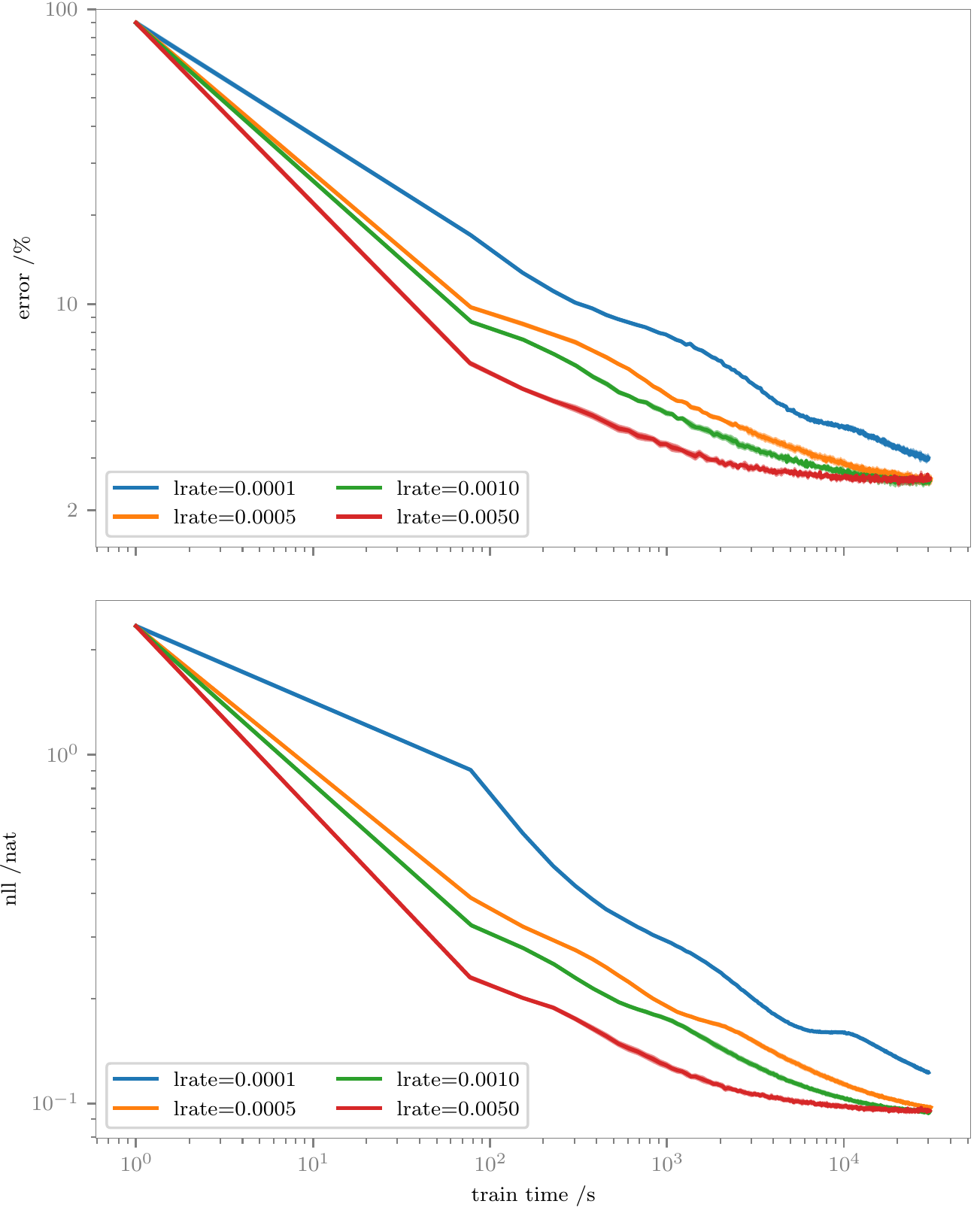}
        \caption{one compute node with one core and iid mini-batches}
    \end{subfigure}
    \hfill
    \begin{subfigure}[h]{0.5\linewidth}
        \includegraphics[width=\linewidth]{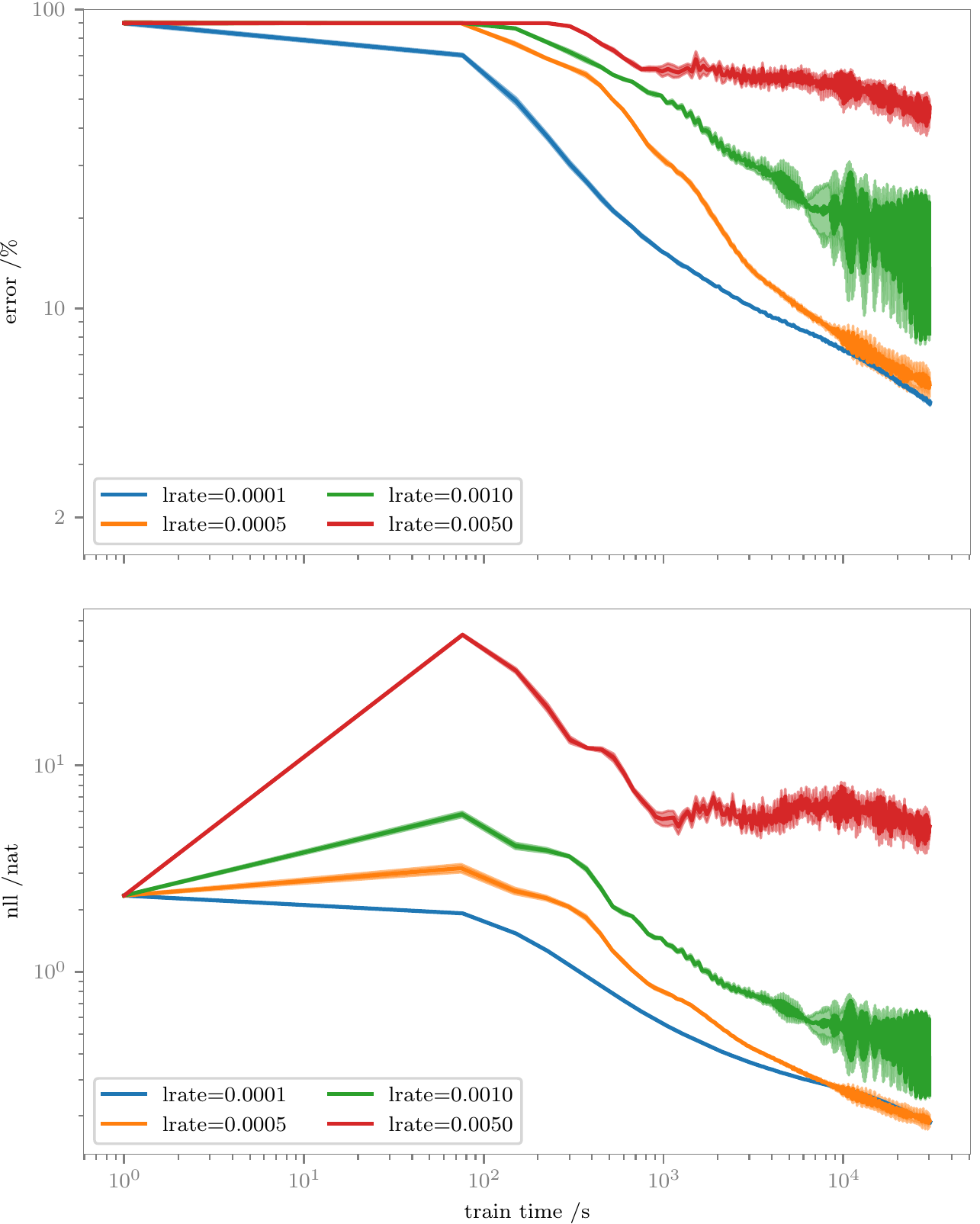}
        \caption{one compute node with one core and non-iid mini-batches}
    \end{subfigure}%
    \caption{The performance of global VI on the test set in the iid [left] and non-iid [right] settings, when the compute node has only one core. Different traces correspond to different learning rate hyperparameters of Adam.}
    \label{fig:app:res_dist_gvi_one}
\end{figure}
\end{landscape}

\begin{landscape}
\begin{figure}
    \begin{subfigure}[h]{0.5\linewidth}
        \includegraphics[width=\linewidth]{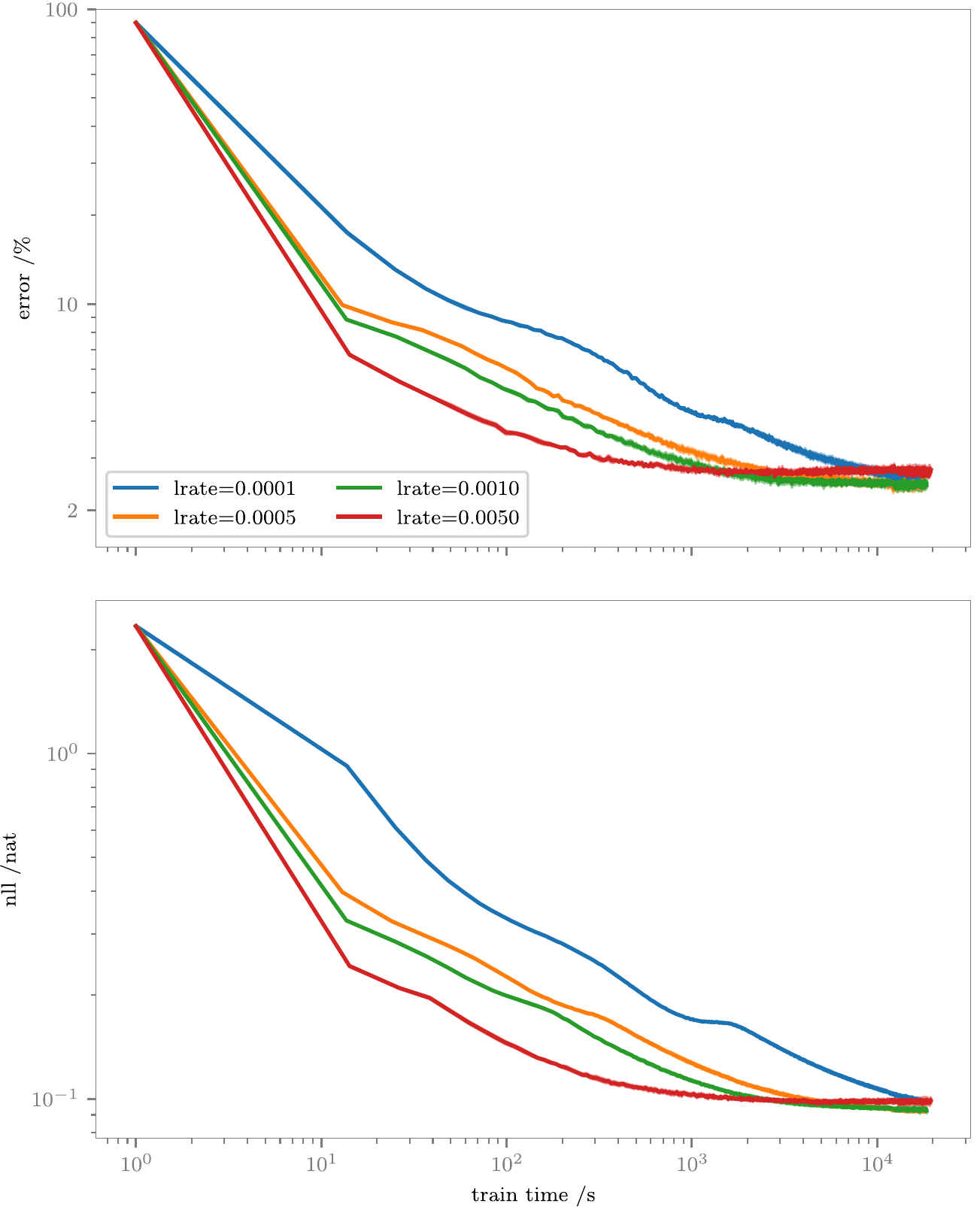}
        \caption{one compute node with ten cores and iid mini-batches}
    \end{subfigure}
    \hfill
    \begin{subfigure}[h]{0.5\linewidth}
        \includegraphics[width=\linewidth]{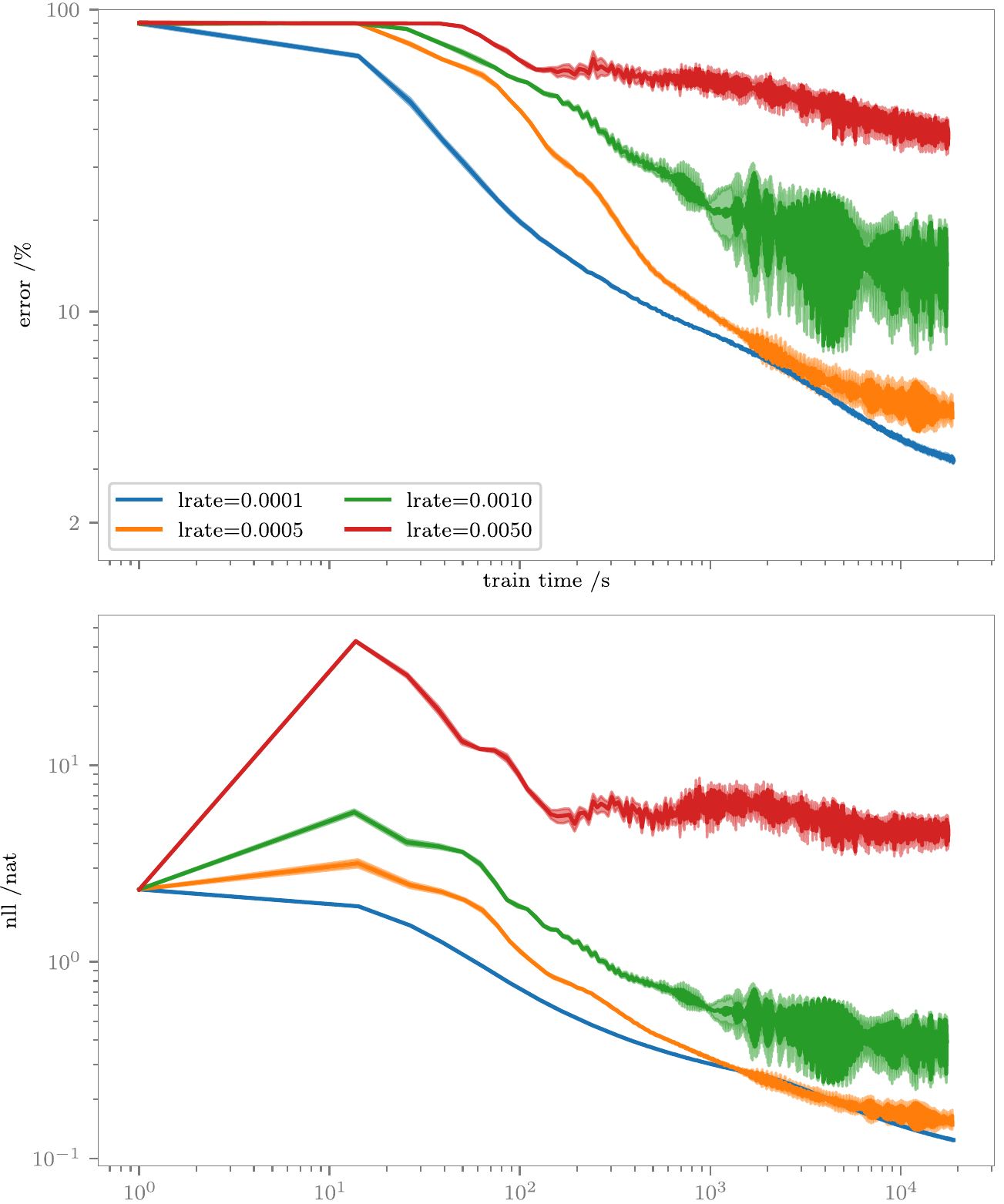}
        \caption{one compute node with ten cores and non-iid mini-batches}
    \end{subfigure}%
    \caption{The performance of global VI on the test set in the iid and non-iid settings, when the compute node has ten cores. Different traces correspond to different learning rate hyperparameters of Adam.}
    \label{fig:app:res_dist_gvi_ten}
\end{figure}
\end{landscape}

We first considered global variational inference, as described in \cref{sec:federated}, for getting an approximate posterior over the parameters. The variational lower bound (\cref{eqn:bnn_gvi_vfe}) is optimized using Adam \citep{kingma+ba:2014}. We considered one compute node (with either one core or ten cores) that can access the entire data set, and simulates the data distribution by sequentially showing mini-batches that can potentially have all ten classes (iid) or that have data of only one class (non-iid). The full performance on the test set during training for different learning rate hyperparameters of the Adam optimizer are shown in \cref{fig:app:res_dist_gvi_one,fig:app:res_dist_gvi_ten}. Notice that in the iid setting, larger learning rates tend to yield faster convergence but can give a slightly poorer predictive performance on the test set at the end of training (see \cref{fig:app:res_dist_gvi_ten} with a learning rate of 0.005). The non-iid is arguably more difficult and the performance can oscillate if the learning rate is too large.

\subsection{Bayesian committee machine}
\begin{landscape}
\begin{figure}
    \begin{subfigure}[h]{0.5\linewidth}
        \includegraphics[width=\linewidth]{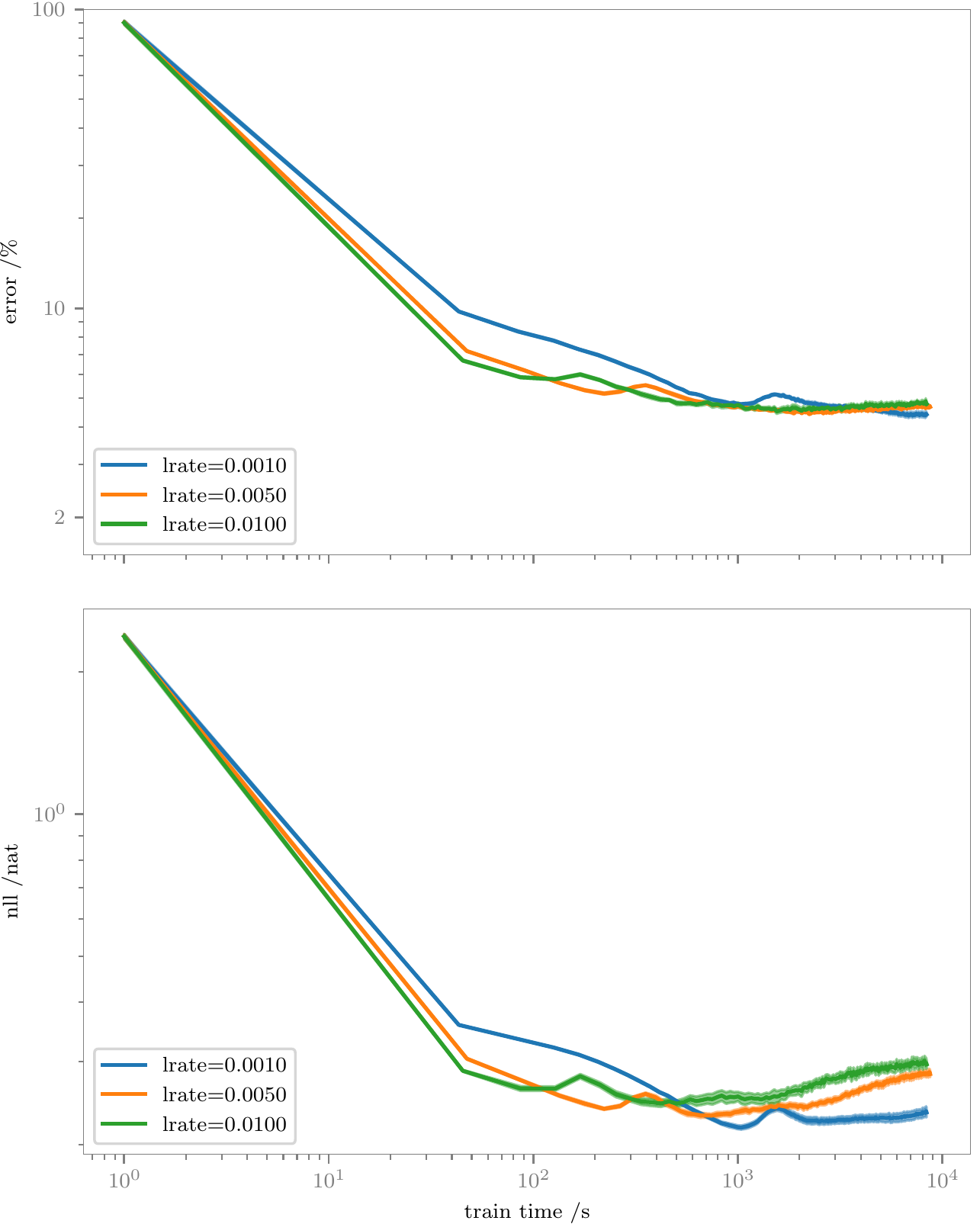}
        \caption{BCM with the same $\norm(0,1)$ prior across 10 workers and iid data}
    \end{subfigure}
    \hfill
    \begin{subfigure}[h]{0.5\linewidth}
        \includegraphics[width=\linewidth]{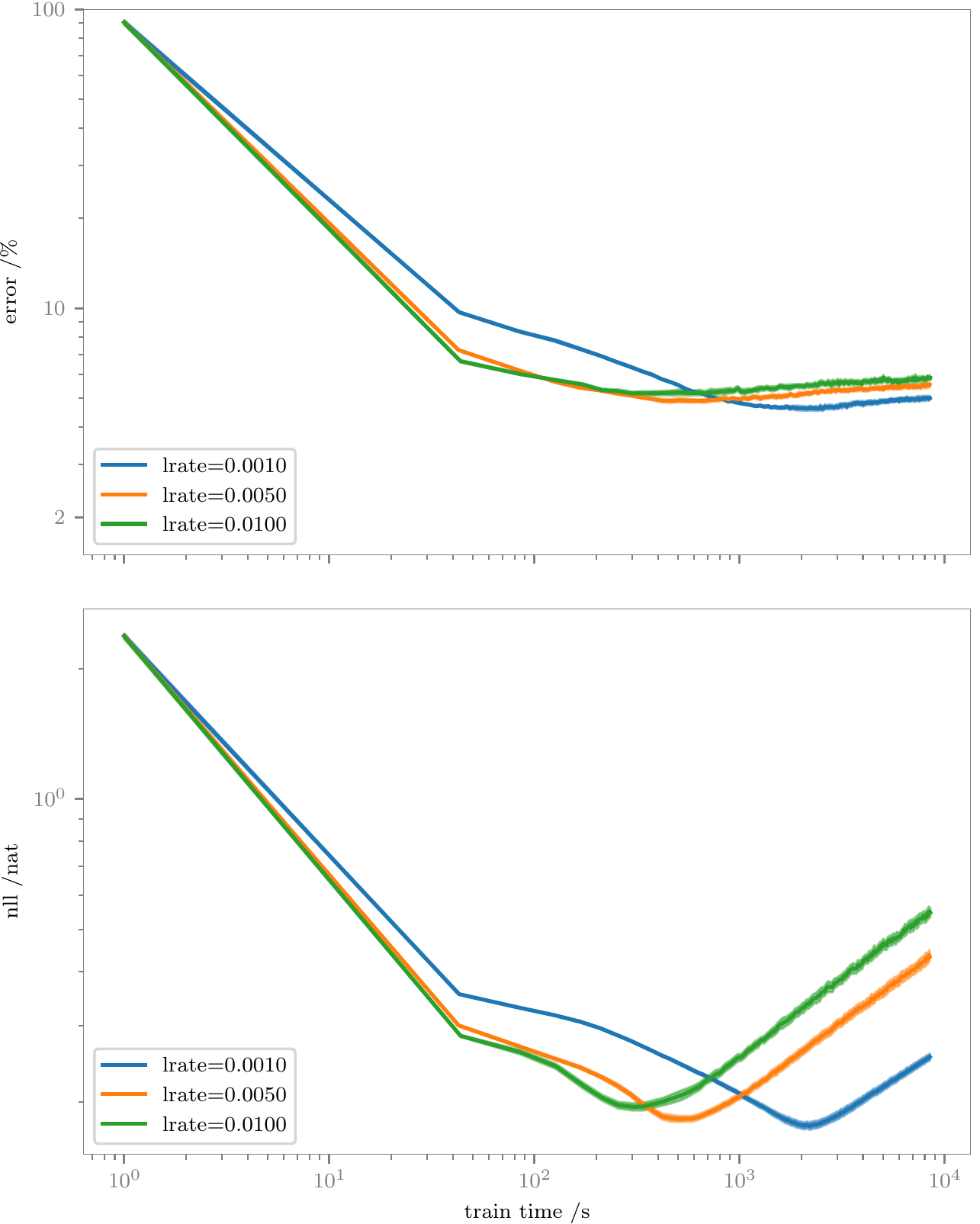}
        \caption{BCM with the prior $\norm(0,1)$ being split equally across 10 workers and iid data}
    \end{subfigure}
    \caption{Performance of BCM with two prior sharing strategies on the iid setting, for various learning rates. Best viewed in colour.}
    \label{fig:app:res_dist_bcm_iid}
\end{figure}
\end{landscape}

\begin{landscape}
\begin{figure}
    \begin{subfigure}[h]{0.5\linewidth}
        \includegraphics[width=\linewidth]{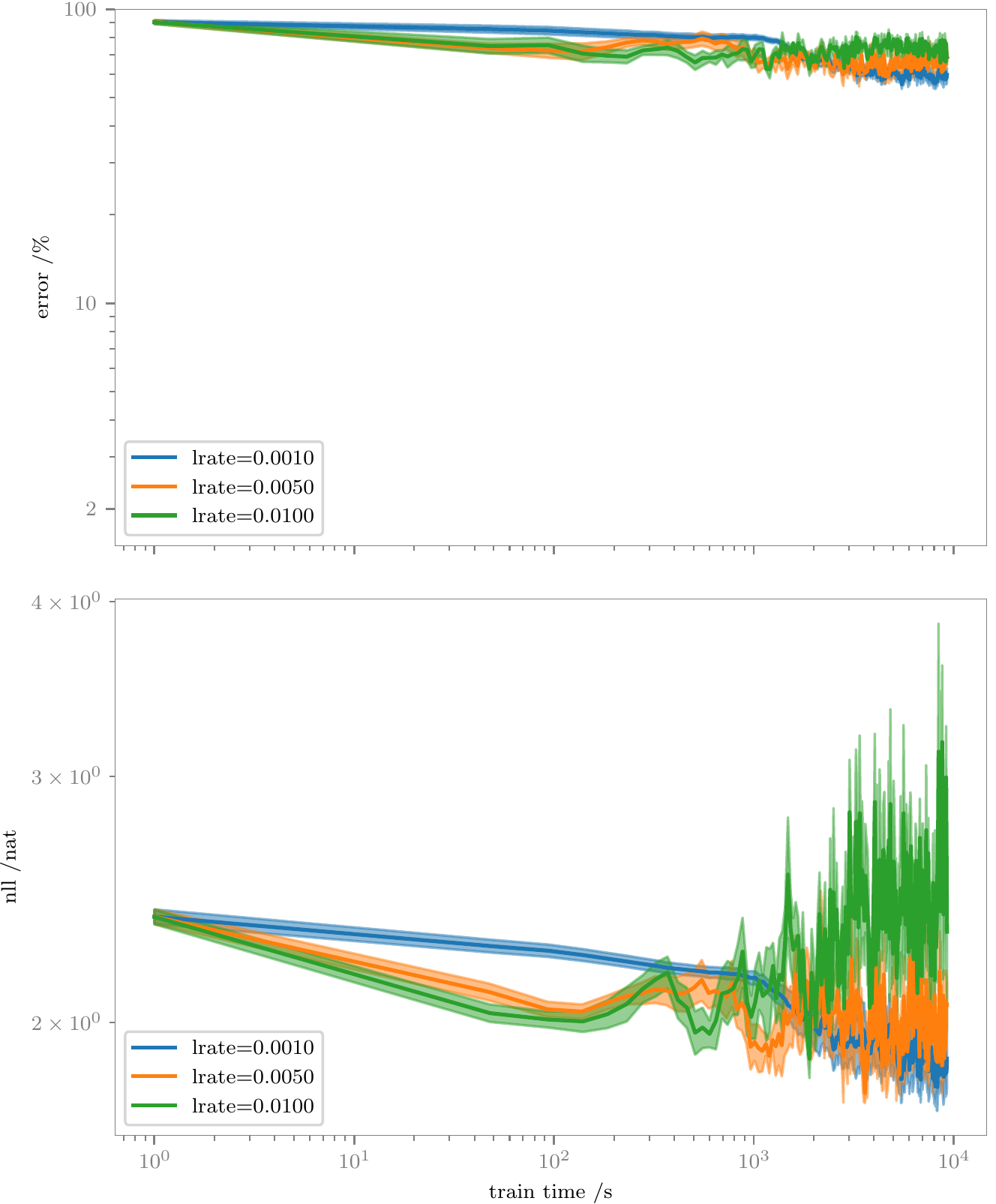}
        \caption{BCM with the same $\norm(0,1)$ prior across workers and non-iid data}
    \end{subfigure}
    \hfill
    \begin{subfigure}[h]{0.5\linewidth}
        \includegraphics[width=\linewidth]{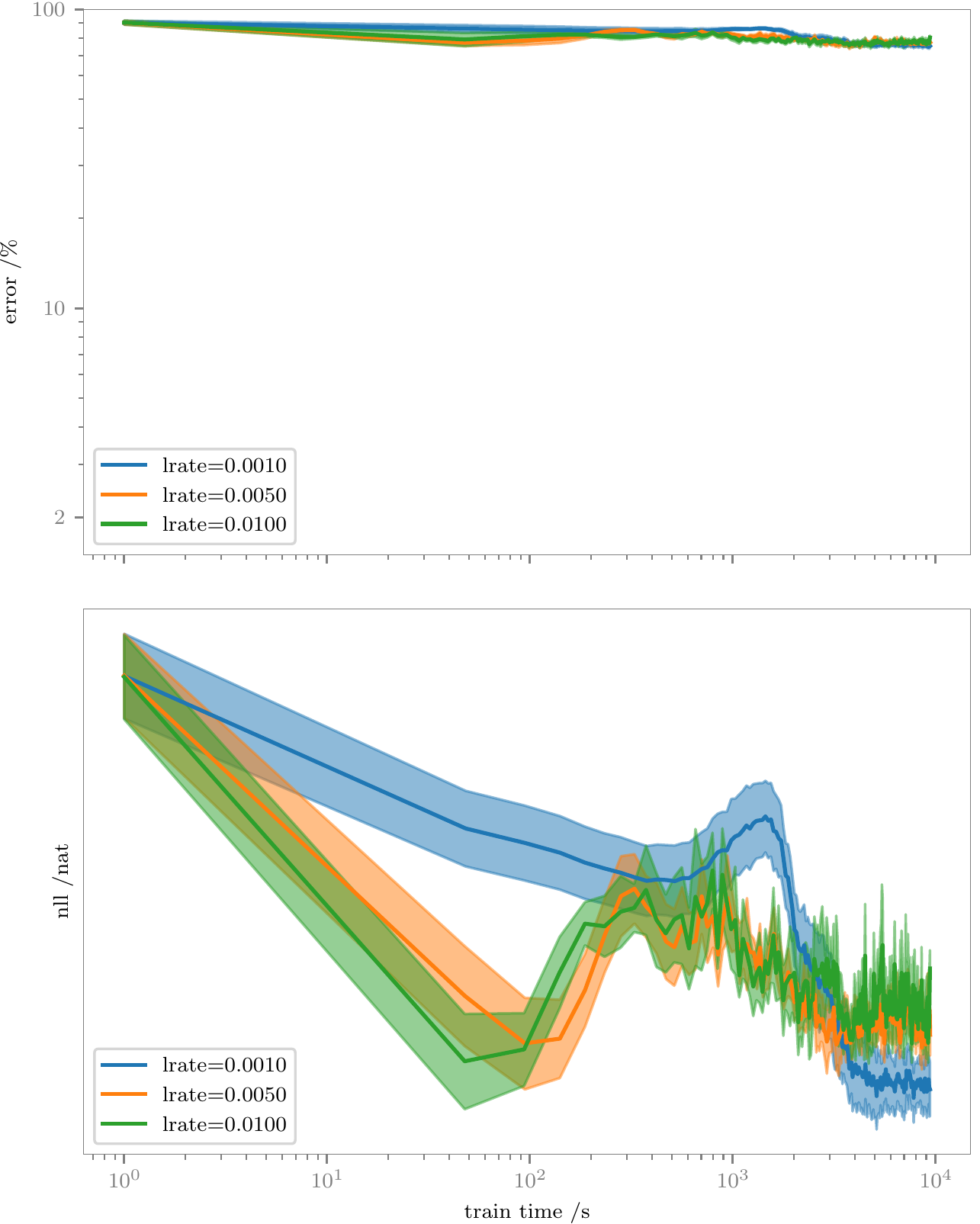}
        \caption{BCM with the prior $\norm(0,1)$ being split equally across workers and non-iid data}
    \end{subfigure}
    \caption{Performance of BCM with two prior sharing strategies on the non-iid setting, for various learning rates. Best viewed in colour.}
    \label{fig:app:res_dist_bcm_noniid}
\end{figure}
\end{landscape}

We next considered an embarrassingly parallel scheme based on the Bayesian committee machine \cite{tresp:00}. In particular, two prior sharing strategies as described in \cref{sec:exp_federated}, BCM - same prior and BCM - split prior, were considered. Each worker has access to one memory shard and performs global variational inference independently. While the workers were running, we occasionally polled the approximate posteriors from all workers, merged them using BCM and computed the test performance using the merged posterior. We report the test performance during training for different prior sharing schemes in both iid and non-iid settings in \cref{fig:app:res_dist_bcm_iid,fig:app:res_dist_bcm_noniid}, respectively. Note that, we also varied the learning rate of Adam for each worker. Both prior sharing strategies in combination with BCM performs surprisingly well in the iid setting. However, they fell short in the non-iid setting as the Gaussian sub-posteriors can potentially have different supports and, if this the case, multiplying them will not give a good global approximation.

\subsection{Sequential, one-pass PVI}
\begin{landscape}
\begin{figure}
    \begin{subfigure}[h]{0.5\linewidth}
        \includegraphics[width=\linewidth]{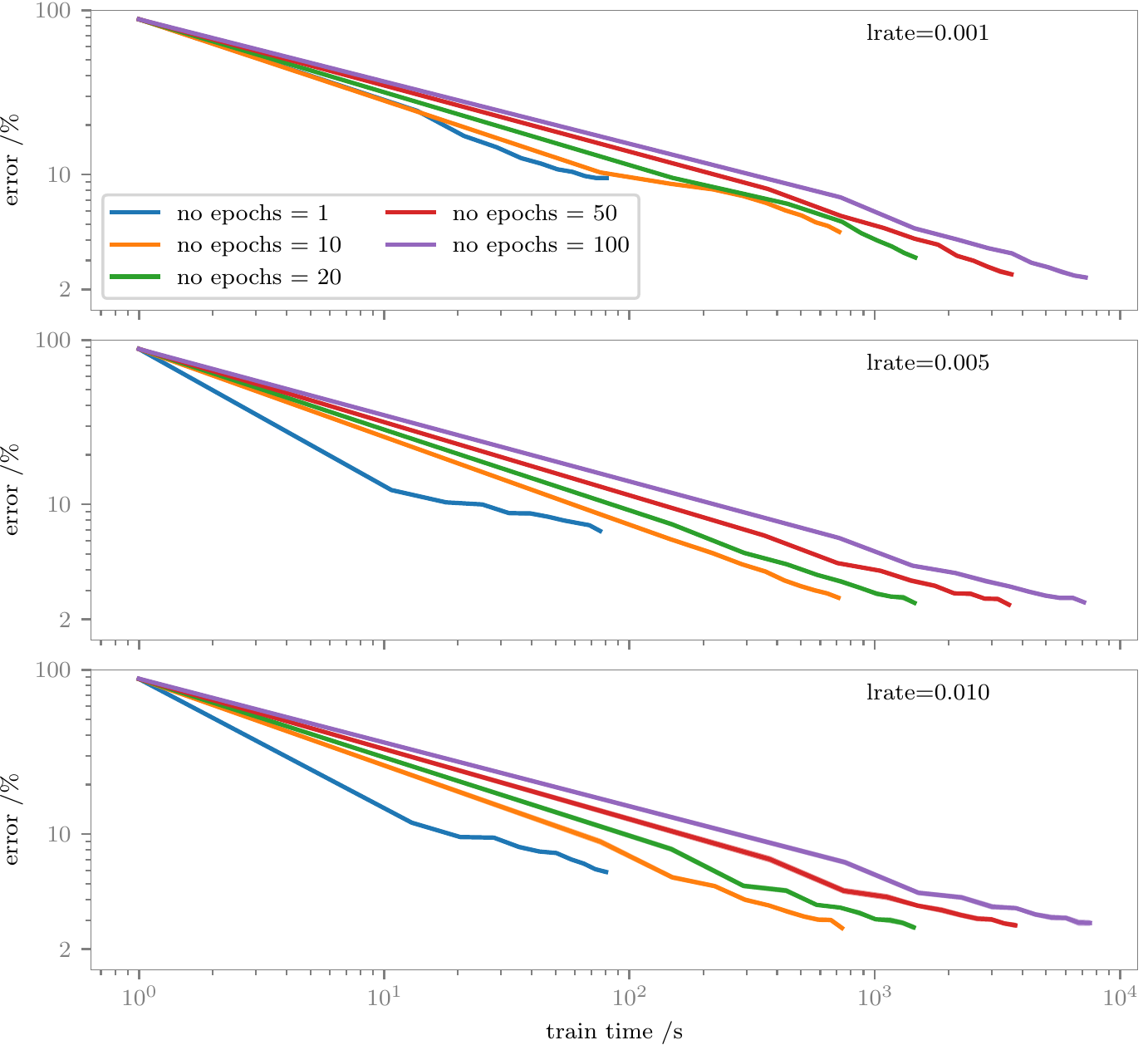}
        \caption{Error}
    \end{subfigure}
    \hfill
    \begin{subfigure}[h]{0.5\linewidth}
        \includegraphics[width=\linewidth]{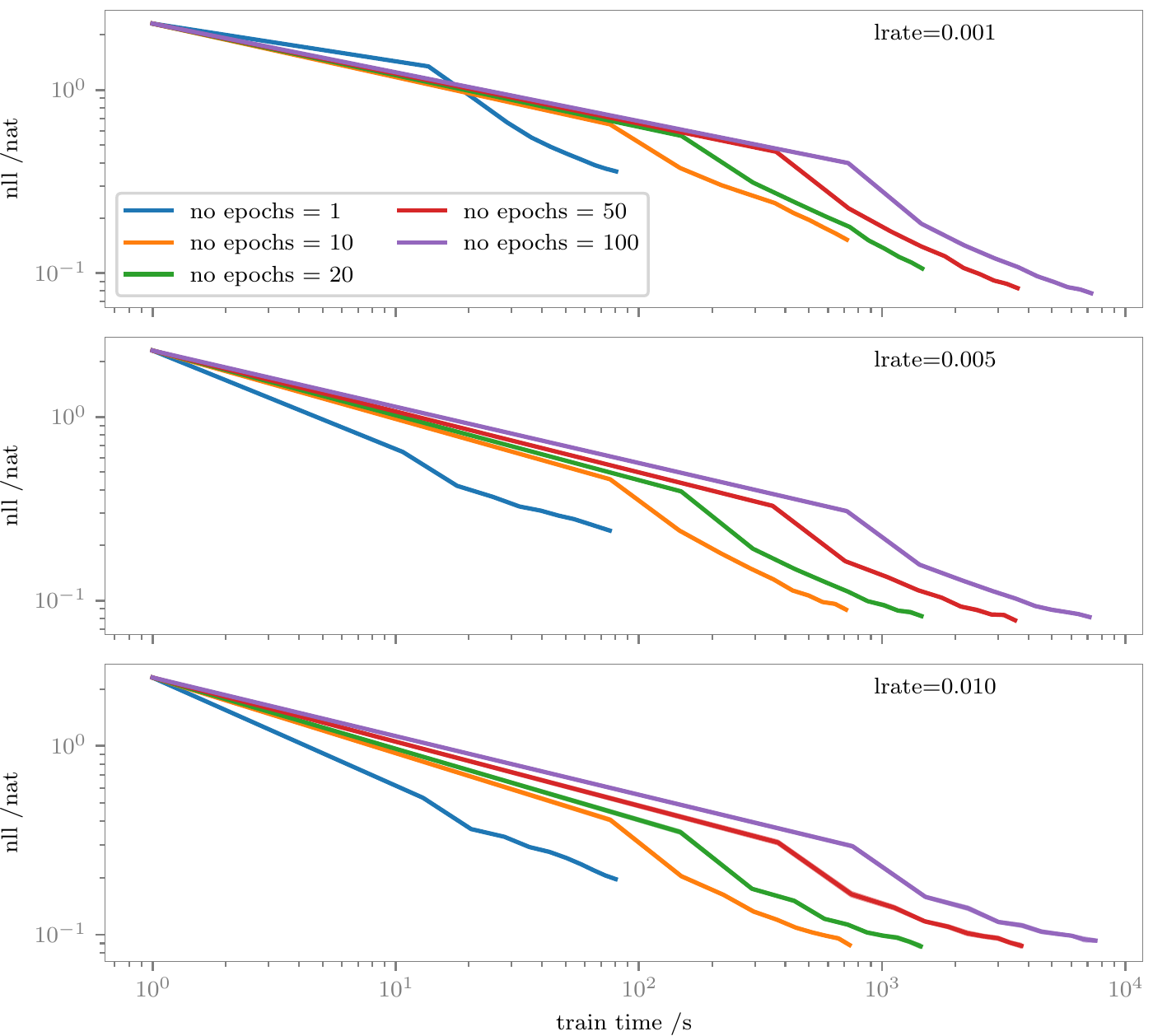}
        \caption{NLL}
    \end{subfigure}
    \caption{Performance of sequential PVI with only one pass through all memory shards when the data are iid. The number of epochs for each worker and the learning rate hyperparameter of Adam were varied. Best viewed in colour.}
    \label{fig:app:res_dist_pvi_one_pass_iid}
\end{figure}
\end{landscape}

\begin{landscape}
\begin{figure}
    \begin{subfigure}[h]{0.5\linewidth}
        \includegraphics[width=\linewidth]{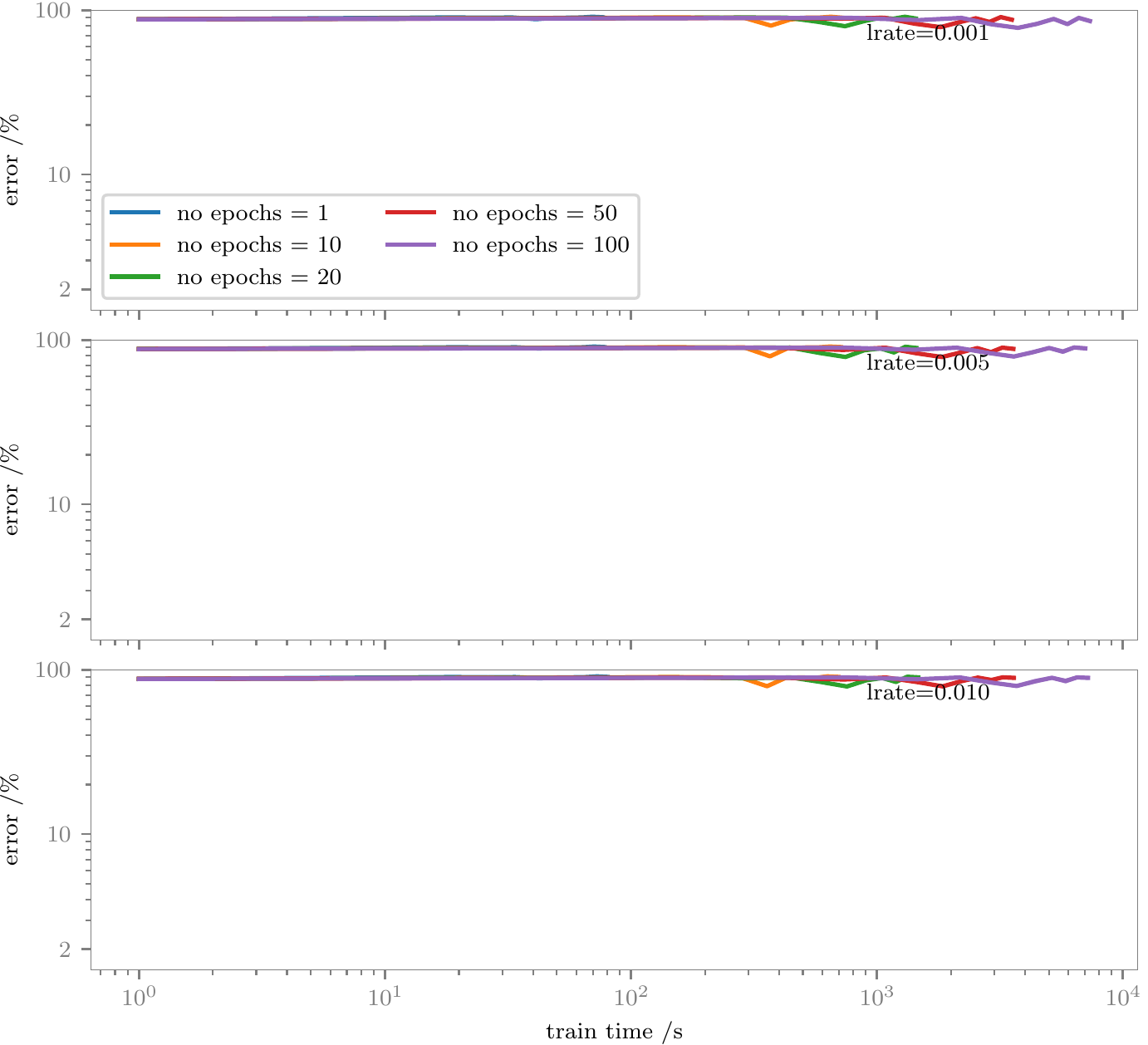}
        \caption{Error}
    \end{subfigure}
    \hfill
    \begin{subfigure}[h]{0.5\linewidth}
        \includegraphics[width=\linewidth]{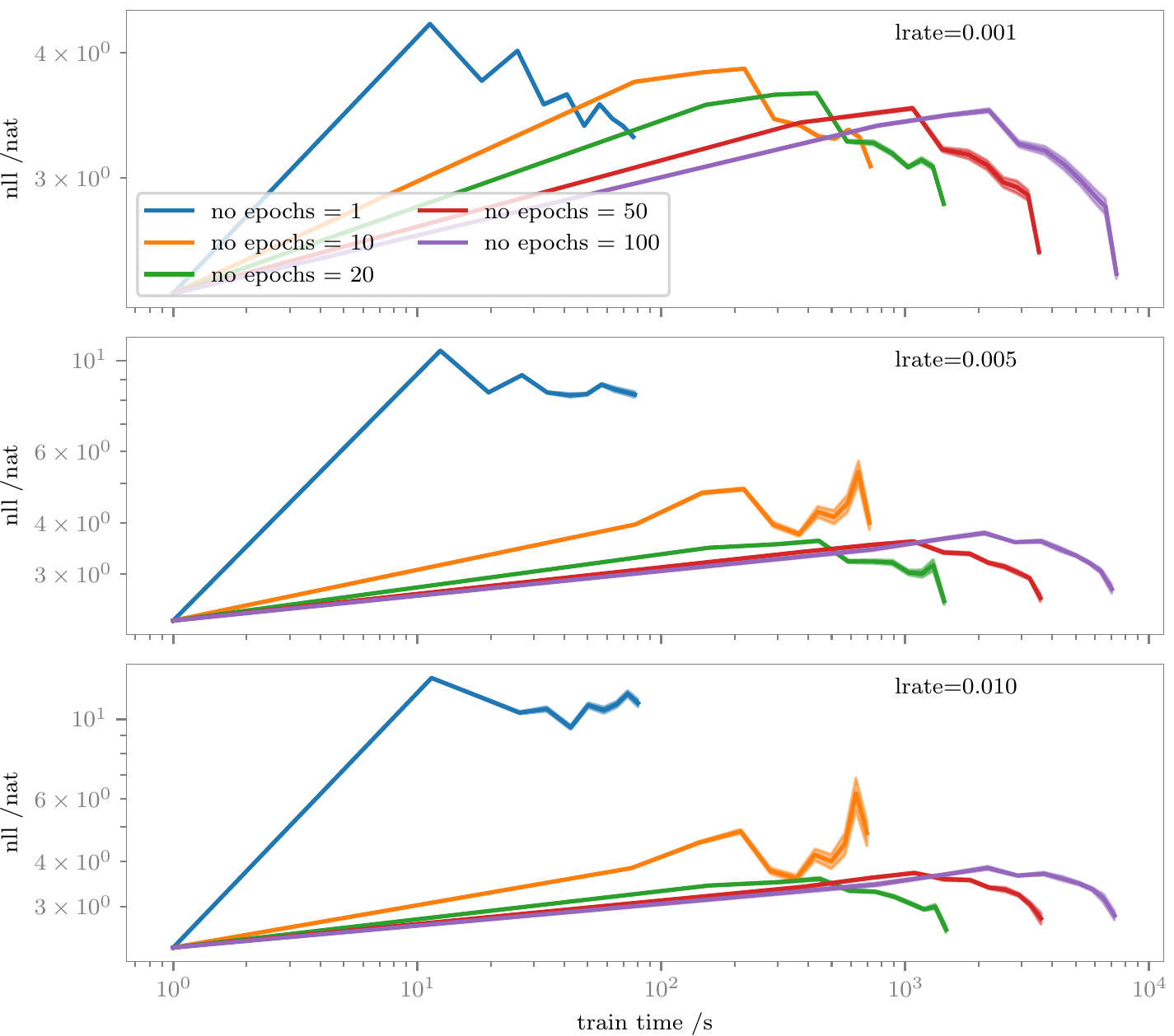}
        \caption{NLL}
    \end{subfigure}
    \caption{Performance of sequential PVI with only one pass through all memory shards when the data are non-iid. The number of epochs for each worker and the learning rate hyperparameter of Adam were varied. Best viewed in colour.}
    \label{fig:app:res_dist_pvi_one_pass_noniid}
\end{figure}
\end{landscape}

We next considered a sequential training scheme using PVI. Each worker, in turn, performs global variational inference with the prior being the posterior of the last trained worker. Learning using this schedule is identical to the Variational Continual Learning approach of \cite{nguyen+al:2018}. We varied the learning rates of Adam (used to optimize the variational lower bound for each worker) and the number of epochs of data used for each worker. The test performance was recorded after each worker finished its training, and the full results for the iid and non-iid settings are shown in \cref{fig:app:res_dist_pvi_one_pass_iid,fig:app:res_dist_pvi_one_pass_noniid}, respectively. This schedule performs well in the iid setting, but struggles when the data across workers are non-iid.

\subsection{PVI with synchronous updates}
\begin{landscape}
\begin{figure}
    \begin{subfigure}[h]{0.5\linewidth}
        \includegraphics[width=\linewidth]{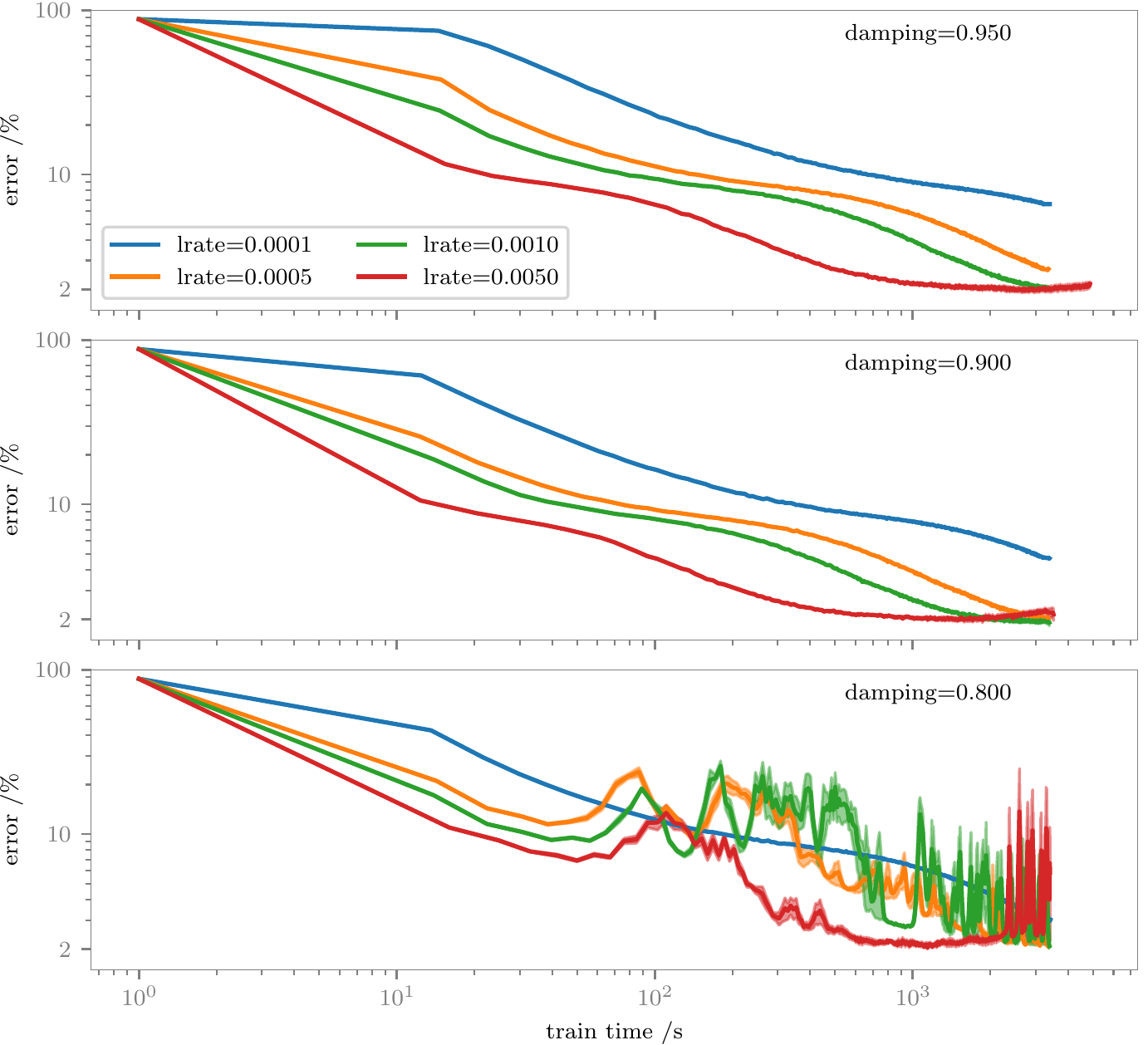}
        \caption{Error}
    \end{subfigure}
    \hfill
    \begin{subfigure}[h]{0.5\linewidth}
        \includegraphics[width=\linewidth]{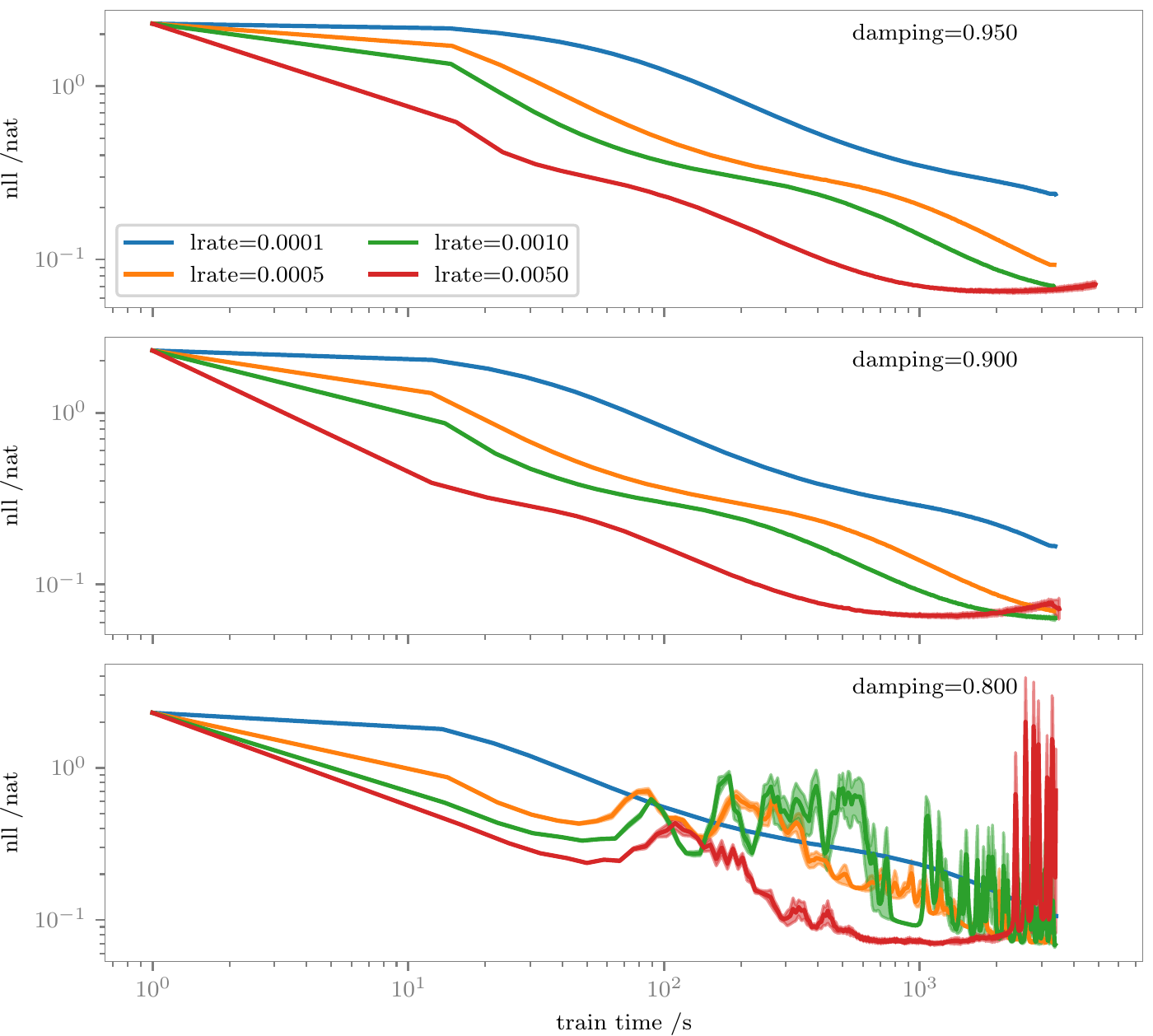}
        \caption{NLL}
    \end{subfigure}
    \caption{Performance of PVI with synchronous updates when the data are iid. In this experiment, each worker communicates with the central server after one epoch. The learning rate hyperparameter of Adam and the damping factor were varied. Best viewed in colour.}
    \label{fig:app:res_dist_pvi_sync_iid}
\end{figure}
\end{landscape}

\begin{landscape}
\begin{figure}
    \begin{subfigure}[h]{0.5\linewidth}
        \includegraphics[width=\linewidth]{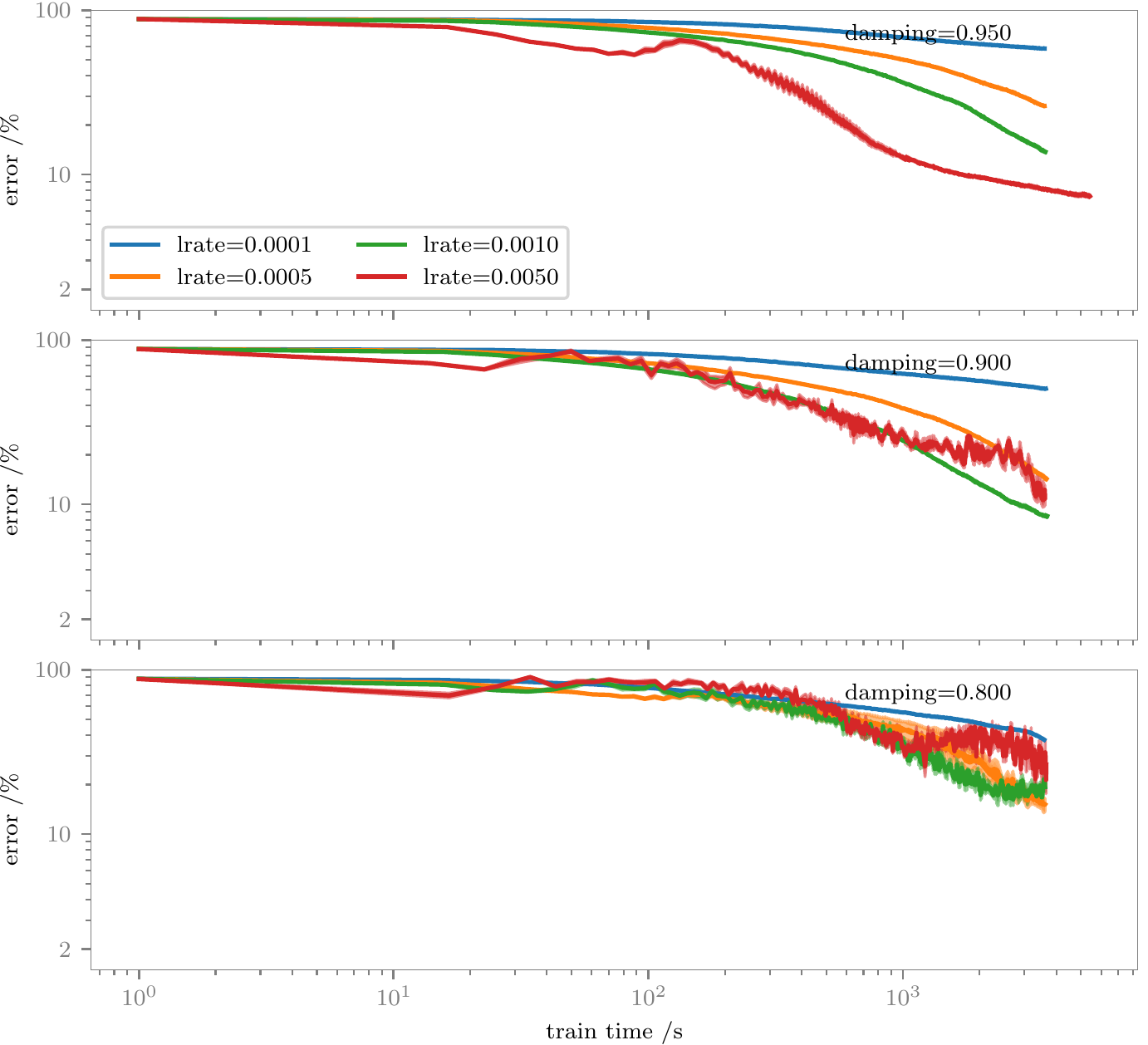}
        \caption{Error}
    \end{subfigure}
    \hfill
    \begin{subfigure}[h]{0.5\linewidth}
        \includegraphics[width=\linewidth]{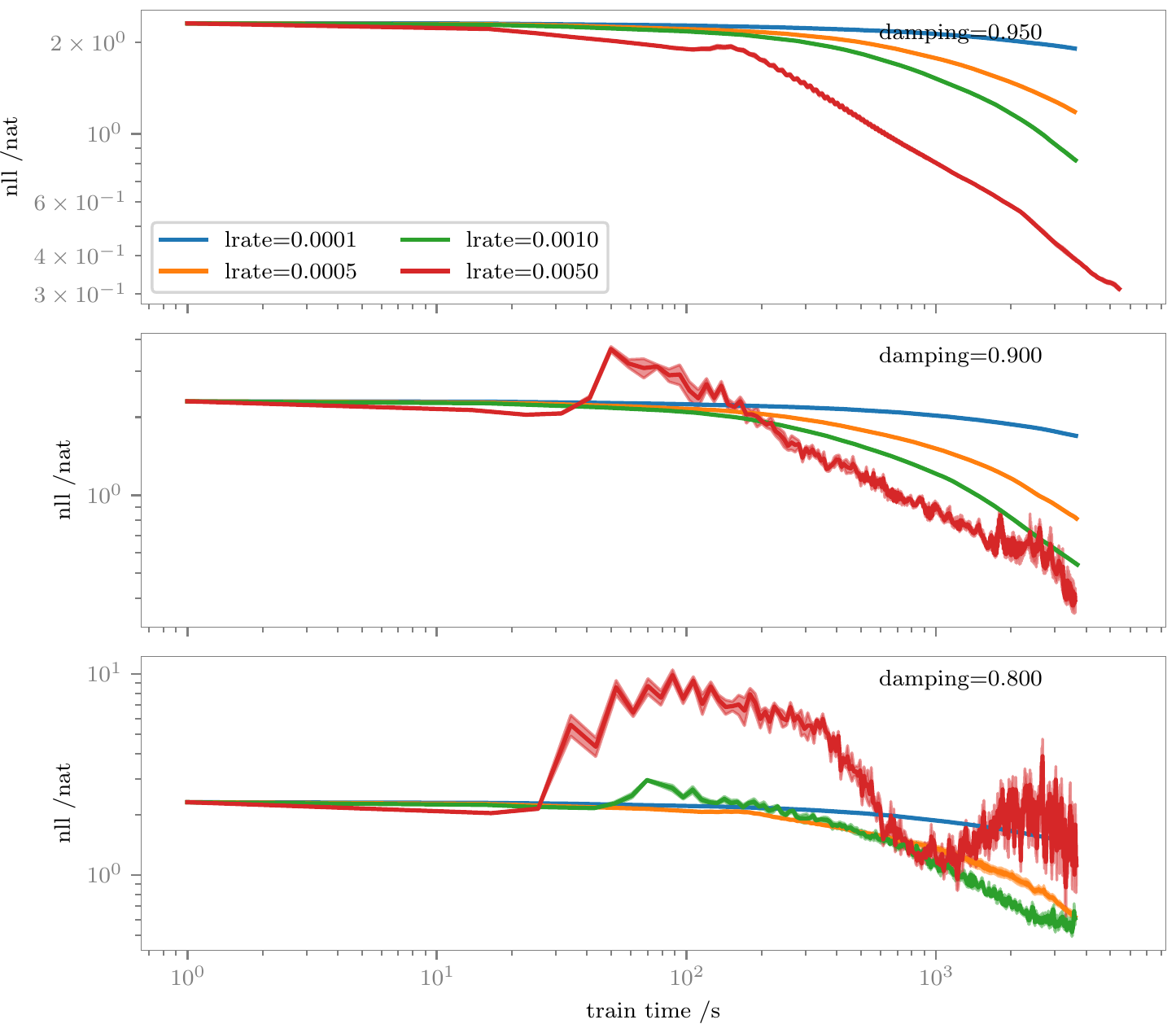}
        \caption{NLL}
    \end{subfigure}
    \caption{Performance of PVI with synchronous updates when the data are non-iid. In this experiment, each worker communicates with the central server after one epoch. The learning rate hyperparameter of Adam and the damping factor were varied. Best viewed in colour.}
    \label{fig:app:res_dist_pvi_sync_noniid}
\end{figure}
\end{landscape}

We considered PVI with synchronous updates, i.e.~the central parameter waits for all workers to finish before gathering the approximate likelihoods together and then sending out the new approximate posterior. As typically done in (parallel) Power-EP, we also considered damped updates, i.e.~the new approximate likelihood is a linear combination of the old approximate likelihood and the factor provided by a worker. We explored different damping factors (higher means slower updates), and different learning rates for the optimization at the local worker. The full results on the test set during training are shown in \cref{fig:app:res_dist_pvi_sync_iid,fig:app:res_dist_pvi_sync_noniid} for the iid and non-iid settings, respectively. 

\subsection{PVI with asynchronous updates}
\begin{landscape}
\begin{figure}
    \begin{subfigure}[h]{0.5\linewidth}
        \includegraphics[width=\linewidth]{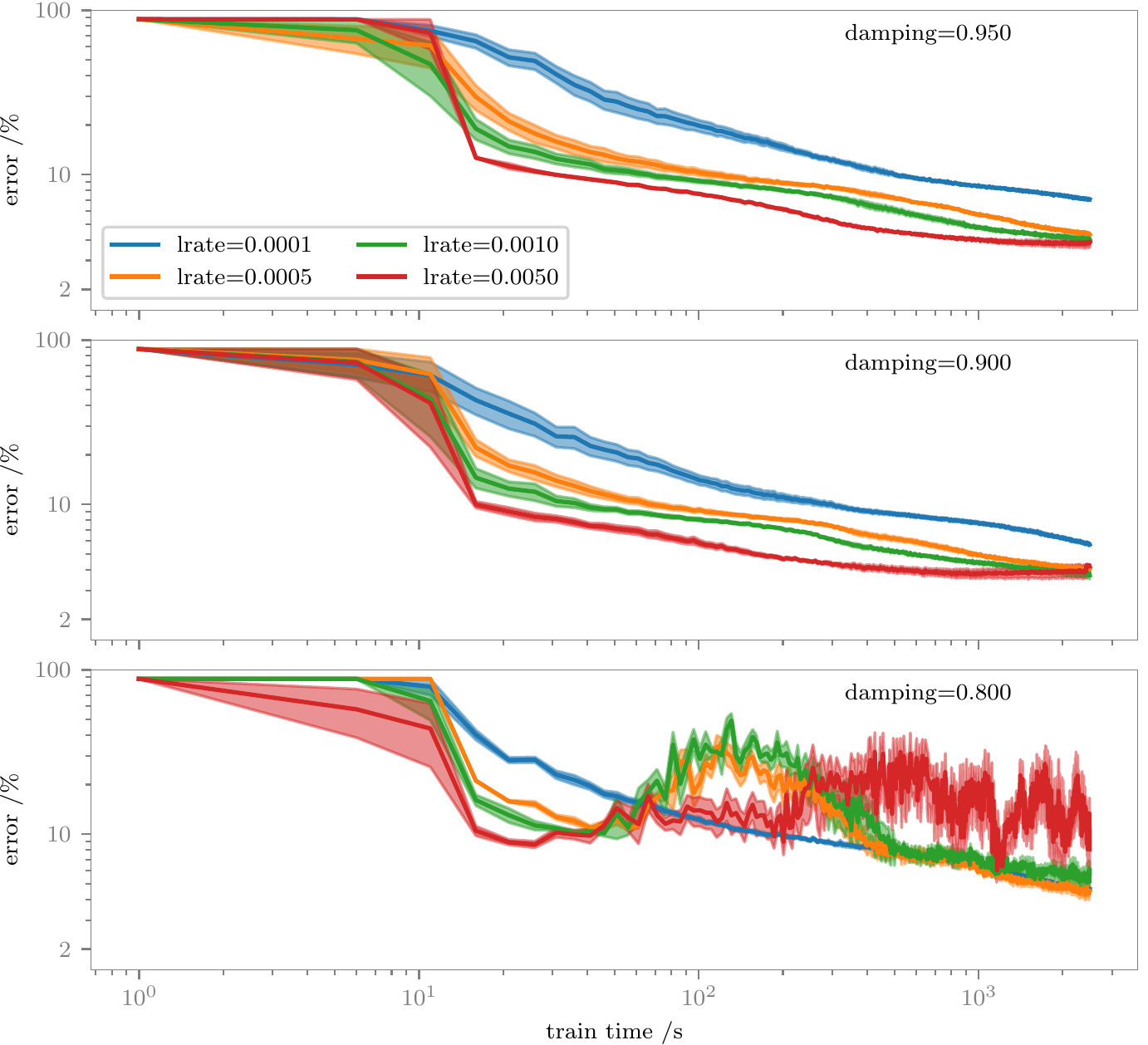}
        \caption{Error}
    \end{subfigure}
    \hfill
    \begin{subfigure}[h]{0.5\linewidth}
        \includegraphics[width=\linewidth]{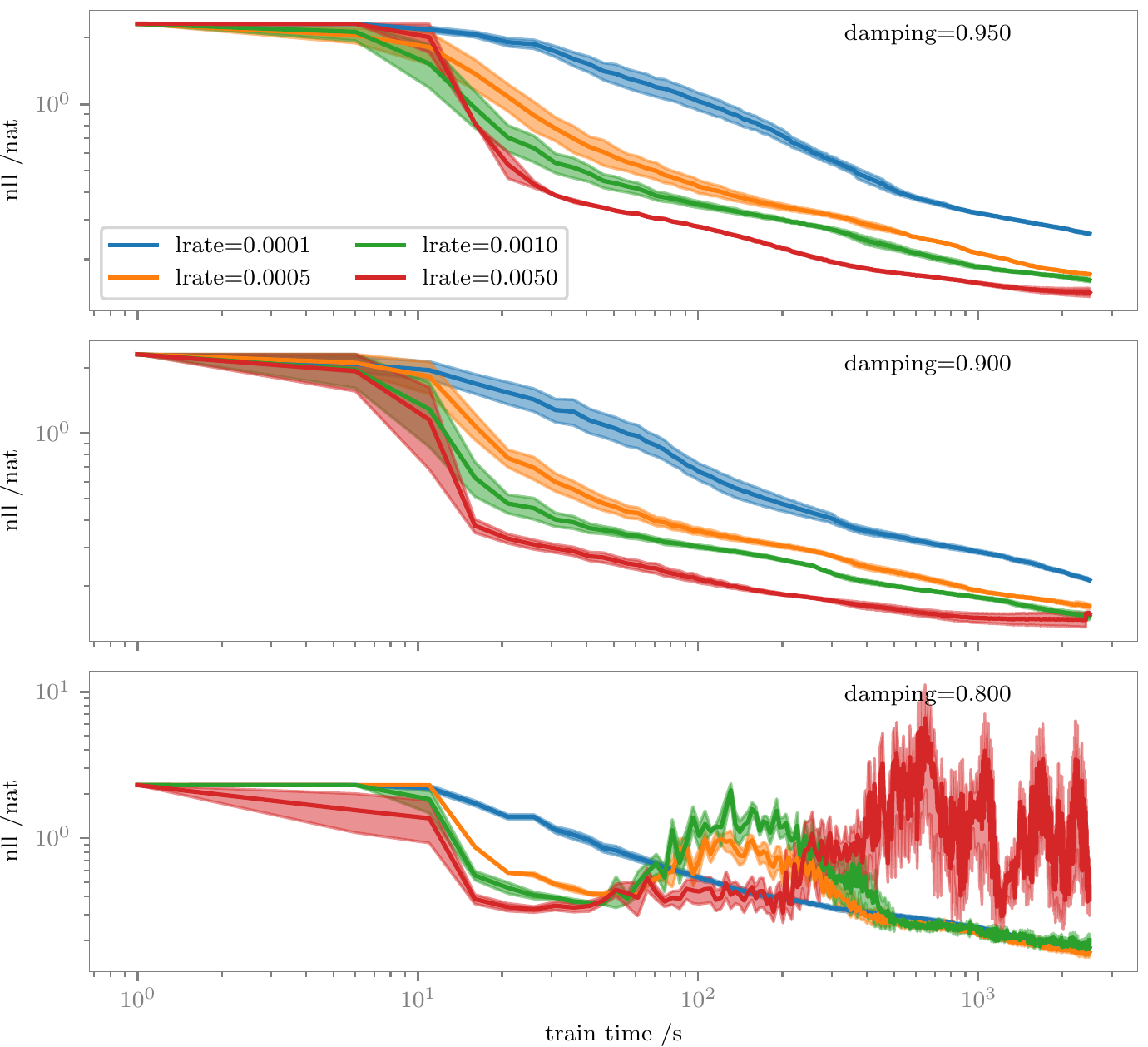}
        \caption{NLL}
    \end{subfigure}
    \caption{Performance of PVI with asynchronous, lock-free updates when the data are iid. In this experiment, each worker communicates with the central server after one epoch. The learning rate hyperparameter of Adam and the damping factor were varied. Best viewed in colour.}
    \label{fig:app:res_dist_pvi_async_iid}
\end{figure}
\end{landscape}

\begin{landscape}
\begin{figure}
    \begin{subfigure}[h]{0.5\linewidth}
        \includegraphics[width=\linewidth]{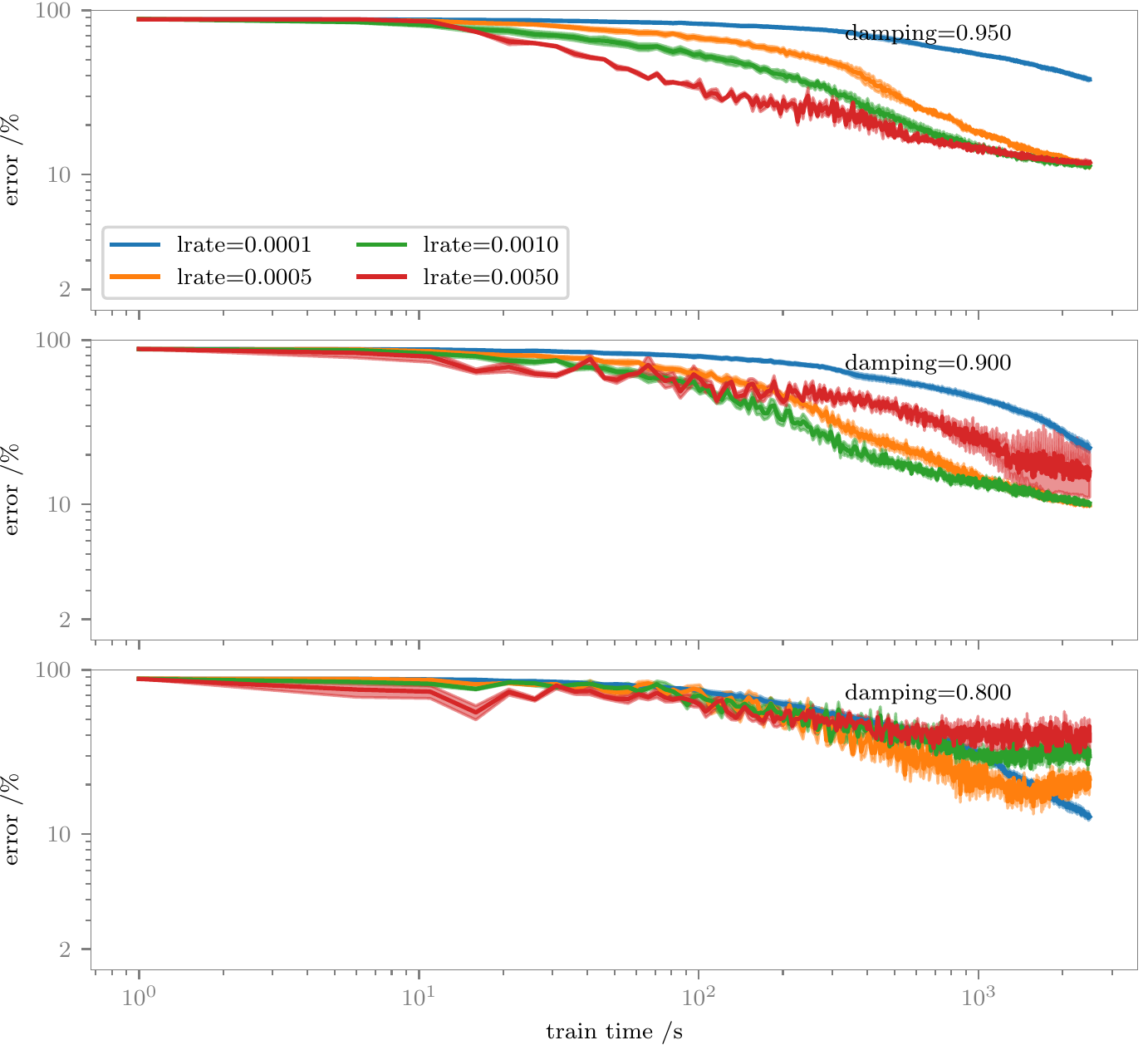}
        \caption{Error}
    \end{subfigure}
    \hfill
    \begin{subfigure}[h]{0.5\linewidth}
        \includegraphics[width=\linewidth]{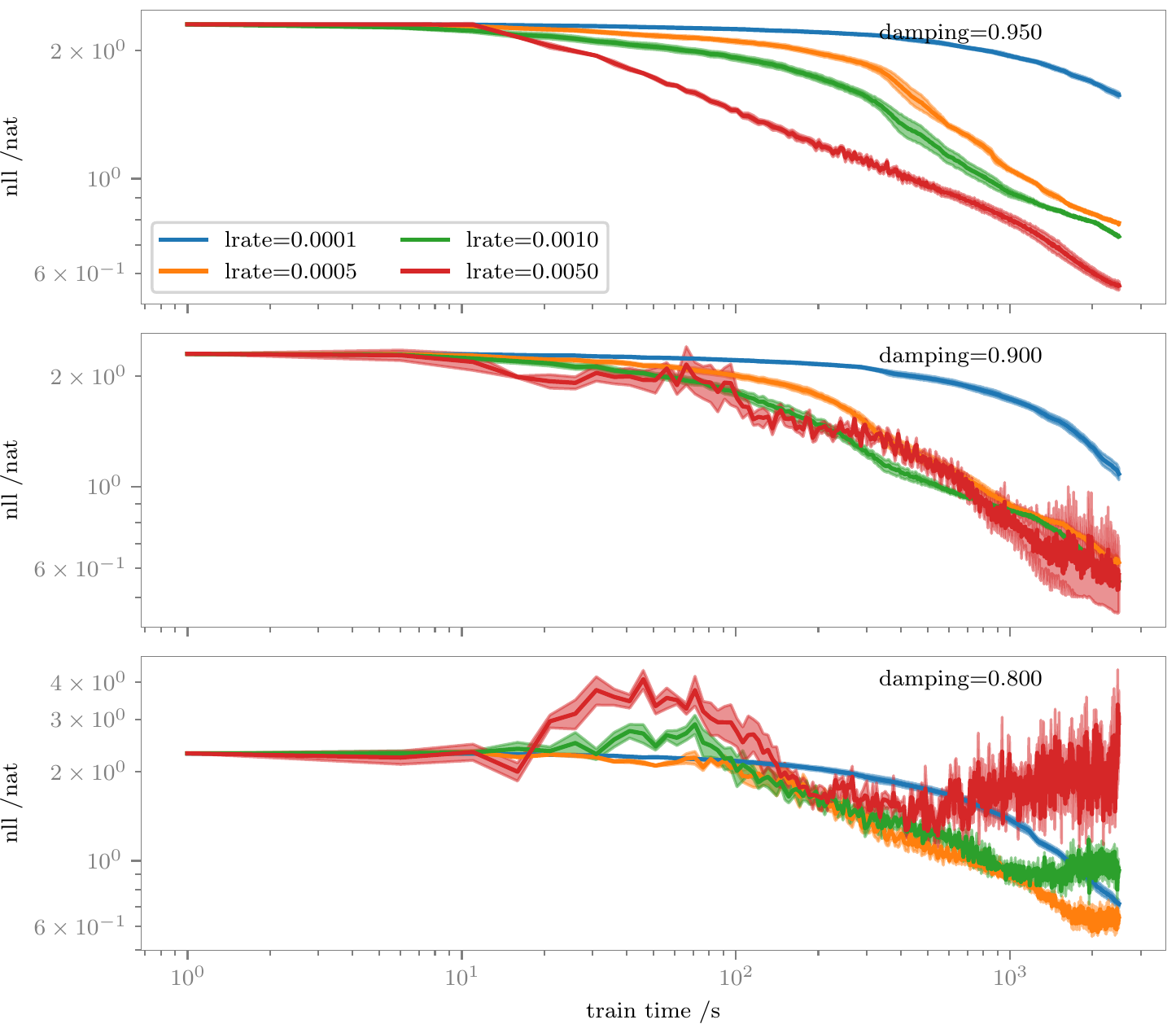}
        \caption{NLL}
    \end{subfigure}
    \caption{Performance of PVI with asynchronous, lock-free updates when the data are non-iid. In this experiment, each worker communicates with the central server after one epoch. The learning rate hyperparameter of Adam and the damping factor were varied. Best viewed in colour.}
    \label{fig:app:res_dist_pvi_async_noniid}
\end{figure}
\end{landscape}

Finally, we allowed the parameter server to update the approximate posterior as soon as a worker has finished its computation. This schedule can potentially cause stale updates. However, this is more suitable to cases when the communication between the workers and the parameter server is unreliable, or when the distribution of data across workers is skewed, i.e.~one worker might have a lot more data points than others and consequently might require a longer time to finish a computation round. As in the synchronous case, we varied the damping factor and learning rate. The full results are shown in  \cref{fig:app:res_dist_pvi_async_iid,fig:app:res_dist_pvi_async_noniid} for the iid and non-iid settings, respectively.

\begin{figure}[h]
    \includegraphics[width=\linewidth]{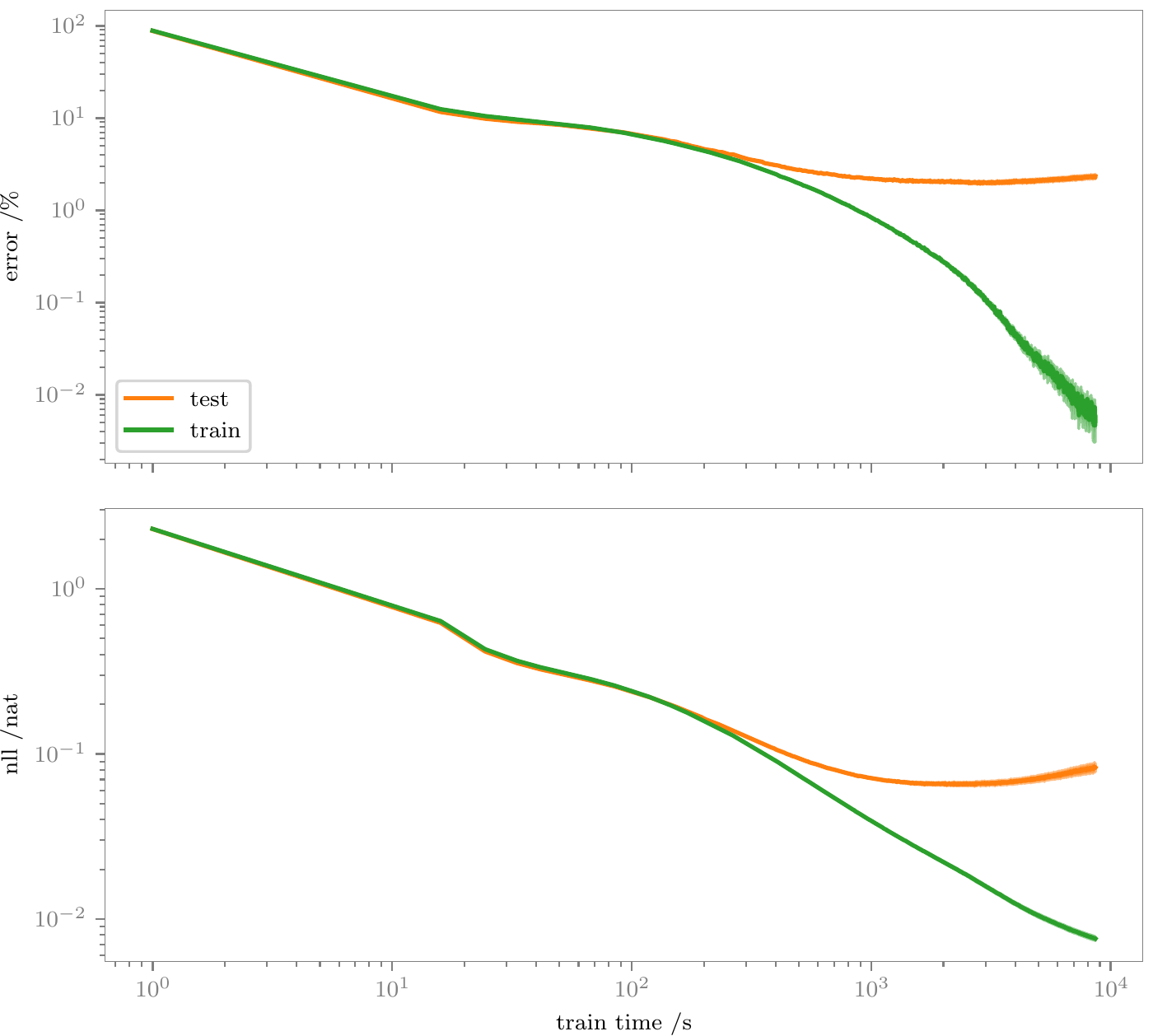}
    \caption{For certain hyperparameter settings, PVI with synchronous updates worryingly exhibits over-fitting.}
    \label{fig:app:res_dist_pvi_sync_overfit}
\end{figure}

\subsection{Stochastic Natural Gradient Variational Inference for Bayesian Neural Networks\label{sec:app:bnn_opt}}
\begin{figure}[!htb]
	\centering
	\includegraphics[width=\linewidth]{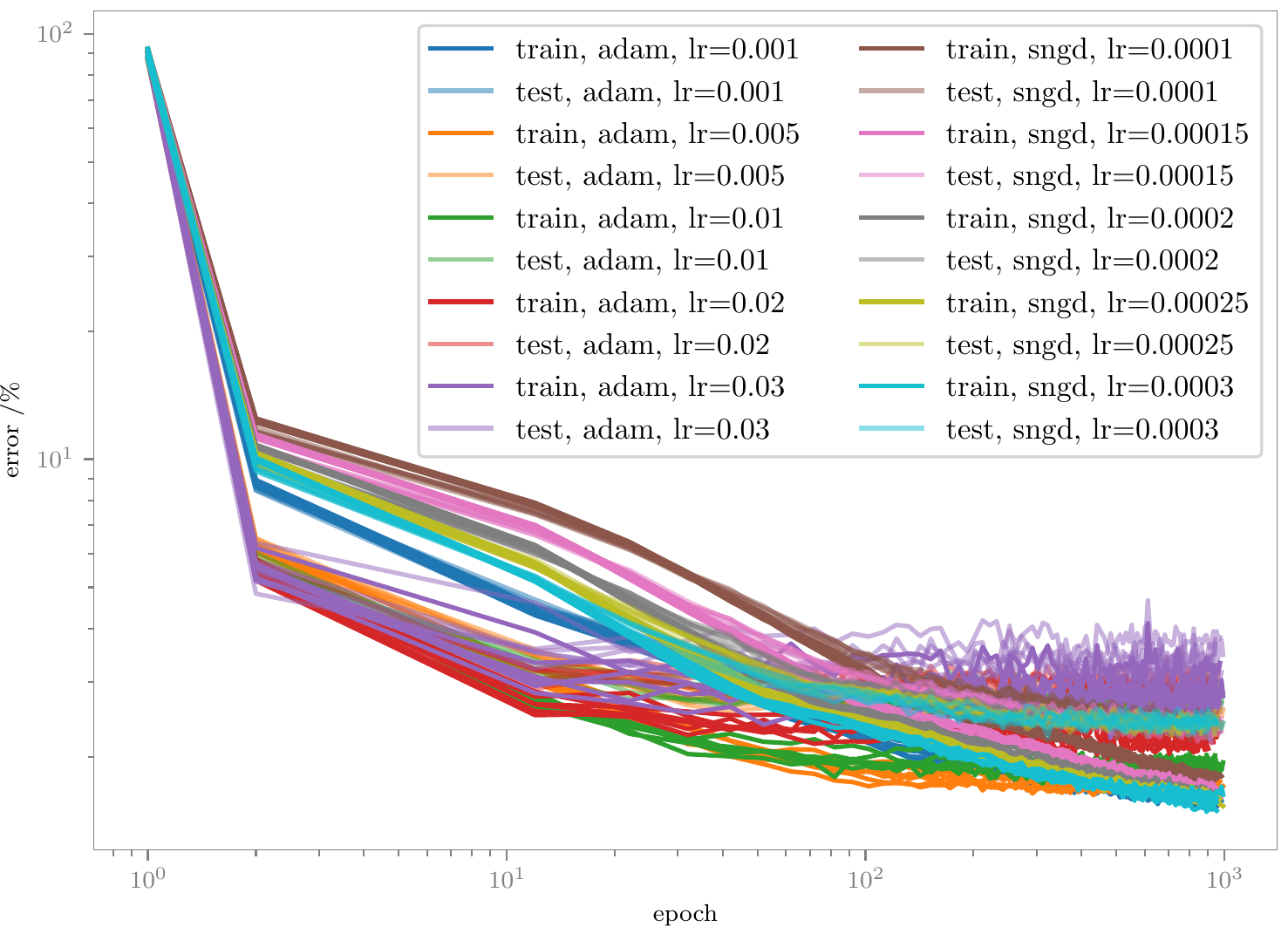}
	\caption{Classification error rates on the train and test sets during training using Adam and Stochastic Natural Gradient (SNGD) methods on the MNIST classification task with a Bayesian neural network with one hidden layer of 200 rectified linear units. The final performance of all settings are shown in \cref{fig:res_bnn_mnist_200_last}. For both Adam and SNGD, the performance highly depends on the learning rate, but the best learning rates for both methods give similar train and test results and yield similar convergence. Note that while Adam adaptively changes the learning rate based on the gradient statistics, SNGD employs a fixed step size. See text for more details. Best viewed in colour.\label{fig:res_bnn_mnist_200_error}}
\end{figure}

\begin{figure}[!htb]
	\centering
	\includegraphics[width=\linewidth]{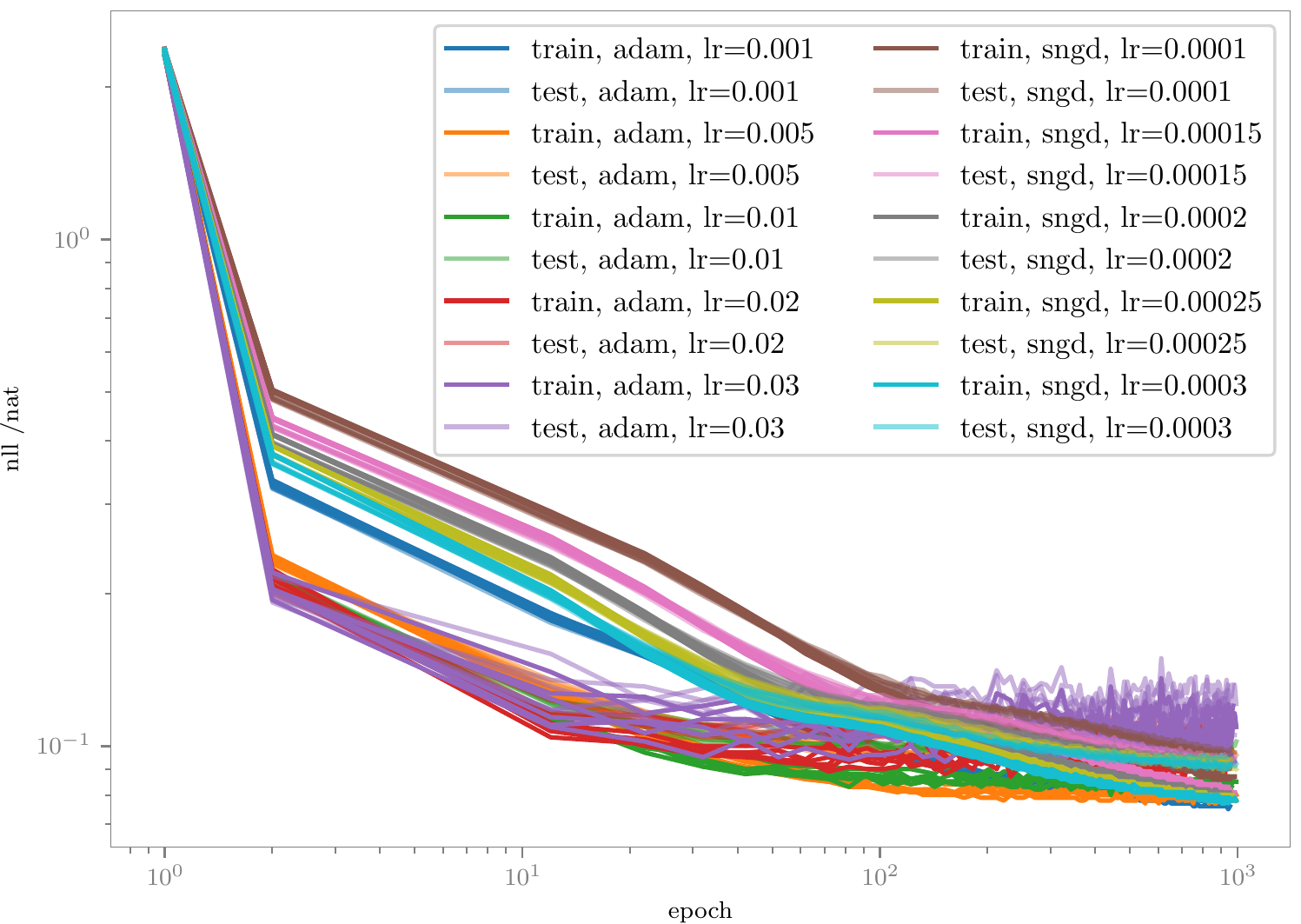}
	\caption{Negative log-likelihoods on the train and test sets during training using Adam and Stochastic Natural Gradient (SNGD) methods on the MNIST classification task with a Bayesian neural network with one hidden layer of 200 rectified linear units. The final performance of all settings are shown in \cref{fig:res_bnn_mnist_200_last}. For both Adam and SNGD, the performance highly depends on the learning rate, but the best learning rates for both methods give similar train and test results and yield similar convergence. Note that while Adam adaptively changes the learning rate based on the gradient statistics, SNGD employs a fixed step size. See text for more details. Best viewed in colour.\label{fig:res_bnn_mnist_200_ll}}
\end{figure}

\begin{landscape}
\begin{figure}
    \centering
	\includegraphics[width=\linewidth]{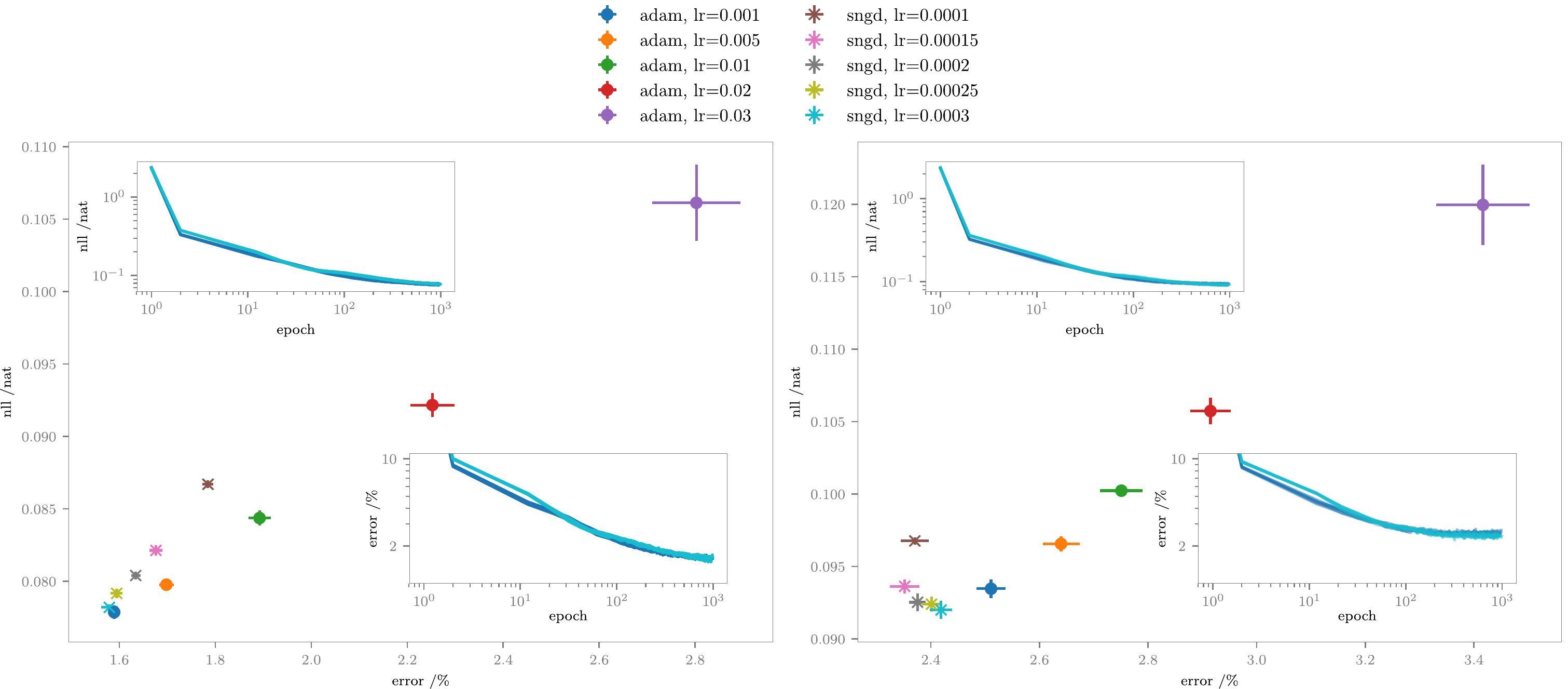}
	\caption{Performance on the train set [left] and test set [right] after 1000 epochs using Adam and Stochastic Natural Gradient (SNGD) methods on the MNIST classification task with a Bayesian neural network with one hidden layer of 200 rectified linear units, and the typical performance traces as training progress [inset plots]. This figure summarizes the full results in \cref{fig:res_bnn_mnist_200_error,fig:res_bnn_mnist_200_ll}. The performance is measured using the classification error [error] and the negative log-likelihood [nll], and for both measures, lower is better and, as such, closer to the bottom left is better. For both Adam and SNGD, the performance highly depends on the learning rate, but the best learning rates for both methods give similar train and test results and yield similar convergence. Note that while Adam adaptively changes the learning rate based on the gradient statistics, SNGD employs a fixed step size. See text for more details. Best viewed in colour.\label{fig:res_bnn_mnist_200_last}}
\end{figure}
\end{landscape}

\begin{figure}[!htb]
	\centering
	\includegraphics[width=\linewidth]{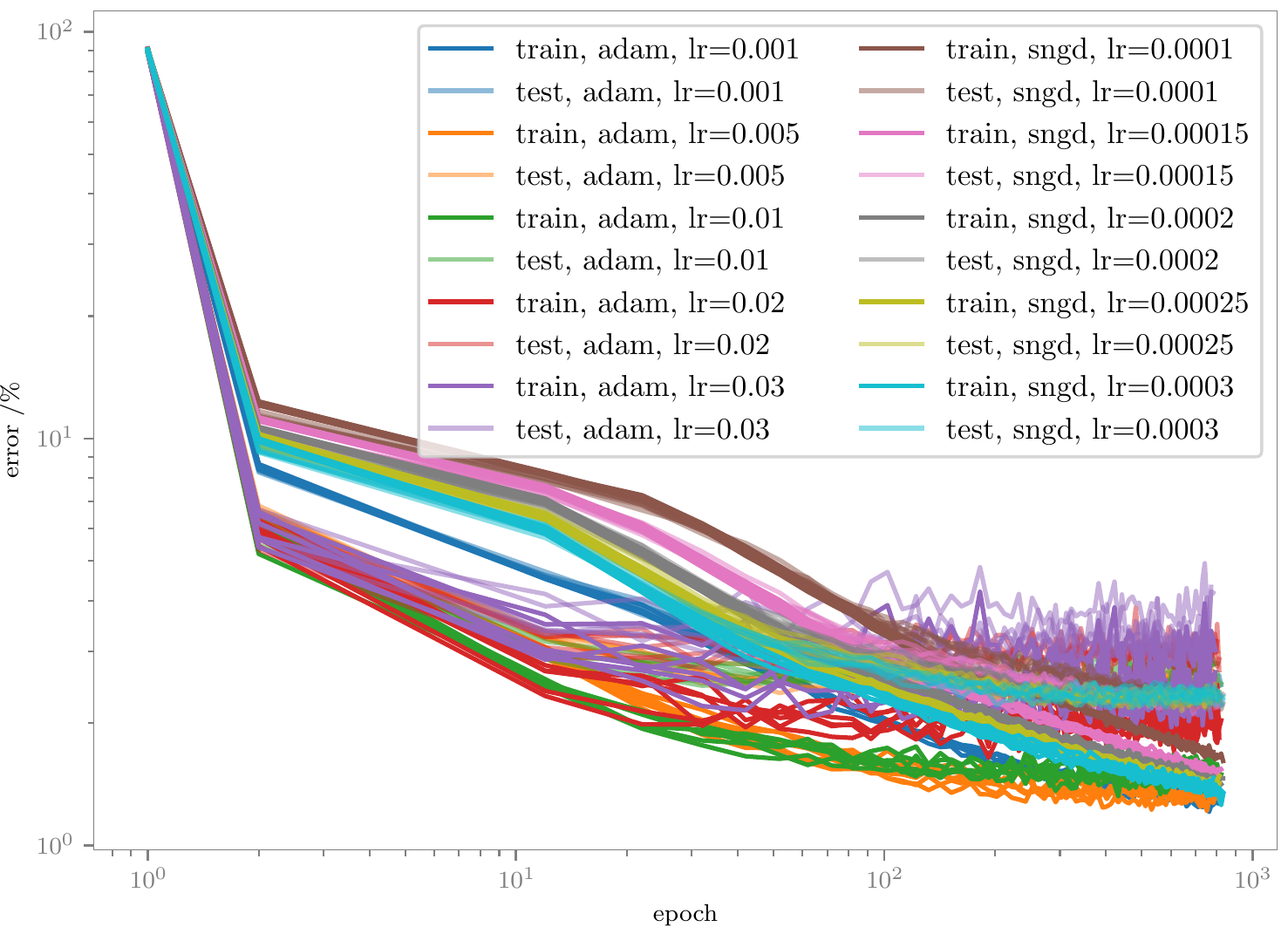}
	\caption{Classification error rates on the train and test sets during training using Adam and Stochastic Natural Gradient (SNGD) methods on the MNIST classification task with a Bayesian neural network with one hidden layer of 500 rectified linear units. The final performance of all settings are shown in \cref{fig:res_bnn_mnist_500_last}. For both Adam and SNGD, the performance highly depends on the learning rate, but the best learning rates for both methods give similar train and test results and yield similar convergence. Note that while Adam adaptively changes the learning rate based on the gradient statistics, SNGD employs a fixed step size. See text for more details. Best viewed in colour.\label{fig:res_bnn_mnist_500_error}}
\end{figure}

\begin{figure}[!htb]
	\centering
	\includegraphics[width=\linewidth]{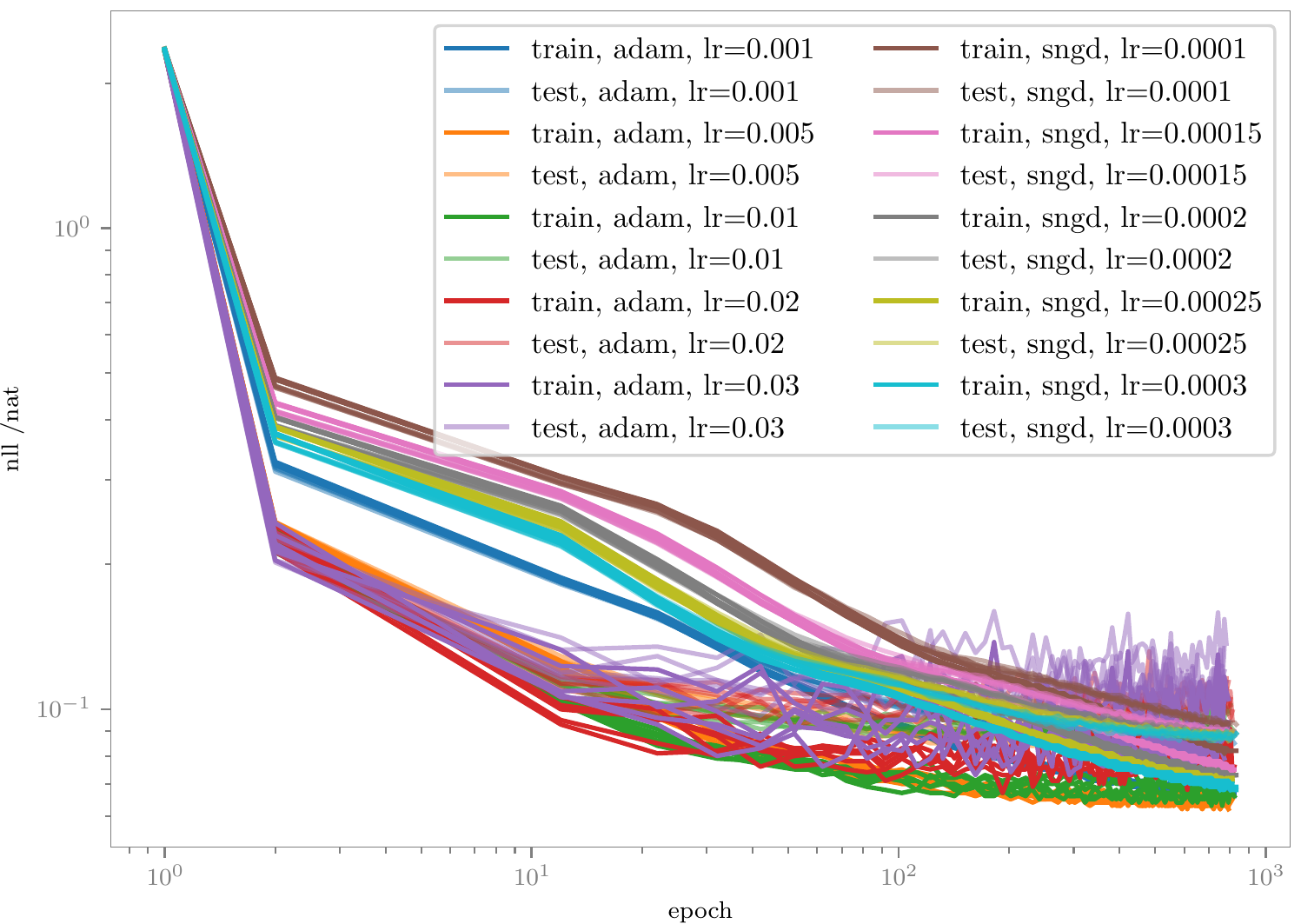}
	\caption{Negative log-likelihoods on the train and test sets during training using Adam and Stochastic Natural Gradient (SNGD) methods on the MNIST classification task with a Bayesian neural network with one hidden layer of 500 rectified linear units. The final performance of all settings are shown in \cref{fig:res_bnn_mnist_500_last}. For both Adam and SNGD, the performance highly depends on the learning rate, but the best learning rates for both methods give similar train and test results and yield similar convergence. Note that while Adam adaptively changes the learning rate based on the gradient statistics, SNGD employs a fixed step size. See text for more details. Best viewed in colour.\label{fig:res_bnn_mnist_500_ll}}
\end{figure}

\begin{landscape}
\begin{figure}
    \centering
	\includegraphics[width=\linewidth]{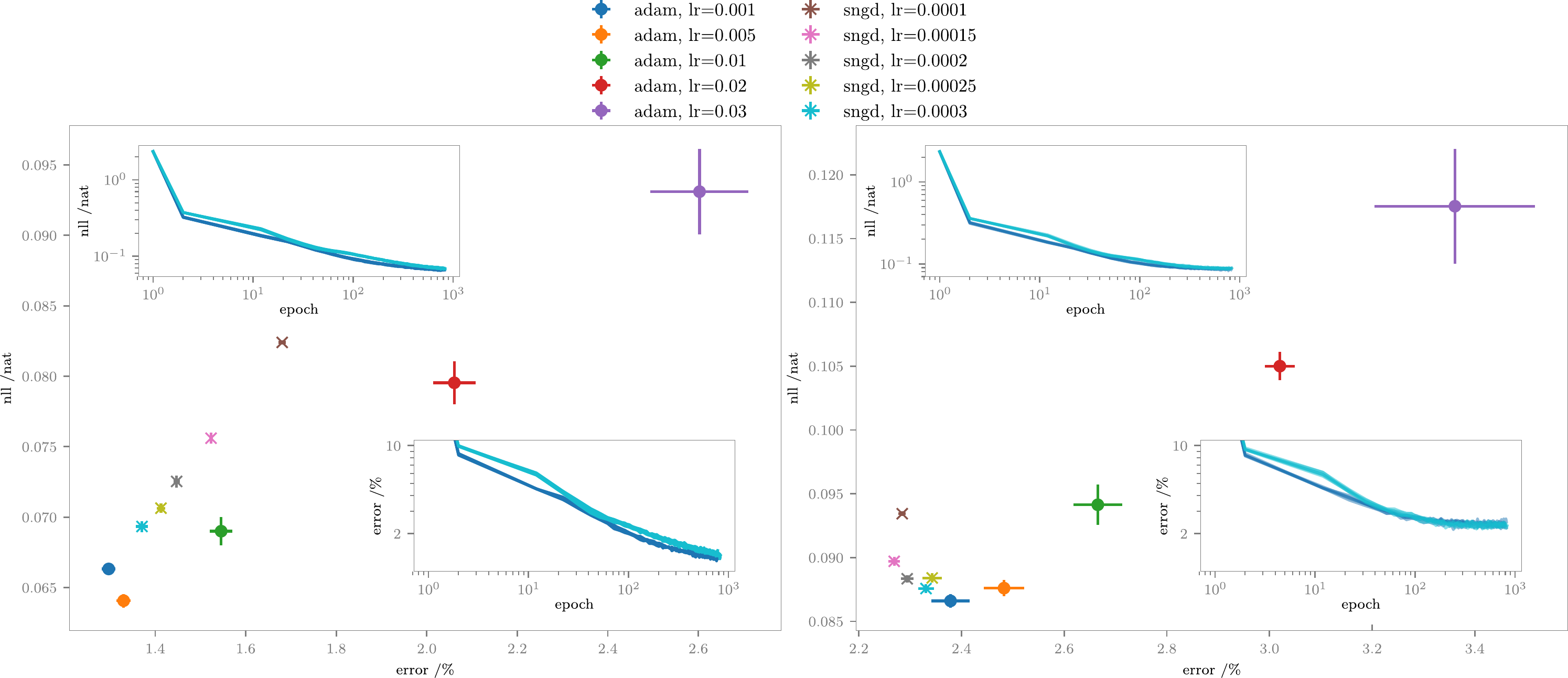}
	\caption{Performance on the train set [left] and test set [right] after 800 epochs using Adam and Stochastic Natural Gradient (SNGD) methods on the MNIST classification task with a Bayesian neural network with one hidden layer of 500 rectified linear units, and the typical performance traces as training progress [inset plots]. This figure summarizes the full results in \cref{fig:res_bnn_mnist_500_error,fig:res_bnn_mnist_500_ll}. The performance is measured using the classification error [error] and the negative log-likelihood [nll], and for both measures, lower is better and, as such, closer to the bottom left is better. Full training and test performance results are included in the appendix. For both Adam and SNGD, the performance highly depends on the learning rate, but the best learning rates for both methods give similar train and test results and yield similar convergence. Note that while Adam adaptively changes the learning rate based on the gradient statistics, SNGD employs a fixed step size. See text for more details. Best viewed in colour.\label{fig:res_bnn_mnist_500_last}}
\end{figure}
\end{landscape}

In this experiment, we stress-test various optimization methods for global variational inference for Bayesian neural networks. In particular, we consider two methods: (i) stochastic natural-gradient global VI with a fixed learning rate (SNGD, see \cref{eq:stoch-global-main}), and (ii) stochastic {\it flat} gradient global VI with an adaptive learning rate provided by Adam \citep{kingma+ba:2014}. Two Bayesian neural networks with one hidden layer of 200 or 500 Relu hidden units, and the standard MNIST ten-class classification problem are employed for this experiment. The network is trained using mini-batches of 200 data points and 800 or 1000 epochs. Both optimization methods considered have similar running time. The full results are included in \cref{fig:res_bnn_mnist_200_error,fig:res_bnn_mnist_200_ll,fig:res_bnn_mnist_500_error,fig:res_bnn_mnist_500_ll} and key results are shown in \cref{fig:res_bnn_mnist_500_last,fig:res_bnn_mnist_200_last}. It can be noticed from \cref{fig:res_bnn_mnist_500_last,fig:res_bnn_mnist_200_last} that the best versions of SNDG and Adam perform similarly in terms of both classification errors and convergence speed/data efficiency. However, both methods do require tuning of the learning rate hyperparameter. As already observed in the global VI experiment in \cref{sec:exp_federated}, signs of fast convergence early during training when using Adam do not necessarily result in a good predictive performance at the end. 

As mentioned in the main text, while natural gradients has been shown to be effective in the batch, global VI settings \citep{honkela+al:2010}, the result presented here could be seen as a negative result for natural-gradient based methods --- a stochastic natural-gradient/fixed-point method with fixed learning rate does not outperform an adaptive stochastic flat-gradient method. However, it might not be surprising as Adam adjusts its step-sizes based on approximate second-order information of the objective. This also suggests a future research venue to develop effective adaptive optimization schemes for stochastic natural-gradient variational inference.

\end{document}